\documentclass[moor,copyedit]{informs-draft}

\usepackage{endnotes}
\let\footnote=\endnote

\usepackage{natbib}
\bibpunct[, ]{(}{)}{,}{a}{}{,}%
\def\bibfont{\small}%

\graphicspath{./Figures, ./}

\TheoremsNumberedThrough     

\EquationsNumberedThrough    

\usepackage{amsmath}
\usepackage{amssymb}
\usepackage{bbold}
\usepackage{graphics}
\usepackage{graphicx}
\usepackage{placeins}
\usepackage{tikz}
\usetikzlibrary{automata,arrows,positioning,calc}
\usepackage{color}
\usepackage{algorithmic}
\usepackage{algorithm}
\usepackage{bbold}
\usepackage{paralist}
\usepackage{mathtools}

\newcommand{\Ghat}{\hat{\Gamma}}
\newcommand{\va}{a} 
\newcommand{\vb}{b} 
\newcommand{\vz}{f} 
\newcommand{\vc}{p} 
\newcommand{\vd}{d} 
\newcommand{\vp}{p} 
\newcommand{\vq}{q} 
\newcommand{\Tb}{T_B} 
\newcommand{\Tl}{T_L} 
\newcommand{\ebf}{\mathbf{\ve}}
\newcommand{\emax}{\ve_{max}}
\newcommand{\ve}{\epsilon}

\newcommand{\p}[1]{\left(#1\right)}

\newcommand{\cb}[1]{\left\{#1\right\}}

\newcommand{\EE}[2][]{\mathbb{E}_{#1}\left[#2\right]}

\DeclarePairedDelimiter{\ceil}{\lceil}{\rceil}
\DeclarePairedDelimiter\floor{\lfloor}{\rfloor} 
\newcommand{\D}{I_{\Gamma}}
\newcommand{\N}{N}
\newcommand{\e}{\omega}
\newcommand{\jlow}{\underline{J}}

\newcommand{\E}{\mathbf{E}}
\renewcommand{\P}{\mathbf{P}}
\newcommand{\feas}{\mathcal{X}}
\newcommand{\actions}{\mathcal{A}}
\newcommand{\rewards}{\mathcal{Y}}
\newcommand{\reals}{\mathbf{R}}
\newcommand{\W}{A}
\newcommand{\QE}{\hat{Q}}
\newcommand{\QI}{\tilde{Q}}

\newcommand{\V}{\mathbf{Var}}
\newcommand{\pol}{\pi}
\newcommand{\polc}{\tilde{\pi}}
\newcommand{\pold}{\hat{\pi}}
\newcommand{\polo}{\pi^*}
\newcommand{\diff}{\Delta}
\newcommand{\diffi}{\tilde{\Delta}}
\newcommand{\diffd}{\hat{\Delta}}
\newcommand{\factor}{\Gamma}
\newcommand{\factorf}{\hat{\factor}}
\newcommand{\indep}{\rotatebox[origin=c]{90}{$\models$}}

\newcommand{\PiD}{\Pi^D}
\newcounter{proofstep}
\newenvironment{proofstep}[1]
{\vspace{3pt}
	\refstepcounter{proofstep}%
	\par\textit{Step~\arabic{proofstep}:~#1}.\,}
{\vspace{2pt}}

\begin{document}
\TITLE{Offline Multi-Action Policy Learning:\\
	 Generalization and Optimization }

\ARTICLEAUTHORS{
	\AUTHOR{Zhengyuan Zhou}
	\AFF{Department of Electrical Engineering, Stanford University, \EMAIL{zyzhou@stanford.edu}}
	\AUTHOR{Susan Athey}
	\AFF{Graduate School of Business, Stanford University, \EMAIL{athey@susanathey.com}}
	\AUTHOR{Stefan Wager}
	\AFF{Graduate School of Business, Stanford University, \EMAIL{swager@stanford.edu}}
}
	
	\ABSTRACT{
In many settings, a decision-maker wishes to learn a rule, or policy, that maps from observable characteristics of an individual to an action. Examples include selecting offers, prices, advertisements, or emails to send to consumers, as well as the problem of determining which medication to prescribe to a patient. In this paper, we study the offline multi-action policy learning problem with observational data and where the policy may need to respect budget constraints or belong to a restricted policy class such as decision trees. We build on the theory of efficient semi-parametric inference in order to propose and implement a policy learning algorithm that achieves asymptotically minimax-optimal regret. To the best of our knowledge, this is the first result of this type in the multi-action setup, and it provides a substantial performance improvement over the existing learning algorithms. We then consider additional computational challenges that arise in implementing our method for the case where the policy is restricted to take the form of a decision tree. We propose two different approaches, one using a mixed integer program formulation and the other using a tree-search based algorithm.

	}


\maketitle

\section{Introduction}
As a result of digitization of the economy, more and more
decision-makers from a wide range of domains have gained the ability to target products, services, and information provision based on individual characteristics
 \citep{kim2011battle, chan2012optimizing, ozanne2014development, bertsimas2014predictive,bastani2015online, kallus2016dynamic, athey2017beyond, ban2018big, li2018}.
Examples include:

$\bullet$\textit{Health Care.}
Different medical treatment options (medicine, surgeries, therapies etc.) need to be selected
for different patients, depending on each patient's distinct medical profile (e.g. age, weight, medical history). 

$\bullet$\textit{Digital Advertising and Product Recommendations.}
Different ads, products, or offers are directed
to different consumers based on user characteristics including estimated age, gender, education background, websites previously visited, products previously purchased, etc..

$\bullet$\textit{Education.}
Digital tutors, online training programs, and educational apps select which lessons to offer a student on the basis of characteristics, including past performance.

$\bullet$\textit{ Public Policies.}
Governments may need to decide which financial-aid package (if any) should be given out to which college students.
IRS may need to decide whether to perform tax audits depending on the information provided on the filers' tax returns. 
Education boards may have to decide what type of remedial education is
needed for different groups of students.  Re-employment services may be offered to different types of unemployed workers.

A common theme underlying these treatment assignment problems is heterogeneity: different individuals respond differently to different treatments. The presence of such
heterogeneity is a blessing, as it provides us with an opportunity to personalize the service decisions, which in turn leads to improved outcomes. However, to exploit such heterogeneity, one needs to efficiently learn a personalized decision-making rule, hereafter referred to as a policy, that maps individual characteristics to the available treatments. Towards this goal, we study in this paper the problem of learning treatment assignment policies from offline, observational data, which has become increasingly available across different domains and applications as a result of recent advances in data collection technologies. The observational data has three key components: features representing the characteristics of individuals, actions representing the treatments applied to the individuals, and the corresponding outcomes observed. Our goal is then to use the collected batch of observational data to learn a policy to be used for future treatments. 

As recognized by the existing literature (discussed in more detail in Section~\ref{subsec:related}), there are several difficulties that make the offline policy learning problem challenging. First, counterfactual outcomes are missing in the observational data: We only observe the outcome for the action that was selected in the historical data, but not for any other action that could have been selected.
Second, unlike in online policy learning, we do not have control over action selection
in the offline case, and this can lead to selection bias: There may be an association
between who receives a treatment and patient-specific characteristics (e.g., in a medical application,
sicker patients may be more likely to receive a more aggressive intervention), and a failure to account for selection
effects will lead to inconsistent estimation of the benefits of counterfactual policies, and thus inconsistent estimation of the optimal treatment assignment rule \citep{rubin1974estimating,heckman1979sample}.
Moreover, we may not have knowledge about the form of these selection effects because the historical policy that was used to collect the observational data may not be explicit (e.g. in the case of Electronic Medical Records) or available (e.g. in the case of customer log data from technology companies), and the analyst thus must learn them from the data.
Third, in many applications (particularly health care and public policy), the decision maker is further constrained to consider policies within a restricted set, due to considerations such as interpretability or fairness. This often creates additional computational challenges as the resulting learned policy must respect such preimposed constraints.

In this paper we tackle these challenges in the context of multi-action policy learning problem, and focus on the general setting where the historical treatment assignment mechanism is not necessarily known \textit{a priori} and must be estimated from the data. 
Suppose that we have $n$ independent and identically distributed samples $(X_i, \, A_i, \, Y_i)$, where
$X_i \in \mathcal{X}$ denote pre-treatment features (also called covariates), $A_i \in \cb{a^1, \, ..., \, a^d}$ denotes the action taken, and $Y_i$ is
the observed outcome. Then, following the potential outcomes framework \citep{imbens2015causal}, we posit
potential outcomes $\cb{Y_i(a^j)}_{j = 1}^d$ corresponding to the reward we would have received by taking any
action $a^j$, such that $Y_i = Y_i(A_i)$. A policy $\pi$ is a mapping from features $x \in \mathcal{X}$
to a decision $a \in \cb{a^1, \, ..., \, a^d}$, and the expected reward from deploying a policy $\pi$ is $Q(\pi) = \EE{Y_i(\pi(X_i))}$.
Given this setting, a fundamental problem in off-policy evaluation (i.e., in estimating $Q(\pi)$) is unless $\pi(X_i)=A_i$, the realized
outcome $Y_i(A_i)$ that we actually observe for an individual is not equal to the potential outcomes $Y_i(\pi(X_i))$ we would
have observed under $\pi$.  Estimating $\EE{Y_i(\pi(X_i))}$ is thus a challenging problem.
Our main assumption is that we have observed enough covariates to explain any selection
effects, and that actions $A_i$ observed in the training data is as good as random once we condition on $X_i$.
\begin{equation}
\label{eq:unconf}
\cb{Y_i(a^j)}_{j = 1}^d \, \indep\,  A_i \ \big|\ X_i.
\end{equation}
If this \emph{unconfoundedness} assumption holds \citep{rosenbaum1983central},
and we are furthermore willing to assume that the propensity scores $e_a(x)$ are known,
then it is well known (see \citet{swaminathan2015batch}) that we can obtain $\sqrt{n}$-rates for $Q(\pi)$ via inverse-propensity weighting (IPW),
\begin{equation}
\label{eq:ipw}
\widehat{Q}_{IPW}(\pi) = \frac{1}{n} \sum_{i = 1}^n \frac{1\p{\cb{A_i = \pi(X_i)}} Y_i }{ e_{A_i}(X_i)}.
\end{equation}
Further, we can consistently learn policies by taking
$\hat{\pi} = \argmax\cb{\widehat{Q}_{IPW}(\pi) : \pi \in \Pi}$,
where $\Pi$ is the set of policies we wish to optimize over, and 
this strategy attains the optimal $1/\sqrt{n}$ rate of convergence.
As discussed in \citet{kitagawa2015should}, the class $\Pi$ can play many valuable roles: It can encode
constraints relative to fairness ($\pi$ shouldn't unfairly discriminate based on protected characteristics),
budget (we can't treat more people than our resources allow),
functional form (in many institutions, decision rules need to take a simple form so that they can be deployed
and audited), etc.

In the case where propensity scores $e_a(x)$ are not known a-priori, however, existing results are not
comprehensive despite a considerable amount of work across several fields \citep{manski2004statistical,
hirano2009asymptotics, stoye2009minimax, langford2011doubly, zhang2012estimating,
zhao2012estimating, zhao2014doubly, kitagawa2015should, swaminathan2015batch, athey2017efficient,
zhou2015residual, kallus2018confounding}.
As shown in \citet{kitagawa2015should}, if we have consistent estimates \smash{$\hat{e}_a(x)$}
of the propensities and plug them into \eqref{eq:ipw}, we can still learn a consistent policy $\hat{\pi}$
by maximizing $\widehat{Q}$. But unlike in the case with known propensities, the regret of $\hat{\pi}$
will now depend on the rate at which converges \smash{$\hat{e}_a(x)$} in root-mean-squared error; and
in all but the simplest cases, this will result in sub-optimal slower-than-$1/\sqrt{n}$ rates for $\hat{\pi}$.
This leads us to our main question: What is the optimal sample complexity for multi-action policy learning
in the observational setting with unknown treatment propensities, and how can we design practical algorithms
that attain this optimal rate?

\subsection{Our Contributions}

We study a family of doubly robust algorithms for multi-action policy learning,
and show that they have desirable properties both in terms of statistical generalizability and
and computational performance. At a high level, these algorithms start by solving non-parametric
regression problems to estimate various ``nuisance components,'' including the propensity score,
and then using the output of these regressions to form an objective function that can be used to
learn a decision rule in a way that is robust to selection bias.

From a generalization perspective, our main result is that the doubly robust approach to policy learning
can achieve the minimax optimal
$O_p(\frac{1}{\sqrt{n}})$ decay rate for regret, even if the non-parametric regressions used to estimate
nuisance components may converge slower than $O_p(\frac{1}{\sqrt{n}})$. This is in contrast to standard
methods based on inverse-propensity weighting using \eqref{eq:ipw}: As a concrete example, in a problem
where all nuisance components are estimated at $O_p(\frac{1}{n^{1/4}})$ in root-mean-squared error,
then methods based on
\eqref{eq:ipw} only satisfy $O_p(\frac{1}{n^{1/4}})$ regret bounds, whereas we prove $O_p(\frac{1}{\sqrt{n}})$
bounds.
In earlier work, \citet{swaminathan2015batch} proved regret bounds for multi-action policy learning; however,
they assume that propensities are known and get slower-than-$O_p(\frac{1}{\sqrt{n}})$ rates. Meanwhile, \citet{athey2017efficient}
provide results on $O_p(\frac{1}{\sqrt{n}})$-regret policy learning with observational data, but their results only
apply to the binary setting (see Section~\ref{subsec:related} for a detailed discussion).
Our analysis builds on classical results on semiparametrically efficient treatment effect estimation from the
causal inference literature \citep{newey1994asymptotic,robins1995semiparametric,hahn1998role,belloni2017program} and,
as in \citet{athey2017efficient}, we find that these tools enable us to considerably sharpen our results.
To the best of our knowledge, this is the first minimax optimal learning algorithm for the general, multi-action offline policy learning problem.

Meanwhile, on the optimization front, we provide  
two concrete implementations of CAIPWL for tree-based policy class, a widely used class of policies in practical settings due to its interpretability: one based on mixed integer program (MIP) and the other based on tree search.
In the first implementation, inspired by the recent work~\cite{bertsimas2017optimal},
we formulate the problem of CAIPWL with trees as a MIP. 
In particular, by solving this MIP, we can find an \textit{exact optimal} tree
that maximizes the policy value estimator.
An important benefit of the MIP formulation is that it conveniently allows us to take advantage of the rapid development of high-performance commercial MIP solvers as well as the quickly growing availability of computing power (examples includes GUROBI~(\cite{gurobi}) and CPLEX~(\cite{cplex})): one can simply 
translate the MIP formulation into code and directly call the underlying blackbox solver.
While the MIP formulation provides a convenient way to address policy optimization that can leverage the off-the-shelf solvers, the scale that it can currently handle is still quite limited. 
Motivated by the computational concerns, we develop a customized tree-search based algorithm that again finds the exact optimal tree in the policy optimization step. In comparison to the MIP approach, this can be viewed as a white-box approach, where the running time can now be easily analyzed in terms of the problem-specific parameters (e.g. number of data points, tree-depth, feature dimension, number of actions).  It can also be controlled by limiting the space of trees that the algorithm searches, e.g. by only considering a limited number of ``split points'' for each covariate.
In our applications, this latter strategy enables us to scale up to larger problems.

We view our result as complementary to the growing recent literature on online policy learning (i.e. online contextual bandits). When possible,
online policy learning presents a power joint approach to exploration and exploration. In some settings, however, online policy learning may be difficult or impossible as
decision rules may need to be stable over time for a variety of reasons: Medical treatment guidelines
may need peer or scientific review, regulators may be required to review changes in bank lending guidelines, and firms may wish to avoid frequent disruptions to their processes.  Implementation of decisions may need to be delegated to humans who may need to be retrained each time decision rules change.  In addition, the growing capabilities of firms and governments (hospitals, technology companies, educational institutions etc.) in the areas of collecting and maintaining data, as well as the increasing trend of running randomized experiments,  imply that many decision-makers have access to large historical datasets that can be used for offline learning.  Given that treatment effects may be small and
outcomes noisy, large historical datasets may be helpful for estimating policies even in environments
where ongoing online experiments are possible.  Finally, estimating and evaluating a treatment assignment policy in an offline setting may be the first step towards understanding the benefits of further investment in online learning. Thus, we believe it is important to understand the statistical difficulty of offline policy learning.

\subsection{Related Work}\label{subsec:related}
In the past decade, the importance and broad applicability of this emerging area (offline policy learning) have drawn a rapidly growing line of research efforts (\cite{langford2011doubly, zhang2012estimating,zhao2012estimating, zhao2014doubly, kitagawa2015should,swaminathan2015batch, athey2017efficient, zhou2015residual, kallus2018confounding}).
In the challenging landscape mentioned above, the existing literature has mostly focused on
binary-action policy learning (i.e. only two actions exist, typically referred to as the control action and the treatment action); notable contributions include~\citep{zhang2012estimating,zhao2012estimating, zhao2014doubly, kitagawa2015should, athey2017efficient}.
\cite{zhao2014doubly} proposed an algorithm and established a $O_p(\frac{1}{n^{\frac{1}{2+1/q}}})$ regret bound\footnote{\cite{zhao2012estimating}'s bound is worse than the bound in~\cite{zhao2014doubly} and \cite{zhang2012estimating} did not provide any bound.} ($0 < q < \infty$ is a problem-specific quantity).
\cite{kitagawa2015should} further improved it to $O_p(\frac{1}{\sqrt{n}})$ (by a different learning algorithm), albeit with the assumption
that the underlying propensities are exactly known. Further, \cite{kitagawa2015should}
established a matching lower bound of $\Omega_p(\frac{1}{\sqrt{n}})$ for policy learning.
\cite{kitagawa2015should} also provides a regret bound for the more general but harder case where the propensities are not known; however, that regret bound
does not have the optimal $O_p(\frac{1}{\sqrt{n}})$ dependence.
More recently, \cite{athey2017efficient} designed another learning algorithm and established the same $O_p(\frac{1}{\sqrt{n}})$ regret bound (with sharper problem-specific constants that rule out certain types of learning algorithms) even when propensities are unknown (also extending their analysis to some cases where unconfoundedness fails but instrumental variables are available). Consequently, this line of work produced a sequence of refinements, culminating in the optimal regret rate $O_p(\frac{1}{\sqrt{n}})$ for binary-action policy learning.
 
Unfortunately, the existing policy learning algorithms for the binary-action are not directly generalizable to the general, multi-action setting for two reasons: First, most of them (particularly~\cite{kitagawa2015should}, \cite{athey2017efficient}) rely on the special embedding 
 of the two actions as $+1$ and $-1$ in the learning algorithms. Hence it's not clear what embedding scheme would be effective for multi-action settings. In particular, these algorithms rely on estimating the impact of assigning the treatment action rather than the control action, something known as treatment effect in the causal inference literature. This treatment effect estimate plays a central role in the algorithm and the theory, making the extension to the multi-action case non-trivial. It is also important to point out that not only are the algorithms in binary-action learning not applicable, but the state-of-the-art proof techniques developed therein are also far from being sufficient. Take~\cite{athey2017efficient} for example, where a sharp analysis is used to obtain a tight regret bound for a customized doubly robust algorithm in the binary-action case. There, the regret analysis depends crucially on bounding the Rademacher complexity, a classical quantity that measures how much the policy class can overfit. Although Rademacher complexity is unambiguously defined in binary-action settings, it is not clear which generalization should be used in multi-action settings. In fact, different generalizations of Rademacher complexity exist \citep{rakhlin2016bistro, kallus2017balanced}, and not choosing the proper generalizaiton can potentially lead to suboptimal bounds (this is discussed in more detail in Section~\ref{subsec:comparisons}).

Although the literature on offline multi-action policy learning as not as well developed as the literature on
the binary case, \citet{swaminathan2015batch}, \citet{zhou2015residual} and
\citet{kallus2016learning,kallus2017balanced} have recently proposed methods with formal consistency guarantees.
However, none of the existing regret bounds has an optimal dependence on the sample size $n$ (see Section~\ref{subsec:comparisons} for a detailed comparison). This leaves open a question of both theoretical and practical significance, namely which algorithm should be used for the multi-action policy learning problem in order to achieve maximum statistical efficiency.


In closing, for completeness, we mention that in online policy learning, (asymptotically) optimal algorithms are known for a class of problems known as linear contextual bandits, where the underlying reward model is linear in the features/contexts (also known as the linear realizablity assumption). 
In particular, UCB and Thompson sampling style algorithms are both provably optimal in linear contextual bandits: they both enjoy the $O(\sqrt{T})$ cumulative regret,\footnote{Depending on the specific variants used, there can be certain logarithmic factors hiding in $O(\cdot)$.} which are known to be tight (with respect to the dependence on $T$). Note that the time horizon $T$ in the online context corresponds to $n$ in the offline context: typically one data sample is received in each online iteration. 
The very recent work~\cite{li2017provably} has further extended to generalized linear contextual bandits.  These methods all rely on functional form assumptions (such as linearity of the outcome model for each possible treatment), which greatly simplifies the estimation problem.  In contrast, the literature on estimating the impact of a treatment (reviewed in \cite{imbens2015causal}) has focused on the case where the functional form of the outcome model is unknown, and indeed a large body of work demonstrates that functional form assumptions matter a lot in practice and can lead to misleading results. Much of the recent offline policy evaluation literature follows in this tradition, and the efficiency results of \cite{athey2017efficient} do not make use of functional form assumptions, but instead build on results from semi-parametric econometrics and statistics. In this paper, we also follow the semi-parametric approach.

Beyond functional form assumptions, it is important to point out that one cannot directly turn an online bandit algorithm into an offline policy learning problem via the standard online-to-batch conversion. Emulating an online bandit algorithm when only historical data is available would require the decision maker to select an action that is different from the actual data collector's action on a particular iteration; but outcome data for counterfactual actions is not available in historical data.  
 Consequently, offline policy learning algorithms are conceptually distinct.
For further discussion of the online learning literature, see, e.g., \cite{ lai1985asymptotically,besbes2009dynamic,Li:2010:CAP:1772690.1772758,rigollet2010nonparametric,chu2011contextual,abbasi2011improved,bubeck2012regret, agrawal2013thompson, russo2014learning,bastani2015online,li2017provably, dimakopoulou2017estimation}.

\section{Problem Setup}\label{sec:model}
Let $\actions$ be the set of $d$ possible actions: $\mathcal{A} = \{a^1, a^2, \dots, a^d\}$.
For notational convenience, throughout the paper, we identify the action set $\actions$ with the set of $d$-dimensional standard basis vectors: $a^j = (0, 0, \dots, 1, \dots, 0)$, where $1$ appears and only appears in the $j$-th component and the rest are all zeros.
Following the classical potential outcome model~(\cite{neyman1923applications, rubin1974estimating}), we posit the existence of a fixed underlying data-generating distribution on $(X, Y(a^1), Y(a^2), \dots, Y(a^d)) \in \feas \times \prod_{j=1}^d \rewards_j$, where $\feas$ is an arbitrary feature set (but typically a subset of $\mathbb{R}^p$),
and each $Y(a^j)$ denotes the random reward realized under the feature vector $X$,  when the action $a^j$ is selected.

Let $\{(X_i, \W_i, Y_i)\}_{i=1}^n$ be $n$ \textbf{iid} observed triples that comprise of the observational dataset,
where $(X_i, Y_i(a^1), \dots, Y_i(a^d))$ are drawn \textbf{iid} from the fixed underlying distribution described above. Further, in the $i$-th datapoint $(X_i, \W_i, Y_i)$, $A_i$ denotes the action selected 
and $Y_i = Y_i(\W_i)$. In other words, $Y_i$ in the $i$-th datapoint is the observed reward under the feature vector $X_i$ and action $A_i$. 
Note that all the other rewards $Y_i(a)$ (i.e. for $a \in \actions - \{A_i\}$), even though they exist in the model (and have been drawn according to the underlying joint distribution), are not observed.
For any $(x, a) \in \feas \times \actions$, we define the following two quantities:

\begin{definition}\label{def:two_quantities}\quad
$e_{a}(x) \triangleq \P[\W_i = a \mid X_i = x]$
and $\mu_{a} (x)  \triangleq \E[Y_i(a) \mid X_i = x].$ 
\end{definition}

\begin{remark}
	The above setup is a standard model that is also used in
	contextual bandits~(\cite{bubeck2012regret}), where each action $a^j$ is known as an arm and the feature vector $x$ is called a context. 
	In general contextual bandits problems, $\mu_{a} (x)$ can be an arbitrary function of $x$ for each $a$. When $\mu_{a} (x)$ is a linear function of $x$, this is known as linear contextual bandits, an important and perhaps most extensively studied subclass of contextual bandits in the online learning context.
	In this paper, we do not make any structural assumption on $\mu_{a} (x)$ and instead work with general
	underlying data-generating distributions. 
	Furthermore, it should also be clear that the problem setup considered here is inherently offline (as opposed to online), since we work with data that is previously collected in one batch. 
\end{remark}

We make the following assumptions about the data-generating process, which are standard in the literature:
\begin{assumption}\label{assump:classical}\quad
	The joint distribution $(X, Y(a^1), Y(a^2), \dots, Y(a^d), A_i)$ satisfies:
	\begin{enumerate}
		\item Unconfoundedness: 
		$(Y_i(a^1), Y_i(a^2), \dots, Y_i(a^d)) \indep \W_i \mid X_i$.
		\item Overlap:
		There exists some $\eta > 0$,  $e_a(x) \ge  \eta$, $\forall (x, a) \in \feas \times \actions$.
		\item Bounded reward support: $(Y(a^1), Y(a^2), \dots, Y(a^d))$ is supported on a bounded set in $\mathbf{R}^d$.
	\end{enumerate}
\end{assumption}

\begin{remark}\quad
The unconfoundedness assumption says that the action taken is indepedent of all the reward outcomes conditioned on the feature vector. 
We emphasize that the actions generated this way provide a significant relaxation of randomized trials, a widely adopted method 
in practice for collecting data, where $A_i$ is chosen at random, independent of everything else. The overlap assumption says that any action should have some minimum probability of being selected, no matter what the feature is. 
Both of these assumptions are standard and commonly adopted in both the estimation literature (\cite{rosenbaum1983central, imbens2004nonparametric, imbens2015causal, athey2016approximate}) and the existing offline policy learning literature (\cite{zhang2012estimating,zhao2012estimating, kitagawa2015should,swaminathan2015batch, zhou2015residual}).
In particular, in the estimation literature, a common setup considers the binary action case (where $|\actions| = 2$); and the goal is to estimate the treatment effect. It is well-known (\cite{rosenbaum1983central}) that these two assumptions enable identification of causal effects in observational studies.
Finally, the bounded support assumption is also standard in the existing literature. Mathematically, it is not essential and can be generalized to unbounded but sub-Gaussian random variables. However, we proceed with bounded rewards for two reasons: first, it simplies certain notation and proofs (the existing theoretical framework is already quite complex); second, outcomes in most practical applications are bounded; hence this assumption is not restrictive from a practical standpoint.
\end{remark}

With the above setup, our goal is to learn a good policy from a fixed ambient policy class $\Pi$ using the observational dataset. A policy $\pi: \feas \rightarrow \Delta(\actions)$ is a function that 
maps a feature vector $x$ to a point
 in the probability simplex\footnote{Here we are considering randomized policies. They contain as a subclass the deterministic policies $\pi: \feas \rightarrow \actions$, which is a map that specifies which one of the $d$ actions to take under a given feature $x$.
Although in the current setup where data is drawn \textbf{iid} from some fixed distribution, 
there is no need to make the distinction: there is always a deterministic policy that achieves the optimal value even if we enlarge the policy class to include randomized policies. Randomized policies can have strictly better values in advesarial contextual bandits (where an adversary adaptively and adversarially chooses the features), a case that belongs to online learning context and that does not concern us here.
} $\Delta(\actions)$ of the action set.
For a given policy
$\pi \in \Pi$, the performance of $\pi$ is measured by the expected reward this policy generates, as characterized by the policy value function:
\begin{definition}\quad
The policy value function $Q: \Pi \rightarrow \reals$ is defined as: $Q(\pi) = \E[Y(\pi(X))]$, where the expectation is taken with respect to the randomness in both the underlying joint distribution and the random policy.
\end{definition}

With this definition, the optimal policy $\pi^*$ is a policy that maximizes the policy value function:
$\pi^* = \arg\max_{\pi \in \Pi} = \E[Y(\pi(X))]$.
The objective in the current policy learning context is to learn a policy $\pi$ that has the policy value as large as possible, or equivalently, to minimize the discrepancy between the performance of the optimal policy and the performance of the learned policy $\pi$.
This discrepancy is formalized by the notion of regret, as given in the next definition: 
\begin{definition}\quad
The regret $R(\pi)$ of a  policy $\pi \in \Pi$ is defined as:
$$R(\pi) =\E[Y(\pi^{*}(X))] - \E[Y(\pi(X))].$$
\end{definition}

Several things to note.
First, per its definition, we can rewrite regret as 
$R(\pi) = Q(\polo) - Q(\pi)$.	
Second, the underlying random policy that has generated the observational data (specifically the $A_i$'s)
need not be in $\Pi$.
Third, when a policy $\hat{\pi}$ is learned from data and hence a random variable (as will be the case in the current policy learning context),
$R(\hat{\pi})$ is a random variable. 
A regret bound in such cases is customarily a high probability bound that highlights
how regret scales as a function of the size $n$ of the dataset, the error probability and other important parameters of the problem (e.g. the complexity of the policy class $\Pi$).

\section{Algorithm: Cross-fitted Augmented Inverse Propensity Weighted Learning}
\label{sec:alg}

We propose an offline policy learning algorithm (which includes two implementation variants discussed in more detail later), called Cross-fitted Augmented Inverse Propensity Weighted Learning (hereafter abbreviated as CAIPWL for short). CAIPWL has three main components:  (i) policy value approximation via an Augmented Inverse Propensity Weights (AIPW) estimator, (ii) a $K$-fold algorithmic structure for computing ``scores'' for each observation, and (iii) policy optimization (via policy-class dependent procedures). Each component merits further discussion. 

 First, the AIPW estimators (\cite{robins1994estimation}), sometimes also called doubly robust estimators, are a well-known family of estimators that are widely used in the literature on the estimation of causal effects. In particular, doubly robustness refers to the following property of estimators: a doubly robust estimator of treatment effects is consistent if either the estimate of each unit's treatment assignment probability as a function of characteristics (the propensity score) is consistent or the estimate of a unit's expected outcome for each treatment arm given covariates (the outcome model) is consistent (see~\cite{scharfstein1999adjusting}). Different variants of AIPW/doubly robust estimators exist, and policy learning algorithms derived from those estimators are broadly referred to as doubly robust learning algorithms~\cite{zhang2012estimating, zhao2014doubly, athey2017efficient}).
 Among approaches using doubly robust estimators, the work in~\cite{langford2011doubly} is the most directly related to ours here. In particular, ~\cite{langford2011doubly}  used the same AIPW estimator as ours to evaluate counterfactual policies, a key step in policy learning. However, ~\cite{langford2011doubly} did not study the policy learning problem: the efficiency gains of using AIPW to learn a policy are not characterized. Further, AIPW by itself is not sufficient for policy learning as the $K$-fold cross-fitting component of our algorithm is quite important in obtaining the overall regret performance.

The second component of our approach, using the $K$-fold structure, builds on a commonly employed method in statistical inference (see~\cite{schick1986asymptotically, chernozhukov2016double, athey2017efficient}) to
reduce model overfit. It is closely related to cross-validation, in that the data is divided into folds to estimate models using all data save one fold.  However, the $K$-fold structure in CAIPWL is not used to select hyperparameters or tune models.  Instead, CAIPWL requires a ``score'' to be constructed for each observation and each treatment arm, and some components of the score must be estimated from the data. The $K$-fold structure is used for estimating these scores, ensuring that the estimated components for each unit are independent of the outcome for that unit.  This independence property is crucial for the theoretical guarantees, and also is important for practical performance, as it reduces generalization error. 

The third component is policy optimization, which concerns selecting a policy that maximizes 
the value estimate constructed from the AIPW scores. This is typically the most computationally intensive step in the entire learning algorithm, and it constitues a crucial component in policy learning: a step that generates the learned policy by optimizing an objective function. The specific implementation depends on the particular policy class that is \textit{a priori} decided upon by the decision maker. We discuss two implementations that achieve exact optimality for decision trees, arguably the most widely used class of policies. 

We next discuss CAIPWL in detail. First, we estimate the two quantities
given in Definition~\ref{def:two_quantities}: $e_{a}(x)$
and $\mu_{a} (x)$.
To do so, we divide the dataset into $K$ evenly-sized folds (hence the name cross-fitted) and for each fold $k$, estimate
$e_{a^1}(x), e_{a^2}(x), \dots, e_{a^d}(x)$ and $\mu_{a^1} (x), \mu_{a^2} (x), \dots, \mu_{a^d} (x)$ using the rest $K-1$ folds. We denote the resulting estimates by $\hat{e}_{a^j}^{-k}(x)$ and  
$\hat{\mu}^{-k}_{a^j} (x)$: $\hat{e}_{a^j}^{-k}(x)$ and  
$\hat{\mu}^{-k}_{a^j} (x)$ are estimates of $e_{a^j}(x)$ and  
$\mu_{a^j} (x)$ using all the data except fold $k$.

Highlighting the dependence on the number of data points used in constructing the estimates,
we use $(\hat{e}_{a^j}^{(n)}(x), \hat{\mu}^{(n)}_{a^j} (x))$ to denote one such generic pair of estimators. Note that $\hat{e}_{a^j}^{-k}(\cdot)$ only uses $\frac{n(K-1)}{K}$ points in the training data to build the estimator because fold $k$ is excluded.
Note also that for different $k = 1, 2,\dots, K$, the estimates $(\hat{e}_{a^j}^{-k}(x),  
\hat{\mu}^{-k}_{a^j} (x))$ can be obtained via different ways (i.e. using different choices of estimators).
In particular, we can draw upon the existing and extensive literaure on parametric, semi-parametric and non-parametric estimation (\cite{hastie2013elements}) to produce effective estimators for the quantities of interest, depending on the particular setup at hand.
We only require each of these $K$ estimators to satisfy the following consistency assumption: 

\begin{assumption}\label{assump:consistency}\quad
All estimators $\hat{e}_{a^j}^{(n)}(x)$, $\hat{\mu}^{(n)}_{a^j} (x)$ used satisfy the following errror bounds:
\begin{equation}\label{assump:consistency_eqv}
\E\Bigg[ \Big( \hat{\mu}_{a^j}^{(n)}(X) - \mu_{a^j}(X) \Big)^2 \Bigg] \cdot
\E\Bigg[ \Big( \hat{e}_{a^j}^{(n)}(X) - e_{a^j}(X) \Big)^2 \Bigg] = \frac{o(1)}{n},
\forall a^j \in \actions.
\end{equation}
\end{assumption}

\begin{remark}\quad
The above assumption is not strong or restrictive. In fact, ~\cite{farrell2015robust} has made clear the importance of Assumption~\ref{assump:consistency} in achieving accurate treatment effect estimation (in binary treatment cases) for doubly robust/AIPW estimators.
Furthermore, the standard $\sqrt{n}$-consistency assumption in the semi-parametric estimation literature (\cite{chernozhukov2016double}) given in Equation~\eqref{eq:help}:
\begin{equation}\label{eq:help}
\E\Bigg[ \Big( \hat{\mu}_{a^j}^{(n)}(X) - \mu_{a^j}(X) \Big)^2 \Bigg] \le \frac{o(1)}{\sqrt{n}}, \quad
\E\Bigg[ \Big( \hat{e}_{a^j}^{(n)}(X) - e_{a^j}(X) \Big)^2 \Bigg] \le \frac{o(1)}{\sqrt{n}},
\forall a^j \in \actions,
\end{equation}
is a special case of Assumption~\ref{assump:consistency}.
On the other hand, the condition given in Assumption~\ref{assump:consistency} is more general and flexible as it allows one to trade-off the accuracy of estimating 
$\mu_{a}(x)$ with the accuracy of estimating $e_a(x)$: it suffices for the product of the two error terms to grow sublinearly.
Furthermore, whether estimators $\hat{e}_{a^j}^{(n)}(x)$, $\hat{\mu}^{(n)}_{a^j} (x)$ exist that satisfy 
Assumption~\ref{assump:consistency} depends on the regularity of the underlying function being estimated. In general, provided $\mu_a(\cdot)$ and $e_a(\cdot)$ are sufficiently smooth, Assumption~\ref{assump:consistency} easily holds (and many semi-parametric or non-parametric estimators fulfill the requirement). The precise such conditions as well as the difference choices of estimators have been extensively studied in the estimation literature (see for instance~\cite{newey1994asymptotic, robins1995semiparametric, bickel,farrell2015robust, belloni2017program, hirshberg2017balancing,newey2018cross}). Below we give four classes of examples:
\begin{enumerate}
	\item \textbf{Parametric Families} When $\mu(\cdot)$ or $e(\cdot)$ is parametric (such as linear models or generalized linear models), estimation can be done efficiently: the achievable mean square error is
	$O(\frac{1}{n})$. In such settings (assuming both are parametric), the product error bound is $O(\frac{1}{n^2})$.
	\item \textbf{Holder classes.}
	When $\mu(\cdot)$'s or $e(\cdot)$'s $(\beta - 1)$-th derivative
	is Lipschitz, and when the feature space $\feas$ has dimension $d$ (i.e. $\feas \subset \mathbb{R}^d$), then the mean-squared-error
	is $O(n^{-\frac{2\beta}{2\beta+d}})$. Hence, for $d < 2\beta$, each of the two corresponding mean squared errors will be $o(n^{-\frac{1}{2}})$, in which case Assumption~\ref{assump:consistency} holds. More generally, assume $\mu(\cdot)$'s and $e(\cdot)$ belong to the Holder class of functions $\mathcal{H}_{\beta, \alpha}$,
	which is the set of all functions that map $\mathcal{X} \subset \mathbb{R}^d$ to $\mathbb{R}$ such that
	$\|D^{(\beta)} (x) - D^{(\beta)} (y)\|_2 \le L \|x - y\|_2^{\alpha}$, for some $0< \alpha < 1$ and some $L > 0$, where $D^{(k)}(\cdot)$ gives the $k$-th order partial derivatives. Then the achievable mean square 
	error is $O(n^{-\frac{2(\beta+\alpha)}{2(\beta+\alpha)+d}})$.
	\item \textbf{Sobolev classes.} If $\mu(\cdot)$ or $e(\cdot)$ has bounded $\beta$-th derivatives, with feature space of dimension $d$,
	then the achievable mean-squared-error is  $O(n^{-\frac{2\beta}{2\beta+d}})$.
	\item \textbf{Reproducing Kernel Hilbert Space (RKHS).}
	Let $K(x, x)$ be the kernel associated with a RKHS that contains
	$\mu(\cdot)$ (or $e(\cdot)$).
	Suppose $K$ belongs to some Besov space $B_{2,\infty}^{\alpha}$ and $\feas \in \mathbb{R}^p$ is locally the graph of a Lipschitz function. Then 
	the mean-squared-error is $O(\frac{1}{n^{l+1}})$, where $l = \frac{1}{ \frac{\alpha}{p}+ 1}$.
	Consequently, since $l < 1$, $o(\frac{1}{\sqrt{n}})$ rate is always achievable when the kernel satisfies the smoothness property. For more, see~\cite{mendelson2010regularization}.
	An RKHS with the Gaussian kernel and all finite dimensional RKHSs are simple special cases.
\end{enumerate}

\end{remark}

With those estimators, we can now  learn a policy as follows. 
As a way to approximate the value function $Q$ for a given policy $\pi \in \Pi$, we first define the agumented inverse propensity weighting estimator
$\QE_{CAIPWL}$:
$\QE_{CAIPWL} (\pi) = \frac{1}{n} \sum_{i=1}^n \langle \pi(X_i), \factorf_i \rangle$,
where $\langle \cdot, \cdot \rangle$ denotes the inner product (note that $\pi(X_i)$ is one of the $d$-dimensional basis vectors ) and $ \factorf_i$ is the following vector constructed from the data and the $K$-folds:
\begin{equation}\label{eq:urgent}
\factorf_i =  \frac{Y_i - \hat{\mu}_{\W_i}^{-k(i)}(X_i)}{\hat{e}^{-k(i)}_{\W_i}(X_i)} \cdot  \W_i
+
\begin{bmatrix}
\hat{\mu}_{a^1}^{-k(i)}(X_i) \\
\hat{\mu}_{a^2}^{-k(i)}(X_i)\\
...\\
\hat{\mu}_{a^d}^{-k(i)}(X_i) 
\end{bmatrix},
\end{equation}
where $k(i)$ denotes the fold that contains the $i$-th data point.
We then find the best candidate policy by selecting from $\Pi$ the policy that maxmizes this approximating
value function.
More specifically,
$\pold_{CAIPWL} = \arg\max_{\pi \in \Pi} \QE_{CAIPWL} (\pi)$.
Combining all of the above steps together, we obtain the CAIPWL algorithm, which is formally summarized in Algorithm~\ref{alg:DRL}. Several definitions then follow.

\begin{algorithm}
	\caption{Cross-fitted Augmented Inverse Propensity Weighted Learning (CAIPWL)} 
	\label{alg:DRL}
	\begin{algorithmic}[1]
		\STATE \textbf{Input:} Dataset $\{(X_i, \W_i, Y_i)\}_{i=1}^n$.
		\STATE Choose $K > 1$.
		\FOR {$k = 1, 2, \dots, K$}
		\STATE Build estimators $\hat{\mu}^{-k}(\cdot) = \begin{bmatrix}
		\hat{\mu}_{a^1}^{-k}(\cdot) \\
		\hat{\mu}_{a^2}^{-k}(\cdot)\\
		...\\
		\hat{\mu}_{a^d}^{-k}(\cdot) 
		\end{bmatrix}, \hat{e}^{-k}(\cdot) =\begin{bmatrix}
		\hat{e}_{a^1}^{-k}(\cdot) \\
		\hat{e}_{a^2}^{-k}(\cdot)\\
		...\\
		\hat{e}_{a^d}^{-k}(\cdot) 
		\end{bmatrix}$
		using the rest $K-1$ folds.
		\ENDFOR
		\STATE Form the approximate value function $\QE_{CAIPWL} (\pi) = \frac{1}{n} \sum_{i=1}^n \langle \pi(X_i),  \frac{Y_i - \hat{\mu}_{\W_i}^{-k(i)}(X_i)}{\hat{e}^{-k(i)}_{\W_i}(X_i)} \cdot  \W_i
		+
		\begin{bmatrix}
		\hat{\mu}_{a^1}^{-k(i)}(X_i) \\
		\hat{\mu}_{a^2}^{-k(i)}(X_i)\\
		...\\
		\hat{\mu}_{a^d}^{-k(i)}(X_i) 
		\end{bmatrix} \rangle$
		\STATE Return $\hat{\pi}_{CAIPWL} = \arg\max_{\pi \in \Pi} \QE_{CAIPWL} (\pi)$
	\end{algorithmic}
\end{algorithm} 

\begin{definition}\quad
	Given the feature domain $\mathcal{X}$, a policy class $\Pi$, a set of $n$ points $\{x_1, \dots, x_n\} \subset \mathcal{X}$, define:
	\begin{enumerate}
		\item
		Hamming distance between any two policies $\pi_1$ and $\pi_2$ in $\Pi$:
			$H(\pi_1, \pi_2) = \frac{1}{n} \sum_{j=1}^n \mathbf{1}_{\{\pi_1(x_j) \neq \pi_2(x_j)\}}$.
		
		\item 
		$\epsilon$-Hamming covering number of the set $\{x_1, \dots, x_n\}$:
		
			$\N_H(\epsilon, \Pi, \{x_1, \dots, x_n\})$ is the smallest number $K$ of policies $\{\pi_1, \dots, \pi_K\}$ in $\Pi$, such that $\forall \pi \in \Pi, \exists \pi_i, H(\pi, \pi_i) \le \epsilon$.
		
		\item
		$\epsilon$-Hamming covering number of $\Pi$:
		$\N_H(\epsilon, \Pi) = \sup\{\N_H(\epsilon, \Pi, \{x_1, \dots, x_m\}) \mid  m \ge 1,  x_1, \dots, x_m \in \mathcal{X}\}.$
		
		\item
		Entropy integral:
		$\kappa(\Pi) = \int_0^1 \sqrt{\log \N_H(\epsilon^2, \Pi)}d\epsilon$.
	\end{enumerate}
\end{definition}

Note that in the above definition, covering number is a classical notion (see~\cite{maurer2009empirical} for a detailed treatment).
Further,
the entropy integral $\kappa(\Pi)$ is a complexity measure of the policy class $\Pi$ and is a variant of the classical entropy integral introduced\footnote{The version given in~\cite{dudley1967sizes} is $\int_0^1 \sqrt{\log \N_H(\epsilon, \Pi)}d\epsilon$} in~\cite{dudley1967sizes}.

\begin{assumption}\label{assump:1}\quad
	$\forall 0 < \epsilon < 1$, $\N_H(\epsilon,\Pi) \le C\exp(D({\frac{1}{\epsilon}})^{\e})$ for some constants $C,D > 0, 0 < \e < 0.5$.
\end{assumption}

\begin{remark}\quad
	Assumption~\ref{assump:1} essentially says that the covering number of the policy class does not grow too quickly. In particular, this implies that the entropy integral is finite: $\kappa(\Pi) = \int_0^1 \sqrt{\log \N_H(\epsilon^2, \Pi)}d\epsilon \le \int_0^1 \sqrt{\log C +  D(\frac{1}{\epsilon})^{2w}}d\epsilon \le 
	\int_0^1 \sqrt{\log C }d\epsilon+ \int_0^1 \sqrt{D(\frac{1}{\epsilon})^{2w}}d\epsilon = \sqrt{\log C } + \sqrt{D}\int_0^1 \epsilon^{-w}d\epsilon = \sqrt{\log C } + \frac{\sqrt{D}}{1-w} < \infty$.
	Note that this is a rather weak assumption: it allows $\log \N_H(\epsilon,\Pi) $ to grow at a low-order polynomial rate as a function of $\frac{1}{\epsilon}$. For the common policy class of finite-depth trees, we establish (see Lemma~\ref{lem:entropy_integral}) that
	$\log \N_H(\epsilon,\Pi) $ is only $O(\log \frac{1}{\epsilon})$, which is order-of-magnitudes less than the required bound here.
\end{remark}

\section{Oracle Regret Bound for CAIPWL}
\label{sec:theory1}

In this section, as an important intermediate step towards establishing the regret bound for CAIPWL,
we establish a regret bound for the policy when oracle access to
the quantities $e_{a}(x), \mu_a(x)$ are available.
Since $e_{a}(x), \mu_a(x)$ are known,
one can pick the policy $\tilde{\pi} \in \Pi$ that optimizes the influence function $\QI (\pi) = \frac{1}{n} \sum_{i=1}^n \langle \pi(X_i), \Gamma_i \rangle$,
where
$\Gamma_i =  \frac{Y_i(A_i) - \mu_{\W_i}(X_i)}{e_{\W_i}(X_i)} \cdot  \W_i
+
\begin{bmatrix}
\mu_{a^1}(X_i) \\
\mu_{a^2}(X_i)\\
...\\
\mu_{a^d}(X_i) 
\end{bmatrix}, $
and $\tilde{\pi} = \arg\max_{\pi \in \Pi} \QI (\pi)$.
The regret bound obtained on $\tilde{\pi}$ is called an oracle regret bound because
we assume (hypothetically) an oracle is able to provide us the exact quantities $e_{a}(x), \mu_a(x)$.

\subsection{Bouding Rademacher Complexity}

A key quantity that allows us to establish uniform concentration results in this setting is the Rademacher complexity of the function class $\PiD \triangleq \{\langle \pi_a(\cdot) -\pi_b(\cdot), \cdot \rangle \mid \pi_a, \pi_b \in \Pi\}$. 
More specifically, each element of this class is a function that takes $(X_i, \Gamma_i)$ as input and outputs $\langle \pi_a(X_i) - \pi_b(X_i), \Gamma_i \rangle$. The superscript $D$ indicates that each function in this class is built out of the difference between two policies $\pi_a$ and $\pi_b$ in $\Pi$.

Our main objective in this subsection is to provide a sharp bound on a type of multi-class Rademacher complexity
defined below. Other notions of multi-class Rademacher complexity have also been defined in the literature, including
separate proposals by \cite{rakhlin2016bistro} and \cite{kallus2017balanced}; however, we find the definition below to
yield sharper regret bounds.

\begin{definition}\quad
Let $Z_i$'s be \textbf{iid} Rademacher random variables: $\P[Z_i = 1] = \P[Z_i = -1] = \frac{1}{2}$.
\begin{enumerate}
	\item The empirical Rademacher complexity $\mathcal{R}_n(\PiD)$ of the function class $\PiD$ is defined as:
	$$\mathcal{R}_n(\PiD; \{X_i, \Gamma_i\}_{i=1}^n) = \E\Bigg[\sup_{\pi_a, \pi_b \in \Pi} \frac{1}{n} \big|\sum_{i=1}^n Z_i \langle \pi_a(X_i) - \pi_b(X_i), \Gamma_i \rangle \big|\Big| \{X_i, \Gamma_i\}_{i=1}^n \Bigg],$$
	where the expectation is taken with respect to $Z_1, \dots, Z_n$.
	\item The Rademacher complexity $\mathcal{R}_n(\PiD)$ of the function class $\PiD$ is the expectated value (taken with respect to the sample $\{X_i, \Gamma_i\}_{i=1}^n$) of the empirical Rademacher complexity:
	$\mathcal{R}_n(\PiD) = \E[\mathcal{R}_n(\PiD; \{X_i, \Gamma_i\}_{i=1}^n)].$
\end{enumerate}
\end{definition}

In characterizing the Rademacher complexity, we introduce a convenient tool that will be useful:

\begin{definition}\quad
	Given a set of $n$ points $\{x_1, \dots, x_n\} \subset \mathcal{X}$, and a set of $n$ weights  $\Gamma = \{\gamma_1, \dots, \gamma_n\} \subset \mathbf{R}^d$, we define the
	inner product distance $\D(\pi_1, \pi_2)$ between two policies $\pi_1$ and $\pi_2$ in $\Pi$ and the corresponding covering number $N_{\D}(\epsilon, \Pi, \{x_1, \dots, x_n\})$ as follows:
	\begin{enumerate}
		\item 
		$\D(\pi_1, \pi_2) =  \sqrt{\frac{\sum_{i=1}^n |\langle \gamma_i,\pi_1(x_i) - \pi_2(x_i)\rangle|^2}{\sup_{\pi_a, \pi_b \in \Pi}\sum_{i=1}^n |\langle \gamma_i,\pi_a(x_i) - \pi_b(x_i)\rangle|^2}},$ where we set $0 \triangleq  \frac{0}{0}$.
		\item
		$N_{\D}(\epsilon, \Pi, \{x_1, \dots, x_n\})$: the minimum number of policies\footnote{Unless otherwise specified, all policies will be assumed to be in $\Pi$.}  needed to $\epsilon$-cover $\Pi$ under $\D$.
	\end{enumerate}

\end{definition}

We characterize two important properties of inner product distance (proof omitted for space limitation):

\begin{lemma}\label{lem:ipd}\quad
	For any $n$, any $\Gamma = \{\gamma_1, \dots, \gamma_n\}$ and any $\{x_1, \dots, x_n\}$:
	\begin{enumerate}
		\item 
		Triangle inequality holds for inner product distance: $\D(\pi_1, \pi_2) \le \D(\pi_1, \pi_3) + \D(\pi_3, \pi_2)$.
		\item
		$N_{\D}(\epsilon, \Pi, \{x_1, \dots, x_n\}) \le N_H(\epsilon^2, \Pi).$
	\end{enumerate}
\end{lemma}

\begin{remark}\quad
	Per the definition of the inner product distance, Lemma~\ref{lem:ipd} and that
	$\D(\pi_1, \pi_2) = \D(\pi_2, \pi_1)$, we see that it is a pseudometric (note that $\D(\pi_1, \pi_2)$ will be $0$ as long as $\pi_1$ and $\pi_2$ agree on the set of points $\{x_1, \dots, x_n\}$ even though the two policies maybe different). However, for convenience, we will just abuse the terminology slightly and call it a distance. 
  Second, the above lemma also relates the covering number under the inner product distance to the covering number under the Hamming distance . We do so because subsequently, it is more convenient to work with the inner product distance, although the entropy integral $\kappa(\Pi)$ is defined in terms of Hamming distance: we hence need a way to convert from the one to the other. 
\end{remark}		

\begin{theorem}\label{thm:rad}\quad
	Let $\{\Gamma_i\}_{i=1}^n$ be \textbf{iid} random vectors with bounded support.
Then under Assumption~\ref{assump:1}:
	\begin{equation}\label{eq:Rad_bound}
	\mathcal{R}_n(\PiD) = 27.2\sqrt{2}(\kappa(\Pi) + 8) \sqrt{\frac{\sup_{\pi_a, \pi_b \in \Pi}\E[\langle\Gamma_i, \pi_a(X_i) - \pi_b(X_i)\rangle^2]}{n}} + o(\frac{1}{\sqrt{n}}).
	\end{equation}
\end{theorem}

{\em Main Steps of the Proof:}
The proof is quite involved and requires several ideas. We break the proof into 4 main components, each discussed at a high level in a single step (see appendix for details).

\setcounter{proofstep}{0}

\begin{proofstep}{Policy approximations}\label{step:2}
 	
 Conditioned on the data $\{X_1, \dots, X_n\}$, and $\Gamma = \{\Gamma_1, \dots, \Gamma_n\}$,
 we define a sequence of refining approximation operators: $A_0, A_1, A_2, \dots, A_J$,
 where $J = \ceil{\log_2(n) (1-\e)}$ and each $A_j: \Pi \rightarrow \Pi$ is an operator that takes a policy $\pi \in \Pi$ and 
 gives the $j$-th approximation policy $A_j(\pi)$ of $\pi$. In other words,
 $A_j(\pi) : \mathcal{X} \rightarrow \mathcal{A}$ is another policy, that serves as
 a proxy to the original policy $\pi$. Through a careful construction, we can obtain a sequence of refining approximation operators, such 
 that the following list of properties holds:
 \begin{enumerate}
 	\item 
 	$\max_{\pi \in \Pi} \D(\pi, A_{J} (\pi)) \le 2^{-J}$.
 		\item
 	$|\{A_j(\pi) \mid \pi \in \Pi \} | \le \N_{\D}(2^{-j},  \Pi, \{X_1, \dots, X_n\})$, for every $j = 0, 1, 2,\dots, J$.
 	\item 
 	$\max_{\pi \in \Pi} \D(A_j(\pi), A_{j+1} (\pi)) \le 2^{-(j-1)}$, for every $j =0, 1, 2, \dots, J-1$.
 	\item For any $J \ge j^{\prime} \ge j \ge 0$,
 	$|\{(A_j(\pi), A_{j^{\prime}}(\pi)) \mid \pi \in \Pi\}| \le \N_{\D}(2^{-j^{\prime}},  \Pi, \{X_1, \dots, X_n\})$.
 \end{enumerate}
Two things to note here are: First, $\{A_0(\pi)\}$ is a singleton set by property $2$ (since the inner product distance between any two policies in $\Pi$ is at most $1$ by definition). In other words, $A_0$ maps any policy in $\Pi$ to a single policy. This means that $A_0$ is the coarsest approximation. At the other end of the spectrum is $A_J$, which gives the finest approximation. 
Second, when $j=0$, property $3$ is obviously true (again because inner product distance is always bounded by $1$) and the upper bound can in fact be refined to $1$. Since 
the impact for Rademacher bound later is only up to some additive constant factor, for notational simplicity, we will just stick with it (instead of breaking it into multiple statements or taking the minimum).
 \end{proofstep}

 \begin{proofstep}{Chaining policies in the negligible regimes}\label{step:3.1}
 	
 For each policy $\pi \in \Pi$, we can write it in terms of the approximation policies (defined above) as:
$ \pi(x) =  A_0(\pi) + \sum_{j=1}^{\jlow} \{A_j(\pi) - A_{j-1} (\pi)\} 
 + \{A_J(\pi) - A_{\jlow}(\pi)\} + \{\pi - A_J(\pi)\},  $
where
$\jlow = \floor{\frac{1}{2} (1-\omega)\log_2(n) }$.
Note that ``chaining" refers to successive finer approximations of policies. 
Consequently, for any two $\pi_a, \pi_b \in \Pi$,
we have:
\begin{equation}\label{eq:policy_approximation_diff}
\begin{split}
\pi_a - \pi_b 
= &\Big\{A_0(\pi_a) + \sum_{j=1}^{\jlow} \{A_j(\pi_a) - A_{j-1} (\pi_a)\} 
+ \{A_J(\pi_a) - A_{\jlow}(\pi_a)\} + \{\pi_a - A_J(\pi_a)\} \Big\}
-\\
&
\Big\{A_0(\pi_b)+ \sum_{j=1}^{\jlow} \{A_j(\pi_b) - A_{j-1} (\pi_b)\} 
+ \{A_J(\pi_b) - A_{\jlow}(\pi_b)\} + \{\pi_b - A_J(\pi_b)\} \Big\} \\
=&
\Big\{\{\pi_a - A_J(\pi_a)\} 
-\{\pi_b - A_J(\pi_b)\}\Big\} + \Big\{\{A_J(\pi_a) - A_{\jlow}(\pi_a)\} -  \{A_J(\pi_b) - A_{\jlow}(\pi_b)\}\Big\} +\\
& \Big\{\sum_{j=1}^{\jlow} \{A_j(\pi_a) - A_{j-1} (\pi_a)\} - \sum_{j=1}^{\jlow} \{A_j(\pi_b) - A_{j-1} (\pi_b)\}\Big\},  \\
\end{split}
\end{equation}
where the second equality follows from that $\{A_0(\pi)\}$ is a singleton set in Step~\ref{step:2}.

We establish two claims in this step (each requiring a different argument), for any $\pi \in \Pi$:

\begin{enumerate}
	\item 
	$\lim_{n \rightarrow \infty} \sqrt{n} \E\Big[\sup_{\pi \in \Pi}\Big|\frac{1}{n} \sum_{i=1}^n Z_i \langle\Gamma_i, \pi(X_i) - A_J(\pi) (X_i)\rangle\Big|\Big] = 0.$
	\item
	$\lim_{n \rightarrow \infty} \sqrt{n} \E\Big[\sup_{\pi \in \Pi} \Big|\frac{1}{n} \sum_{i=1}^n Z_i \langle\Gamma_i, A_J(\pi)(X_i) - A_{\jlow}(\pi) (X_i)\rangle\Big|\Big] = 0.$
\end{enumerate}

These two statements establish that the first two parts of Equation~\eqref{eq:policy_approximation_diff} (after second equality) only contribute 
$o(\frac{1}{\sqrt{n}})$ to the Rademacher complexity, and are hence negligible.
Consequently, only the third part has a non-negligible contribution to the Rademacher complexity, which we characterize next.
\end{proofstep}

\begin{proofstep}{Chaining policies in the effective regime}\label{step:4}
 By expanding the Rademacher complexity using the approximation policies and using the two conclusions established in the previous step, we can show:
\begin{equation}\label{eq:rad_decompose}
\begin{split}
& \mathcal{R}_n(\PiD)
= \E\Big[\sup_{\pi_a, \pi_b \in \Pi}\frac{1}{n} \Big|\sum_{i=1}^n Z_i \langle\Gamma_i, \pi_a(X_i) - \pi_b(X_i)\rangle \Big|\Big] \\
&\le 2\E\Big[\sup_{\pi \in \Pi}\frac{1}{n} \sum_{i=1}^n Z_i \Big|\langle\Gamma_i, \pi(X_i) - A_J(\pi)(X_i)\rangle\Big|\Big] + 2\E\Big[\sup_{\pi \in \Pi}\frac{1}{n} \Big|\sum_{i=1}^n Z_i \langle\Gamma_i, A_J(\pi)(X_i) - A_{\jlow}(\pi)(X_i)\rangle\Big|\Big] \\
& + 2\E\Big[\sup_{\pi \in \Pi}\frac{1}{n} \Big|\sum_{i=1}^n Z_i \langle\Gamma_i, \sum_{j=1}^{\jlow} \Big\{A_j(\pi)(X_i) - A_{j-1}(\pi)(X_i)\Big\}\rangle\Big|\Big] \\
&= 2\E\Big[\sup_{\pi \in \Pi}\frac{1}{n} \Big|\sum_{i=1}^n Z_i \langle\Gamma_i, \sum_{j=1}^{\jlow} \Big\{A_j(\pi)(X_i) - A_{j-1}(\pi)(X_i)\Big\}\rangle\Big|\Big] + o(\frac{1}{\sqrt{n}}) + o(\frac{1}{\sqrt{n}}) \\
&= 2\E\Big[\sup_{\pi \in \Pi}\frac{1}{n} \Big|\sum_{i=1}^n Z_i \langle\Gamma_i, \sum_{j=1}^{\jlow} \Big\{A_j(\pi)(X_i) - A_{j-1}(\pi)(X_i)\Big\}\rangle\Big|\Big] + o(\frac{1}{\sqrt{n}}).
\end{split}
\end{equation}
	
Consequently, it now remains to bound
$\E\Big[\sup_{\pi \in \Pi}\frac{1}{n} \Big|\sum_{i=1}^n Z_i \langle\Gamma_i, \sum_{j=1}^{\jlow} \Big\{A_j(\pi)(X_i) - A_{j-1}(\pi)(X_i)\Big\}\rangle\Big|\Big]$. Indeed, as it turns out, for each $j$ between $1$ and $\jlow$,
$A_j(\pi)(X_i) - A_{j-1}(\pi)(X_i)$ is large enough that it has a non-negligible contribution to the Rademacher complexity.
In this step, we characterize this contribution and establish that:

$\E\Big[ \sup_{\pi \in \Pi}\frac{1}{\sqrt{n}} \Bigg|\sum_{i=1}^n Z_i \Bigg\langle\Gamma_i, \sum_{j=1}^{\jlow} \Big\{A_j(\pi)(X_i) - A_{j-1}(\pi)(X_i)\Big\}\Bigg\rangle\Bigg| \Big] \\
 \le 13.6\sqrt{2}\Big\{\kappa(\Pi) + 8\Big\} \sqrt{\E\Big[\frac{ \sup_{\pi_a, \pi_b}\sum_{i=1}^n |\langle \Gamma_i, \pi_a(X_i) - \pi_b(X_i)\rangle|^2}{n}\Big] }.$
\end{proofstep}

\begin{proofstep}{Refining the lower range bound using Talagrand's inequality}\label{step:5}

Since $ \sup_{\pi_a, \pi_b \in \Pi}\sum_{i=1}^n |\langle \Gamma_i, \pi_a(X_i) - \pi_b(X_i)\rangle|^2 
\le \sum_{i=1}^n \sup_{\pi_a, \pi_b \in \Pi} |\langle \Gamma_i, \pi_a(X_i) - \pi_b(X_i)\rangle|^2 $,
an easy upper bound on $\E\Big[\frac{ \sup_{\pi_a, \pi_b \in \Pi}\sum_{i=1}^n |\langle \Gamma_i, \pi_a(X_i) - \pi_b(X_i)\rangle|^2}{n}\Big]$ is $ \E\Big[ \sup_{\pi_a, \pi_b \in \Pi} |\langle \Gamma_i, \pi_a(X_i) - \pi_b(X_i)\rangle|^2\Big]$. However, this upper bound is somewhat loose because the maximization over policies happens after the data $\{X_i, \Gamma_i\}$ is seen, as opposed to before.
By using a sharpend version of Talagrand's concentration inequality in~\cite{gine2006}, we can obtain 
a more refined bound on $\E\Big[\frac{ \sup_{\pi_a, \pi_b \in \Pi}\sum_{i=1}^n |\langle \Gamma_i, \pi_a(X_i) - \pi_b(X_i)\rangle|^2}{n}\Big]$ in terms of the tighter population quantity $\sup_{\pi_a, \pi_b \in \Pi} \E \Big[|\langle \Gamma_i, \pi_a(X_i) - \pi_b(X_i)\rangle|^2 \Big]$.
Specifically, we show that the following inequality holds:
\begin{equation}\label{eq:lower3}
\E\Big[\frac{ \sup_{\pi_a, \pi_b \in \Pi}\sum_{i=1}^n |\langle \Gamma_i, \pi_a(X_i) - \pi_b(X_i)\rangle|^2}{n}\Big]
\le   \sup_{\pi_a, \pi_b \in \Pi} \E \Big[\langle \Gamma_i, \pi_a(X_i) - \pi_b(X_i)\rangle^2 \Big] + 8U \mathcal{R}_n(\PiD).
\end{equation}
Combining this equation with Step~\ref{step:4} and then further with Step~\ref{step:3.1}, we obtain the final bound.
\end{proofstep}
$\hfill\blacksquare$

\begin{remark}\label{rem:V_def}\quad
	Two remarks are in order here.
	First, we denote for the rest of the paper $V_* \triangleq \sup_{\pi_a, \pi_b \in \Pi}\E[\langle\Gamma_i, \pi_a(X_i) - \pi_b(X_i)\rangle^2]$, which is an important population quantity in the bound on Rademacher complexity (Equation~\ref{eq:Rad_bound}). 
   $V_*$ can be interpreted as the worst-case variance of evaluating the difference 
	between two policies in $\Pi$: it measures how variable it is to evaluate the difference between
	$\langle\Gamma_i, \pi_a(X_i) \rangle$ (the policy value when using $\pi_a$) and $\langle\Gamma_i, \pi_b(X_i)\rangle$ (the policy value when using $\pi_b$) in the worst-case. Intuitively, the more difficult it is to evaluate this worst-case difference (i.e. the larger the $V_*$), the harder it is to distinguish between two policies, and hence the larger the Rademacher complexity it is (all else fixed).
	
	Second, a further upper bound on $V_*$ is $2\E[\|\Gamma_i\|_2^2]$, because
	by Cauchy-Schwartz, we have:
	\begin{equation}
	\begin{split}
	&\sup_{\pi_a, \pi_b \in \Pi}\E\Big[\langle\Gamma_i, \pi_a(X_i) - \pi_b(X_i)\rangle^2 \Big]\le
	\E\Big[\sup_{\pi_a, \pi_b \in \Pi}\langle\Gamma_i, \pi_a(X_i) - \pi_b(X_i)\rangle^2 \Big]\\
	&\le \E\Big[\sup_{\pi_a, \pi_b \in \Pi}\|\Gamma_i\|_2^2 \|\pi_a(X_i) - \pi_b(X_i)\|_2^2\Big] \le 2\E[\|\Gamma_i\|_2^2].
	\end{split}
	\end{equation}
	However, note that $2\E[\|\Gamma_i\|_2^2]$ can be much larger than $V_*$ because $\langle\Gamma_i, \pi_a(X_i) - \pi_b(X_i)\rangle$ only picks out the components of $\Gamma_i$
	at which $\pi_a(X_i)$ and $\pi_b(X_i)$ differ (where all the other componets are zeroed out).
	This in fact illustrates the sharpness of our bound (even in terms of problem-specific constants): even though  $\E[\|\Gamma_i\|_2^2]$ may be large (representing the fact that the counterfactual estimates $\Gamma_i$ are highly volatile), the final Rademacher bound only depends on the magnitude of the difference induced by two policies.
\end{remark}
\subsection{Uniform Concentration of Influence Difference Functions}

To establish high-probability regret bounds for $\tilde{\pi}$, we now characterize 
the uniform concentration of the influence function on the policy value function. 
However, unlike uniform concentration bounds in most other learning-theoretical contexts,
the influence function does not directly uniformly concentrate on the policy value function (at least not 
at a sharp enough rate). Consequently, we consider a novel concentration of a different set of quantities: that of the differences. More specifically,
we will concentrate the difference of influence functions as opposed to the influence function itself.
We start with a definition that makes this point precise.

\begin{definition}\quad
The influence difference function $\tilde{\Delta}(\cdot, \cdot): \Pi \times \Pi \rightarrow \reals$ and the policy value difference function $\Delta(\cdot, \cdot): \Pi \times \Pi \rightarrow \reals$ are defined respectively as follows: 	
$\tilde{\Delta}(\pi_1, \pi_2) = \QI(\pi_1) - \QI(\pi_2), \Delta(\pi_1, \pi_2) = Q(\pi_1) - Q(\pi_2).$
\end{definition}

\begin{lemma}\label{thm:uniform_counter}\quad
Under Assumptions~\ref{assump:classical} and~\ref{assump:1},  the influence difference function concentrates uniformly around its mean: for any $\delta > 0$, with probability at least $1 - 2\delta$,
\begin{equation}
\begin{split}
\sup_{\pi_1, \pi_2 \in \Pi} \Big|\tilde{\Delta}(\pi_1, \pi_2) - \Delta(\pi_1, \pi_2)\Big|\le  \Big(54.4\sqrt{2}\kappa(\Pi) + 435.2 + \sqrt{2\log\frac{1}{\delta}}\Big) \sqrt{\frac{V_*}{n}} + o(\frac{1}{\sqrt{n}}).
\end{split}
\end{equation}
\end{lemma}

{\em Main Steps of the Proof:}
The proof divides into two main components (all the proof details omitted). 
\setcounter{proofstep}{0}
\begin{proofstep}{Expected uniform bound on maximum deviation}\label{step:1}
From Theorem~\ref{thm:rad}, we have:
\begin{equation}\label{eq:joint_rad}
\begin{split}
& \mathcal{R}_n(\PiD) \le  27.2\sqrt{2}(\kappa(\Pi) + 8) \sqrt{\frac{\sup_{\pi_a, \pi_b \in \Pi}\E[\langle\Gamma_i, \pi_a(X_i) - \pi_b(X_i)\rangle^2]}{n}} + o(\frac{1}{\sqrt{n}}).
\end{split}
\end{equation}
By an explicit computation, we show that $\E[\tilde{\Delta}(\pi_1, \pi_2)] = \Delta(\pi_1, \pi_2)$.
Using this conclusion and Equation~\eqref{eq:joint_rad}, as well as properties of Rademacher complexity, we then establish (after some involved algebra):
\begin{equation}\label{eq:bound1}
\E\Big[\sup_{\pi_1, \pi_2 \in \Pi} \Big|\tilde{\Delta}(\pi_1, \pi_2) - \Delta(\pi_1, \pi_2)\Big| \Big] \le
 54.4\sqrt{2}(\kappa(\Pi) + 8) \sqrt{\frac{\sup_{\pi_a, \pi_b \in \Pi}\E[\langle\Gamma_i, \pi_a(X_i) - \pi_b(X_i)\rangle^2]}{n}} + o(\frac{1}{\sqrt{n}}).
\end{equation}	
This result indicates that at least in expectation, the maximum deviation of the influence difference function from its mean is well-controlled.
\end{proofstep}

\begin{proofstep}{High probability bound on maximum deviation via Talagrand inequality}\label{step:2}

From the previous step, it remains to bound the difference between $\sup_{\pi \in \Pi} \Big|\tilde{\Delta}(\pi_1, \pi_2) - \Delta(\pi_1, \pi_2)\Big|$ and $\E\Big[\sup_{\pi \in \Pi} \Big|\tilde{\Delta}(\pi_1, \pi_2) - \Delta(\pi_1, \pi_2)\Big| \Big]$.
We establish a high-probability bound on this difference by applying the Bennett concentration inequality in~\cite{bousquet2002bennett}, where we 
 identify each $f_i$ with $\frac{\langle \Gamma_i , \pi_1(X_i) - \pi_2(X_i) \rangle - \E[\langle \Gamma_i , \pi_1(X_i) - \pi_2(X_i)]}{2U}$ and
consider:
$\frac{n}{2U}\Big|\tilde{\Delta}(\pi_1, \pi_2) - \Delta(\pi_1, \pi_2)\Big| = \Big|\sum_{i=1}^n \frac{\langle \Gamma_i , \pi_1(X_i) - \pi_2(X_i) \rangle - \E[\langle \Gamma_i , \pi_1(X_i) - \pi_2(X_i)]}{2U} \Big|,$
where $U$ is an upper bound on $\langle \Gamma_i , \pi_1(X_i) - \pi_2(X_i) \rangle$. 
We then establish that with probability at least $1 -2\delta$:
\begin{equation}\label{eq:bound2}
\begin{split}
&\sup_{\pi_1, \pi_2 \in \Pi} \Big|\tilde{\Delta}(\pi_1, \pi_2) - \Delta(\pi_1, \pi_2)\Big|\le   \E\Big\{\sup_{\pi_1, \pi_2 \in \Pi} \Big|\tilde{\Delta}(\pi_1, \pi_2) - \Delta(\pi_1, \pi_2)\Big| \Big\}+ \sqrt{2\log\frac{1}{\delta}}\sqrt{\frac{V_*}{n}} + O(\frac{1}{n^{0.75}}).
\end{split}
\end{equation}

Finally, combining Equation~\eqref{eq:bound1} and Equation~\eqref{eq:bound2} yields that with probability at least $1 - 2\delta$,
\begin{equation}\label{eq:bound3}
\begin{split}
&\sup_{\pi_1, \pi_2 \in \Pi} \Big|\tilde{\Delta}(\pi_1, \pi_2) - \Delta(\pi_1, \pi_2)\Big|\le 54.4\sqrt{2}(\kappa(\Pi) + 8) \sqrt{\frac{V_*}{n}} + o(\frac{1}{\sqrt{n}})  + \sqrt{2\log\frac{1}{\delta}}\sqrt{\frac{V_*}{n}} + O(\frac{1}{n^{0.75}})\\
& = \Big(54.4\sqrt{2}\kappa(\Pi) + 435.2 + \sqrt{2\log\frac{1}{\delta}}\Big) \sqrt{\frac{\sup_{\pi_a, \pi_b \in \Pi}\E[\langle\Gamma_i, \pi_a(X_i) - \pi_b(X_i)\rangle^2]}{n}} + o(\frac{1}{\sqrt{n}}).
\end{split}
\end{equation}
\end{proofstep}
$\hfill\blacksquare$

The uniform concentration established in Lemma~\ref{thm:uniform_counter} then allows us to obtain tight regret bounds on the policy $\tilde{\pi}$ learned via counterfactual risk minimization, as given next.

\begin{theorem}\label{thm:main_counter}\quad
Let $\polc \in \arg\min_{\pi \in \Pi}\QI(\pol)$. Then for any $\delta > 0$, with probability at least $1 - 2\delta$,
$$R(\polc) \le \Big(54.4\sqrt{2}\kappa(\Pi) + 435.2 + \sqrt{2\log\frac{1}{\delta}}\Big) \sqrt{\frac{V_*}{n}} + o(\frac{1}{\sqrt{n}}).$$
\end{theorem}

{\em Proof:} The result follows by applying Theorem~\ref{thm:uniform_counter} and noting that with probability at least $1 - 2\delta$.:
\begin{equation}
\begin{split}
&R(\polc) = Q(\polo) - Q(\polc) = \QI(\polo) - \QI(\polc) + \diff(\polo, \polc) - \diffi(\polo, \polc)
\le \diff(\polo, \polc) - \diffi(\polo, \polc)\\
& \le | \diffi(\polo, \polc) - \diff(\polo, \polc)| \le \sup_{\pi_1,\pi_2 \in \Pi} |\diffi(\pi_1, \pi_2) - \diff(\pi_1, \pi_2)| \\
& \le \Big(54.4\sqrt{2}\kappa(\Pi) + 435.2 + \sqrt{2\log\frac{1}{\delta}}\Big) \sqrt{\frac{\sup_{\pi_a, \pi_b \in \Pi}\E[\langle\Gamma_i, \pi_a(X_i) - \pi_b(X_i)\rangle^2]}{n}} + o(\frac{1}{\sqrt{n}}),
\end{split}
\end{equation}
where the first inequality follows from the definition of $\polc$. 
$\hfill\blacksquare$

\section{Regret Bound for (Feasible) CAIPWL}\label{sec:theory2}

In this section, we establish the main regret bound for the proposed algorithm CAIPWL.
In the previous section, we established uniform concentration between influence difference functions
and the policy value difference functions. 
Since the influence functions assume the knowledge of the actual $\mu(\cdot)$ and $e(\cdot)$,
the oracle regret bound is still some distance away from our desideratum. In this section, we fill in the gap.
Specifically, we show that uniform concentration on differences also holds when one uses the corresponding estimates $\hat{\mu}(\cdot)$ and $\hat{e}(\cdot)$ instead.

\subsection{Uniform Difference Concentration with Actual Estimates}

Our strategy here is again to concentrate the differences of the quantities rather than the quantities themselves. We start with definitions.

\begin{definition}\quad
	Define $\diffd(\cdot, \cdot): \Pi \times \Pi \rightarrow \reals$ as follows:
	$\diffd(\pi_1, \pi_2) = \QE_{CAIPWL}(\pi_1) - \QE_{CAIPWL}(\pi_2).$
\end{definition}

It turns out that , as made formal by Lemma~\ref{lem:uniform_counter},  with high probability,
$\diffd(\cdot, \cdot)$ concentrates around $\diffi(\cdot, \cdot)$
uniformly at a rate strictly faster than $O(\frac{1}{\sqrt{n}})$.
Before stating the result, we first recall the commonly used notation $O_p(\cdot)$ and $o_p(\cdot)$ (which will make the statement of the result less cluttered with cumbersome symbols).
Let $\{X_n\}_{n=1}^{\infty}$ be a given sequence of random variables
and $\{a_n\}_{n=1}^{\infty}$ be a given sequence. Then,
\begin{enumerate}
	\item
	$X_n = O_p(a_n)$ if for any $\epsilon > 0$,  with probability at least $1 - \epsilon$, $|\frac{X_n}{a_n}| \le M, \forall n \ge N,$ for some $M > 0$ and $N > 0$, 
	\item 
	$X_n = o_p(a_n)$ if for any $\epsilon > 0$,  $\lim_{n\rightarrow \infty} P(|\frac{X_n}{a_n}| \ge \epsilon) = 0.$
\end{enumerate}

\begin{lemma}\label{lem:uniform_counter}\quad
	Assume $\kappa(\Pi)  < \infty$ and $K \ge 2$. Under
	Assumptions~\ref{assump:classical} and~\ref{assump:consistency}, 
	$$\sup_{\pi_1, \pi_2 \in \Pi} \Big|\diffd(\pi_1, \pi_2) - \diffi(\pi_1, \pi_2)\Big| = o_p(\frac{1}{\sqrt{n}}).$$
\end{lemma}

\begin{remark}\quad
	The preceding result, equivalently $\sqrt{n}\sup_{\pi_1, \pi_2 \in \Pi} \Big|\diffd(\pi_1, \pi_2) - \diffi(\pi_1, \pi_2)\Big| =  o_p(1)$, is much stronger than those in the classical estimation theory, which establish that $\sqrt{n} \Big|\diffd(\pi_1, \pi_2) - \diffi(\pi_1, \pi_2)\Big| =  o_p(1)$: for a single pair of two policies, $\sqrt{n} \Big|\diffd(\pi_1, \pi_2) - \diffi(\pi_1, \pi_2)\Big| \rightarrow  0$ in probability as $n \rightarrow \infty$. In other words, it's a single pair concentration because convergence only holds when $\pi_1$ and $\pi_2$ are fixed; hence convergence can \textit{a priori} fail easily when one scans through over all possible pair of policies in $\Pi$. 
	Such results are far from sufficient in the current context.
	We have thus generalized the single-pair concentration result to 
	the much stronger uniform convergence result to meet the demanding requirements of the policy learning setting. Finally, we note that a key feature that has enabled this result is cross-fitting, where the training data is divided into different folds and the estimation on a given data point is performed using other data folds. Cross-fitting is a commonly employed technique in statistical estimation~\cite{chernozhukov2016double} to reduce overfit. Here, we show that the efficiency gains in estimation in fact translates to policy learning.
\end{remark}

{\em Main Outline of the Proof:}
Take any two policies $\pi^a, \pi^b \in \Pi$ (here we use superscripts $a,b$ because we will also use subscripts $\pi_j$ to access the $j$-th component of a policy $\pi$).
By some tedious algebra, one can show that
$\diffd(\pi^a, \pi^b) - \diffi(\pi^a, \pi^b) = \sum_{j=1}^d(\diffd_{DR}^{j} (\pi^a, \pi^b) - \diffi^{j} (\pi^a, \pi^b))$,
where we have the following decomposition:
\begin{equation}
\begin{split}
\diffd_{DR}^{j} (\pi^a, \pi^b) - \diffi^{j} (\pi^a, \pi^b) 
& =  \frac{1}{n}\sum_{i=1}^n \Big(\pi_j^a(X_i) - \pi_j^b(X_i)\Big) \Big( 
\hat{\mu}_{a^j}^{-k(i)}(X_i) - \mu_{a^j}(X_i) \Big)\Big(1 - \frac{\mathbf{1}_{\{\W_i = a^j\}}}{e_{a^j}(X_i)} \Big)\\
& +
\frac{1}{n}\sum_{\{i \mid \W_i =a^j\}} \Big(\pi_j^a(X_i) - \pi_j^b(X_i)\Big) \Big( 
Y_i - \mu_{a^j}(X_i) \Big)\Big(\frac{1}{\hat{e}_{a^j}^{-k(i)}(X_i)} - \frac{1}{e_{a^j}(X_i)} \Big)\\
&+ 
\frac{1}{n}\sum_{\{i \mid \W_i =a^j\}} \Big(\pi_j^a(X_i) - \pi_j^b(X_i)\Big) \Big( 
\mu_{a^j}(X_i)  - \hat{\mu}_{a^j}^{-k(i)}(X_i)\Big)\Big(\frac{1}{\hat{e}_{a^j}^{-k(i)}(X_i)} - \frac{1}{e_{a^j}(X_i)} \Big).
\end{split}
\end{equation} 
By a careful analysis, we establish that each of the three components is $o_p(\frac{1}{\sqrt{n}})$ (and hence the claim). The steps are quite involved and all the details are presented in the appendix.
On an intuitive level, this can be best understood by recognizing that each of the three terms is a product of two small terms. The first term $\Big( 
\hat{\mu}_{a^j}^{-k(i)}(X_i) - \mu_{a^j}(X_i) \Big)\Big(1 - \frac{\mathbf{1}_{\{\W_i = a^j\}}}{e_{a^j}(X_i)} \Big)$ is a product of estimation error and noise, the second term $\Big( 
Y_i - \mu_{a^j}(X_i) \Big)\Big(\frac{1}{\hat{e}_{a^j}^{-k(i)}(X_i)} - \frac{1}{e_{a^j}(X_i)} \Big)$ is a product of noise and estimation error and the final term $\Big( 
\mu_{a^j}(X_i)  - \hat{\mu}_{a^j}^{-k(i)}(X_i)\Big)\Big(\frac{1}{\hat{e}_{a^j}^{-k(i)}(X_i)} - \frac{1}{e_{a^j}(X_i)} \Big)$ is a product of estimation error and estimation error. 
A careful analysis utilizing the above observations then establishes the result (all details omitted).

\subsection{Main Regret Bound}
Putting all the results together yields the final regret bound:

\begin{theorem}\label{thm:main}\quad
	Let $\pold_{CAIPWL}$ be the policy learned from Algorithm~\ref{alg:DRL} and let  Assumptions~\ref{assump:classical},\ref{assump:consistency},\ref{assump:1} hold. Then, for any $\delta > 0$, there exists an integer $N_{\delta}$, such that with probability at least $1 -2\delta$ and for all $n \ge N_{\delta}$.:
	$$R(\pold_{CAIPWL}) \le \Big(54.4\sqrt{2}\kappa(\Pi) + 435.2 + \sqrt{2\log\frac{1}{\delta}}\Big) \sqrt{\frac{V_*}{n}},$$
\end{theorem}

\begin{remark}\quad
	The constants in the above theorem can be further tightened.
	The proof is a consequence of Lemma~\ref{thm:uniform_counter} and Lemma~\ref{lem:uniform_counter} with some additional algebra, which we omit here.
	Note that a simpler way of stating the regret bound in Theorem~\ref{thm:main} is
	$R(\pold_{CAIPWL}) =O_p\Big(\kappa(\Pi) \sqrt{\frac{V_*}{n}}\Big)$, which can be quickly seen below:
\begin{equation}
\begin{split}
& R(\pold_{CAIPWL}) = Q(\polo) - Q(\pold_{CAIPWL}) = \QE(\polo) - \QE(\pold_{CAIPWL}) + \diff(\polo, \pold_{CAIPWL}) - \diffd(\polo, \pold_{CAIPWL}) \\
&\le \diff(\polo, \pold_{CAIPWL}) - \diffd(\polo, \pold_{CAIPWL})\\
& \le \Big| \diff(\polo, \pold_{CAIPWL}) - \diffd(\polo, \pold_{CAIPWL}) \Big|\le \sup_{\pi_1, \pi_2 \in \Pi} \Big|\diffd(\pi_1, \pi_2) - \Delta(\pi_1, \pi_2) \Big| \le
 \sup_{\pi_1, \pi_2 \in \Pi} \Big|\diffd(\pi_1, \pi_2) - \diffi(\pi_1, \pi_2)\Big| \\
&+
\sup_{\pi_1, \pi_2 \in \Pi} \Big|\tilde{\Delta}(\pi_1, \pi_2) - \Delta(\pi_1, \pi_2) \Big| 
= O_p\Big((\kappa(\Pi) + 1) \sqrt{\frac{V_*}{n}}\Big) +o_p(\frac{1}{\sqrt{n}}) = O_p\Big((\kappa(\Pi)) \sqrt{\frac{V_*}{n}}\Big).
\end{split}
\end{equation}
\end{remark}

Note that the last equality follows because, except in a fully degenerate case (i.e. $\Pi$ contains only 1 policy), $\kappa(\Pi) = \Omega(1)$. Further,
in the special case of binary action set (i.e. $|\actions| = 2$), one can consider the well-known VC class (\cite{vapnik1971uniform}): $\Pi$ is said to be in VC class, if its VC dimension $VC(\Pi)$ is finite (see appendix for a review of its definition). 
Note that the concept of VC dimension, introduced in~\cite{vapnik1971uniform}, is a widely used notion in machine learning that measures how well a class of functions can fit a set of points with binary labels.
	VC dimension only makes sense  if $|\actions| = 2$.
Here we have:

\begin{corollary}\label{cor:VC}\quad
	Under Assumptions~\ref{assump:classical},\ref{assump:consistency},\ref{assump:1} and assume  $|\actions| = 2$,  $R(\pold_{CAIPWL}) =O_p\Big(\sqrt{\frac{V_* \cdot \text{VC}(\Pi)}{n}}\Big)$.
\end{corollary}

{\em Proof:}
By Theorem 1 in~\cite{haussler1995sphere},
the covering number can be bounded by VC dimension as follows: $\N_H(\epsilon, \Pi) \le  e(VC(\Pi) +1)(\frac{2e}{\epsilon})^{VC(\Pi)}$. From here, by taking the natural log of both sides and computing the entropy integral, one can show that $\kappa(\Pi) \le 2.5\sqrt{VC(\Pi)}$.
\hfill$\blacksquare$

\subsection{Comparisons}\label{subsec:comparisons}
We now provide a discussion on the comparisons between our main regret bound $R(\pold_{CAIPWL}) =O_p\Big((\kappa(\Pi) \sqrt{\frac{V_*}{n}}\Big)$ and the existing regret bounds in the literature.
First, we give a quick comparison of our bound $O_p\Big(\sqrt{\frac{V_* \cdot \text{VC}(\Pi)}{n}}\Big)$ in Corollary~\ref{cor:VC} to the existing bounds
in the special case of binary-action policy learning. Note that we will only compare constants in the regret bound if the dependence on $n$ matches ours.
In~\cite{zhao2014doubly}, a different doubly robust learning algorithm is proposed where the regret bound is $O_p(\frac{1}{n^{\frac{1}{2+1/q}}})$, where $0 < q < \infty$ is a parameter that measures how
well the two actions are separated (the larger the $q$, the better the separation of the two actions).
Note that our bound is strictly better and is indepedent of the extra separation parameter. 
In~\cite{kitagawa2015should}, a particular inverse propensity weighted learning algorithm was proposed, which has regret bound $O_p\Big(\frac{B}{\eta}\sqrt{\frac{ \text{VC}(\Pi)}{n}}\Big)$, provided the propensities are known, where $B$ is the upper bound on the support of $|Y_i|'s$ (that is $|Y_i| \le B$ almost surely for every $i$) and $\eta$ is the uniform lower bound in sampling an action for any given $x$ (Assumption~\ref{assump:classical}). While the bound's dependence on $n$ is optimal and the same as ours, it is looser in the other parameters of the problem: the aboluste support parameters represent the worst-case bound compared to the second moment, and in most cases, $\E[\| \Gamma_i\|_2^2 ] << \frac{B^2}{\eta^2}$ (recall that $V_* =\sup_{\pi_a, \pi_b \in \Pi}\E[\langle\Gamma_i, \pi_a(X_i) - \pi_b(X_i)\rangle^2] \le \E[\| \Gamma_i\|_2^2 ]$.) 
The recent paper \cite{athey2017efficient} proposed another binary-action doubly robust learning algorithm and further tightened the constant to $\E[\| \Gamma_i\|_2^2 ]$ (also with the optimal $\frac{1}{\sqrt{n}}$ dependence), which essentially matches our bound here.
 Consequently, our regret bound provides the best guarantee even in the highly special binary-action setting.
Finally, we mention that all of the existing learning algorithms in the binary-action context relies either 
on comparing one action to the other (a.k.a. treatment effect) or on the special embedding of two actions (e.g. $1$ and $0$ or $1$ and $-1$) or both; hence, it is not at all clear how those algorithms can be generalized to multiple-action contexts.

We next discuss comparisons in multi-action settings. We start by discussing the existing work in mutli-action policy learning that requires known propensity.
First, ~\cite{swaminathan2015batch} proposed an inverse propensity weights based policy learning algorithm. The resulting regret bound is $O_p\Big(\sqrt{\frac{V \log \N_H(\frac{1}{n}, \Pi) }{n}}\Big)$.  While the dependence on the second moment $V$ is desirable compared to support parameters, the dependence on the covering number is not entirely satisfactory.
To put this regret bound in perspective, 
first consider the binary-action setting. 
In this case, by lower bounding the covering number using VC dimension (Theorem 2 in~\cite{haussler1995sphere}),
we have $ \log \N_H(\frac{1}{n}, \Pi) \ge O(VC(\Pi) \log n)$. Consequently, even in the binary-action case, the regret bound is at least $O_p\Big(\sqrt{\frac{V_*    VC(\Pi) \log n  }{n}}\Big)$.
which is non-optimal.
As an extreme case to further illustrate this point:
if $\N_H(\epsilon, \Pi) = \exp((\frac{1}{\epsilon})^{\omega}), \omega < 0.5$, a requirement satisfied by Assumption~\ref{assump:1}, then $\log \N_H(\frac{1}{n}, \Pi) =  n^{\omega}$ and hence the regret bound 
in~\cite{swaminathan2015batch} translates to $O_p\Big(\sqrt{\frac{V }{n^{1-\omega}}}\Big)$, and can be as large as $O_p(\frac{1}{n^{0.25}})$,
which is much worse than $O_p\Big((\kappa(\Pi) \sqrt{\frac{V_*}{n}}\Big)$ given in Theorem~\ref{thm:main}, where we note that $\kappa(\Pi)$ is a fixed constant. 
Finally, as a minor point, since the analysis in~\cite{swaminathan2015batch} does not look at the policy difference $\pi_a(X_i) - \pi_b(X_i)$,  the $V$ term there corresponds to $\E[\|\Gamma_i\|_2^2]$ and is hence larger than $V_*$ here, although we believe a similar analysis to ours would tighten $V$ to $V_*$.
In~\cite{zhou2015residual}, another inverse propensity weighs based learning algorithm (called residual weighted learning) was proposed, where a regret bound of $O_p(n^{-\frac{\beta}{2\beta +1}})$, for some problem specific parameter $0 < \beta \le 1$. 
$\beta$ is a parameter that measures the growth rate of the approximation error function (the difference between the optimal policy value and some regularized policy value used to select a policy) and is restricted to lie between $0$ and $1$. 
Hence, even the best possible regret bound (where $\beta = 1$ and the approximation error function is the smallest) is $O_p(n^{-\frac{1}{3}})$, which is again non-optimal. While this regret bound can be  better (or worse) than the one in~\cite{swaminathan2015batch} (depending on certain problem-specific quantities), neither is optimal.
In addition, both algorithms require the propensities to be known; hence, the resulting regret bounds are 
oracle regret bounds only.

We point out that policy learning algorithms assuming known propensities are of practical utility if the policy used to collect the data is known beforehand, since the propensities can then be recovered. This occurs if, for instance, the data collector is the same as the decision maker, who designed the experiments to gather the dataset. 
On the other hand, in many practical settings, they are two separate entities and the data-collecting policy may not be explicitly encoded, in which case acquiring propensities becomes unrealistic. In such cases, it is important to work with feasible algorithms where propensities must be estimated from data.
Althought it may initially appear surprising that the known propensities assumption can be dispensed with,
fundamentally, it is because we are estimating two ``orthogonal" components at the same time: reward models and propensities. This approach in fact falls into the broader orthogonal moments paradigm (\cite{newey1994asymptotic, robins1995semiparametric, belloni2014inference}), which is a line of research in statistical estimation that advocates the insight that using two (or more) complementary components in estimation can enhance the overall accuracy, and the various instantiations of which have recently been fruitfully applied (\cite{ chernozhukov2016locally}, \cite{belloni2017program}) to policy evaluation. 
In presenting our results, we first establish $O_p(\frac{1}{\sqrt{n}})$ regret bound under known propensities (also known as oracle regret bound) in Section~\ref{sec:theory1} (Theorem~\ref{thm:main_counter}) and then analyze the
$O_p(\frac{1}{\sqrt{n}})$ final regret bound in Section~\ref{sec:theory2} (Theorem~\ref{thm:main}).
That these two bounds have the same asymptotic dependence on $n$ should not be taken for granted 
and is a consequence of both the designed algorithm ($K$-fold structure as well as the underlying AIPW estimator) and a sharp analysis (Lemma~\ref{lem:uniform_counter}).
In particular, it is unlikely that the same regret bounds in~\cite{swaminathan2015batch,zhou2015residual} would still hold when the propensities are not known and must be estimated. 

Finally, we compare our regret bound to that of~\cite{kallus2017balanced}, where a balanced policy learning approach was proposed that, like us, dispenses with the assumption of known propensity.
Several regret bounds are given under different assumptions. For instance, as a bound that is directly comparable to ours, \cite{kallus2017balanced} established that if 
$\E\Bigg[ \Big| \hat{\mu}_{a}^{(n)}(X) - \mu_{a}(X) \Big| \Bigg] = r(n)$, then the regret bound 
is $O_p(r(n) + \mathcal{R}_n(\Pi)+ \frac{1}{n})$. It suffices to point out that, unless $\mu_a(\cdot)$ is realizable by a parametric family, $r(n)$ will be strictly worse than $O(\frac{1}{\sqrt{n}})$, in which case $r(n)$ dominates the regret bound and yields a suboptimal learning rate. In particular, if we can estimate all nuisance components at a fourth-root rate in root-mean squared error, then
the regret bound of \cite{kallus2017balanced} scales as $o_P(\frac{1}{n^{0.25}})$, whereas our regret bound scales as $O_p(\frac{1}{n^{0.5}})$.
Furthermore, our regret bound enjoys the additional benefit of being flexible in trading off the accuracy of estimating the model with that of the propensity, a property missing in~\cite{kallus2017balanced} that can make its regret bound even more disadvantageous. \cite{kallus2017balanced} also presents another stronger regret bound scaling as $O_p(\mathcal{R}_n(\Pi)+ \frac{1}{n})$ for the special case where the Bayes optimal policy lies in an RKHS.
However, such RKHS-realizability is a rather strong assumption (and we do not make such an assumption here).
Moreover, the notion of Rademacher complexity used in~\cite{kallus2017balanced} is rather loose (to be precise, a factor of $d$ loss where recall $d$ is the number actions). One consequence of this is that with trees, even 
in the special case where realizability holds, it results in $O_p(\frac{d^2}{\sqrt{n}})$ regret rather than our $O_p(\frac{\sqrt{\log d}}{\sqrt{n}})$ regret bound, even though both are optimal with respect to $n$. 

\section{Exact Policy Learning with Decision Trees}

In this section, we consider policy learning with a specific policy class: decision trees.
To this end, we start by recalling the definition of decision trees (hereafter referred to as trees for short)
in Section~\ref{subsec:trees_def} and characterizing the complexity of this policy class by bounding its entropy integral.
This complexity characterization then directly allows us to translate the regret bound established in Theorem~\ref{thm:main} for a general, abstract policy class to an application-specific regret bound
in the context of trees.

Next, we show that with trees being the policy class, finding the optimal $\hat{\pi}_{CAIPWL}$ that \textit{exactly} maximizes $\QE_{CAIPWL}$, a key step in Algorithm~\ref{alg:DRL}, can be formulated as a mixed integer program (MIP). We describe the MIP formulation in Section~\ref{subsec:mip}.
Note that since finding an optimal classification tree is generally intractable (\cite{bertsimas2017optimal}), and since classification can be reduced to a special case of policy learning\footnote{This also makes it clear that the MIP we develop here contains as a special case the MIP developed for simple classification purposes~\cite{bertsimas2017optimal}. Although strictly speaking, since \cite{bertsimas2017optimal} also considers several other variants that are helpful for the classification context, it is not quite a strict special case. } (\cite{zhang2012estimating,zhao2012estimating}), policy learning with trees in general is intractable as well. Consequently, the existing implementations (whether binary-action or not)
of various policy learning algorithms using trees (\cite{langford2011doubly, zhang2012estimating,zhao2012estimating}) are rather ad hoc, and often 
greedy-based algorithms.
In particular, this means that the existing approaches for offline multi-action policy learning are neither statistically optimal (regret guarantees) nor computationally optimal (maximizing policy value estimator).
On the other hand, our MIP approach finds an exact optimal policy in the optimization subroutine of CAIPWL; when combined with the above-mentioned regret guarantees, provides a learning algorithm that is both statistically and computationally optimal.

Finally, in Section~\ref{subsec:tree_search}, we discuss the merits and drawbacks of the MIP approach, the latter of which lead to a customized tree-search algorithm that we design and implement, also for the purspose of exact tree-based policy learning. In comparison to the MIP approach, the main advantage of the customized tree-search algorithm is its computational efficiency. In particular, while the number of MIP variables scales linearly with the number of data points (holding other problem parameters fixed), the total running time for solving a MIP scales exponentially with the number of variables. This has two drawbacks: 1) it limits the size of training data that can be solved to optimality, as already mentioned; 2) it is difficult to estimate the amount of time it takes to solve a given policy learning problem due to the blackbox nature of MIP solvers. The first drawback presents a bottleneck from an estimation standpoint: real-world data often has very weak signals, which can only be estimated when the dataset size reaches a particular threshold (typically beyond hundreds of thousands of data points). As such, the MIP-based CAIPWL will fail to operate in such regimes. The second drawback is also significant in practice, as provisioning computation time is often a prerequisite in deciding beforehand whether to use a particular method. 
Practically, the computational efficiency presents the largest bottleneck for using MIP approach.
For instance, 
while it takes MIP around 3 hours to solve a problem with 500 data points in 10 dimensions on a laptop using a depth-3 tree, the solver we provide can handle tens of thousands of data points within a few seconds and millions of data points within hours (both running on the same laptop environment). In the policy learning context, this far exceeds the scale that are dealt with in the current literature and hence pushes the existing evenlope of the regime where policy learning can have a positive impact. 
Note our methods will be particularly effective in the large-$n$-shallow-tree regime\footnote{Of course, as already mentioned, for deep trees, exact optimality is not possible.}, where it is important
to learn the branching decisions right at each level. In such cases, greedy tree learning tends to perform much worse (Section~\ref{subsec:sim} provides a more detailed discussion). At the same time, the tree-search algorithm provided here also allows users to inject approximation in order to gain computational efficiency. This approximation scheme in a sense combines the best of both worlds: on the one hand, it is more efficient and hence can deal with deeper trees and larger scale dataset; on the other hand, it provides a much better approximation than greedy learning.

Throughout this section, we assume that a feature $x$ is a point in a bounded subset of $p$-dimensional Euclidean space: $\feas \subset \mathbf{R}^p$. We use $x$ to denote a point in $\feas$ and $x(i)$ to denote its $i$-th component:
$x_i$ is reserved for denoting the $i$-th point in a sequence of points $x_1, \dots, x_n \in \feas$.

\subsection{Decision Trees}\label{subsec:trees_def}

\begin{figure}[t]
	\centering
		\includegraphics[width=0.5\linewidth]{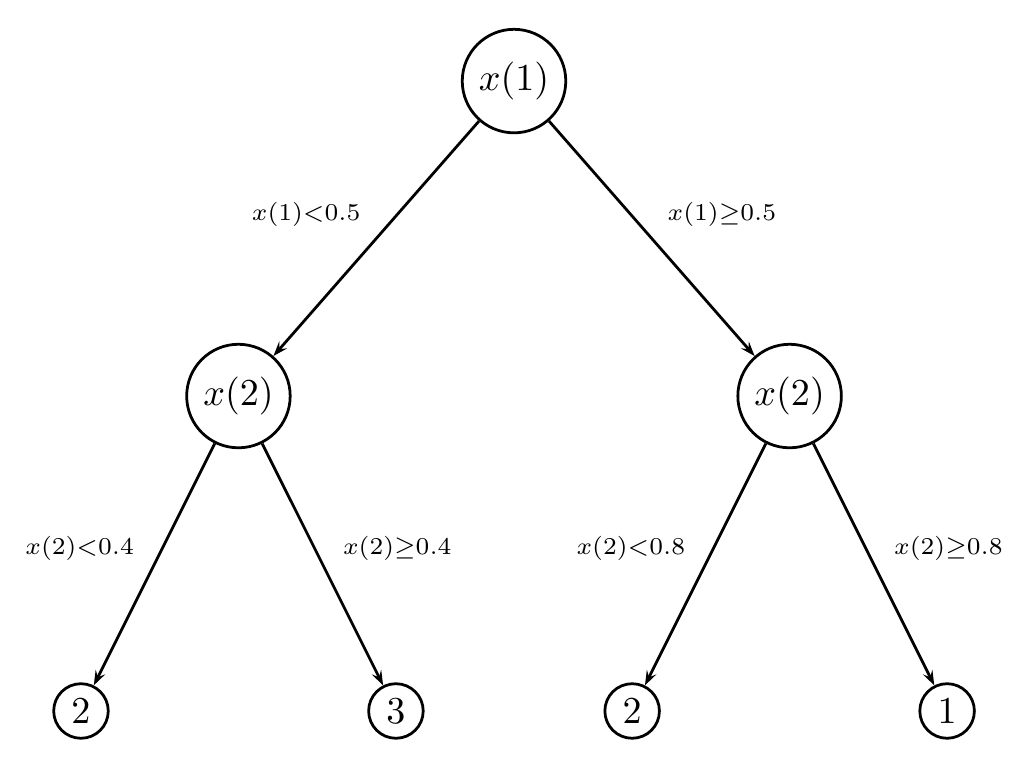}
	\caption{A depth-$3$ tree with three actions $1, 2, 3$.
More specifically, for any data point $x$, this tree will assign an action in the following way: 1) Compare its first component to $0.5$ and branch to the left if it's less than $0.5$ and to the right otherwise.  2) If branched left previously, then compare its second component to $0.4$ and branch to the left if less than $0.4$ and to the right otherwise. If branched right previously, then compare its second component to $0.8$ and branch to the left if less than $0.8$ and to the right otherwise. Whichever leaf the path terminates on, the corresponding action will be assigned. For instance, the point $(0.4, 0.5)$ will be assigned action $3$ while the point $(0.5, 0.7)$ will be assigned action $2$.}
	\label{fig:tree}
\end{figure}

Decision trees~(\cite{breiman1984classification}) map a feature $x$ into an action $a$ by traversing a particular path from the root node to a leaf node.
Specifically, following the convention in~\cite{bertsimas2017optimal}, a depth-$L$ tree has $L$ layers in total: the first $L-1$ layers consist of branch nodes, while the  $L$-th layer consists of leaf nodes. 
Each branch node is specified by two quantities: the variable to be split on and the threshold $b$.
At a branch node, each of the $p$ possible components of $x$ can be chosen as a split variable. 
Suppose $x(i)$ is chosen as the split variable for a branch node. Then if $x(i) < b$,  the left child of the node is followed; otherwise, the right child of the node is followed.
Every path terminates at a leaf node, each of which is assigned a unique label, which corresponds to one of the $d$ possible actions (different leaf nodes can be assigned the same action label).
A simple example is given in Figure~\ref{fig:tree}.
We use $\Pi_L$ to denote the set of all depth-$L$ trees.

%
%
%
%
%

Next, we characterize the complexity of the policy class formed by depth-$L$ trees.
\begin{lemma}\label{lem:entropy_integral}\quad
Let $L$ be any fixed positive integer. Then:
\begin{enumerate}
	\item 
	$\log \N_H(\epsilon, \Pi_L) \le (2^{L}-1)\log p + 2^L\log d  + (2^L -1) \log (\frac{L}{\epsilon} +1) = O(\log \frac{1}{\epsilon})$.
    \item $\kappa(\Pi) \le \sqrt{ (2^{L}-1)\log p + 2^L\log d } + \frac{4}{3}L^{\frac{1}{4}}\sqrt{ (2^L -1) }$.
\end{enumerate}
\end{lemma}

\begin{remark}\quad
Note that all the quantities $L$ (tree-depth), $p$ (feature dimension) and $d$ (action set size) are fixed. The logarithmic growth $O(\log \frac{1}{\epsilon})$ easily satisfies Assumption~\ref{assump:1}.
The tedious proof can be found in the appendix.
\end{remark}

With this characterization, a immediate corollary of Theorem~\ref{thm:main} is given next:
\begin{theorem}\label{thm:trees}\quad
	Let $\Pi_L$ be the underlying policy class. 
	Let $\pold_{CAIPWL}$ be the policy learned from Algorithm~\ref{alg:DRL} and let  Assumptions~\ref{assump:classical},\ref{assump:consistency},\ref{assump:1} hold.
	Then $R(\pold_{CAIPWL}) = O_p\Big(\Bigg(\sqrt{ (2^{L}-1)\log p + 2^L\log d } + \frac{4}{3}L^{\frac{1}{4}}\sqrt{ (2^L -1) }\Bigg) \sqrt{\frac{V_*}{n}}\Big).$ In particular, for fixed $L, p, d$,
	$R(\pold_{CAIPWL}) = O_p\Big(\sqrt{\frac{V_*}{n}}\Big).$
	
\end{theorem}	

\subsection{Policy Learning via a Mixed Integer Program}\label{subsec:mip}

In this subsection, we study the computational aspect of CAIPWL when 
the policy class is trees. Specifically, we wish to solve $\hat{\pi}_{CAIPWL} = \arg\max_{\pi \in \Pi_L} \sum_{i=1}^n \langle \pi(X_i), \Ghat_i \rangle$, where each $\Ghat_i$ is a $\vd$-dimensional vector pre-computed (from data) 
according to Equation~\ref{eq:urgent}. 
Since a feature $x$ is a point in a bounded subset of $p$-dimensional Euclidean space, for convenience and without loss of generality, we normalize $\feas$ to be $[0, 1]^p$. 

We denote by $\Tb$ and $\Tl$ the number of branch nodes and leaf nodes, respectively.
For a depth-$L$ tree, $\Tb = 2^{L-1}  - 1$ and $\Tl = 2^{L-1}$.
We number the branch nodes $1, 2, \dots, \Tb$ and the leaf nodes $1, 2, \dots, \Tl $.
A node will be referred to as a branch node $t$ or a leaf node $t$ (which makes it clear whether it is a branch node or a leaf node, as well as which one of the nodes it is).
We formulate this problem as a mixed integer program. To do so,
we introduce three sets of variables in our MIP formulation:
\begin{enumerate}
	\item $\va_t$ and $\vb_t,  t = 1, ..., \Tb$ (number of branch nodes). They encode the comparisons made at each branch node $t$, which take the form $\va_t^T X_i < \vb_t, \enspace  t = 1, ..., \Tb.$ 
We require $\va_t \in \{0,1\}^{\vp}$ and 
	$\sum_{q=1}^{\vp}  \va_{tq} = 1, \enspace  t = 1, ..., \Tb$. 
	These two constraints together ensure that at each branch node $t$, exactly one of the variables is the split variables.
	For the $\vb_t$ variables, since the data $X_i$ are normalized to lie between 0 and 1, we have $
	0 \leq \vb_t \leq 1, \enspace  t = 1, ..., \Tb$.
	
	\item $\vz_{it},  i = 1, ..., n,  t = 1, ..., \Tl$ (number of leaf nodes). $\vz_{it}$ encodes whether data point $X_i$ fell in leaf node $t$. Hence we restrict $\vz_{it} \in \{0,1\} \quad  i = 1, ..., n, \quad t = 1, ..., \Tl, \quad k = 1, ..., \vd$, and
	$\sum_{t=1}^{\Tl}  \vz_{it} = 1, \enspace  i = 1, ..., n$: each data point $X_i$ enters and only enters one leaf.
	
	\item $\vc_{kt},  k = 1, ..., \vd,  t = 1, ..., \Tl$. $\vc_{kt}$ represents whether action $k$ is assigned to leaf node $t$. If $\vc_{kt} = 1$, then $\pi(x) = k$ for all $x$ in leaf $t$. Each leaf node is associated with one action: $
	\label{vc_sumto1}
	\vc_{kt} \in \{0, 1\}, \quad t = 1, ..., \Tl, \quad k = 1, ..., \vd,
	\sum_{k=1}^{\vd}  \vc_{kt} = 1, \quad  t = 1, ..., \Tl.$
	Note that in general, $\vc_{kt}$ need not be binary: $\sum_{k=1}^{\vd}  \vc_{kt} = 1$ and $ 0 \le \vc_{kt} \le 1$ suffice. In the latter case,  $\vc_{kt}$ represents the probability of taking action $k$ for leaf node $t$: this corresponds to a 
	random decision tree. 
	As mentioned before, when the data is generated independently, there is no loss of generality to only use deterministic policies, as an optimal policy can always be achieved by a deterministic policy. That being said, we will freely switch to random policies if doing so proves more computationally efficient.
	\end{enumerate}
    
With these variables, one can check that the objective function can be expressed as:
\begin{lemma}\quad
	$\sum_{i=1}^n \langle \pi(X_i), \Ghat_i \rangle =\sum_{i=1}^{n} \sum_{k=1}^{\vd} \sum_{t=1}^{\Tl} \vz_{it} \vc_{kt} \Ghat_{ik}$
\end{lemma}

The value of this objective function exactly matches the value of the $Q$-function. Thus, maximizing this objective will give the highest achievable value of $\hat{Q}(\cdot)$ over the space of decision tree policies. To see this, recall that the value $\pi(X_i)$ is a one-hot vector, with a ``1'' in the $k$th position, signifying that action $k$ should be taken for $X_i$. The value of $\langle \pi(X_i), \Ghat_i \rangle$, then, is equal to $\Ghat_{ik}$. 
We only want to add the values of $\Ghat_{ik}$ which correspond to the actions our policy chooses for each data point $i$. However, this information is exactly encoded in the variables $\vz_{it}$ and $ \vc_{kt}$. 
Therefore, this objective function causes the MIP to select $\vz_{it}$ and $ \vc_{kt}$ (i.e. classify the data into leaves and associate actions with the leaves) so that the maximal sum of estimates is achieved. 
Note also that the objective is quadratic and can be written as $y^T Q y$, where $y$ is the vector of decision variables of the MIP that consists of $\va_t, \vb_t,\vz_{it},\vc_{kt}$, and where $Q$ is the data matrix that consists of $\Ghat_{ik}$ and other constants. Such conversion can be done in a straightforward manner with elementary matrix algebra (and some care), which we omit here. It suffices to note that most MIP solvers allow for quadratic objectives.

Our MIP formulation is nearly complete; the final set of elements we need to include are the branch constraints associated with each data point (of the form $\va_t^T X_i < \vb_t$). 
 For each leaf node $t$, we divide the set of nodes traversed to reach that leaf node into two sets: node indices where the left branch was followed, LB$(t)$, and node indices where the right branch was followed, RB$(t)$. 
Note if $X_i$ falls in node $t$, then $\va_l^T X_i < \vb_l$ for all branch nodes $l$ in LB$(t)$, and $\va_r^T X_i \geq \vb_r$ for all branch nodes $r$ in RB$(t)$.
To enforce this, we make use of the binary variables $\vz_{it}$. We only want to enforce the branch constraints for the leaf into which the data point falls, and not any of the other leaf nodes. 
For the right branch constraints, we add an extra term on the right side: $
\va_r^T X_i \geq \vb_r - (1-\vz_{it}), \enspace \forall r \in \text{RB} (t), \enspace  t = 1, ..., \Tl, \enspace  i = 1, ..., n,$
so that the inequality is always true when $\vz_{it} \neq 1$ (the case where $X_i$ does not fall into leaf $t$).  

We proceed similarly with the left branch constraints: $
\va_l^T X_i < \vb_l + (1-\vz_{it}), \enspace \forall l \in \text{LB} (t), \enspace  t = 1, ..., \Tl, \enspace  i = 1, ..., n,$
but here we need to add a small constant $\ve$ to the left side of the strict inequalities (and turn the strict inequalities into inequalities), because MIP solvers (and most general LP solvers) can only handle non-strict inequality constraints. 
Because of numerical considerations, we want this constant to be as large as possible. In the current context, we want to add the largest allowable amount which does not cause any data point to enter a different leaf node. To accomodate this, we follow~\cite{bertsimas2017optimal} and add a vector of small values, $\ebf$, to the data vector $X_i$, instead of adding a small value to the entire left side. 
For each element of $X_i$, the largest $\ve_{\vq}$ we can add is the smallest nonzero difference between the $\vq$th element of any two data points in our dataset:
\begin{align}
\ve_{\vq} = \min_{i,j} \left(X_{i\vq} - X_{j\vq} \mid X_{i\vq} \neq X_{j\vq}\right). 
\end{align}
Our left branch constraints then become:
\begin{align}
\va_l^T \left(X_i + \ebf\right) \leq \vb_l + B_l(1-\vz_{it}), \quad \forall l \in \text{LB} (t), \quad  t = 1, ..., \Tl, \quad  i = 1, ..., n.
\end{align}
However, our addition of $\ebf$ means we can't simply choose $B_l = 1$. Instead, we set $B_l = 1 + \emax$, where $\emax = \max(\ebf)$. Since $\va_{tq}$ lie between 0 and 1, this ensures that the left branch constraints are only enforced for the leaf $t$ into which data point $i$ falls into (i.e., where $\vz_{it} = 1$). 
The left branch constraints then become $\va_l^T X_i + \ebf \leq \vb_l + (1 + \ebf)(1-\vz_{it}), \enspace \forall l \in \text{LB} (t), \enspace  t = 1, ..., \Tl, \enspace  i = 1, ..., n$, where we multiply by $(1 + \ebf)$ to ensure that the left branch constraints are only enforced for the leaf $t$ into which data point $i$ fell into (i.e., where $\vz_{it} = 1$). 

Here is the final MIP, the solution to which is exactly $\hat{\pi}_{CAIPWL}$ (after some reconstruction) :
\begin{align}
\begin{split}
\label{eq:mip_quad}
\text{maximize} \quad &\sum_{i=1}^{n} \sum_{k=1}^{\vd} \sum_{t=1}^{\Tl} \vz_{it} \vc_{kt} \Ghat_{ik}\\
\text{subject to} \quad	
&\sum_{q=1}^{\vp}  \va_{tq} = 1, \quad  t = 1, ..., \Tb,\quad
\sum_{t=1}^{\Tl}  \vz_{it} = 1, \quad  i = 1, ..., n,\quad
\sum_{k=1}^{\vd}  \vc_{kt} = 1, \quad  t = 1, ..., \Tl,\\
&\va_l^T X_i + \ebf - \vb_l + (1+\ebf)\vz_{it} \leq 1+\ebf, \quad \forall l \in \text{LB}(t), \quad  t = 1, ..., \Tl, \quad  i = 1, ..., n, \\
&\va_r^T X_i - \vb_r - \vz_{it} \geq -1, \quad \forall r \in \text{RB} (t), \quad  t = 1, ..., \Tl, \quad  i = 1, ..., n, \\
&\va_{tq} \in \{0, 1\},  \quad  t = 1, ..., \Tb, \quad q = 1, ..., \vp, 0 \leq \vb_t \leq 1, \quad  t = 1, ..., \Tb,\\
&\vz_{it}, \vc_{kt} \in \{0, 1\},\quad  i = 1, ..., n, \quad t = 1, ..., \Tl, \quad k = 1, ..., \vd.
\end{split}
\end{align}


\subsection{Policy Learning via Tree Search}\label{subsec:tree_search}

The MIP based approach for finding the optimal tree given in Section~\ref{subsec:mip} has both advantages and disadvantages. The primary advantage of MIP lies in its convenience and simplicity: once we formulate the tree-based policy learning problem as a MIP given in Equation~\eqref{eq:mip_quad}, the task is essentially completed as one can freely employ available off-the-shelf MIP solvers. Furthermore, as the technology behind these off-the-shelf solvers continue to develop, the size of solvable problems will continue to grow, without any extra effort from the decision maker who utilizes the MIP based approach.
However, a key disadvantage of the MIP based approach, at least at the time of this writing, is its inability to solve problems beyond moderate scale.

To fully explain this point, we start by introducing some notation. Throughout the paper, we have used $n$ to denote the number of training data points, $p$ to denote the feature dimension, $L$ to denote the tree depth, $d$ to denote the number of actions. Consequently, the size of the tree-based policy learning problem can be succinctly characterized by the following list of parameters $(n, p, L, d)$. 
In the MIP formulation, we have four classes of variables: $\va$, $\vb$, $\vz$ and $\vc$.
For a problem of size $(n, p, L, d)$, it easily follows from the definitions of those variables that we have
$pT_B$ variables for $\va$, $T_B$ variables for $\vb$, $nT_L$ variables for $\vz$ and
$dT_L$ variables for $\vc$,
where $T_B = 2^{L-1} -1$ and $T_L = 2^{L-1}$.
Consequently, we have in total $(p+1)(2^{L-1} -1) + (n+d)2^{L-1}$ variables.
In particular, the number of variables in the MIP scales linearly with the number of data points and exponentially with the tree depth. When we fix the tree depth $L$ (as well as $p$ and $d$) to be constant, 
the linear growth of the number of variables in terms of $n$ may seem very mild. However, and perhaps most unfortunately, the amount of time it takes to solve a MIP scales exponentially with the number of variables (in the worst case). Consequently, even holding all other parameters fixed, the running time of the MIP optimization approach scales exponentially at worst with the number of training data points $n$.
To get a quick sense, it takes 3 hours to solve a problem of the scale $(500, 10, 3, 3)$ on a Macbook Pro laptop with a 2.2 GHz Intel Core i7 processor and 16 GB RAM~\footnote{Beyond 1000 points, the program does not finish running within a day.}, a scale that contains 2045 variables in total for the corresponding MIP.
Consequently, even with a powerful server, going beyond low thousands for $n$ using MIP is simply not a possiblity for a depth $3$ tree  at this time of writing.
Furthermore, precisly due to the blackbox nature of a MIP solver, one cannot gauge the running time for a given problem size (as the commerical off-the-shelf solver is typically closed sourced). This creates difficulty for decision makers to decide whether the required running time is the worth the investment for the problem at hand.

Motivated by the above drawbacks of the MIP approach, we provide in this paper a customized tree-search algorithm that finds the optimal policy. The algorithm searches in a principled way through the space of all possible trees in order to identify an optimal tree. While the actual algorithm has many engineering optimization details (in order to run fast practically), the key ingredient of the algorithm relies on the following observation: for a fixed choice of a parent node's parameters (i.e. which variable to split and the split value), the problem of finding an optimal tree for the given data decouples into two independent but smaller subproblems: one for the left subtree, and the other for the right subtree. Consequently, we obtain a recursive tree-search algorithm that, at least in principle, is quite simple (see Algorithm~\ref{alg:ts} for a vanilla version of the algorithm). Putting everything together, we can integrate the tree-search algorithm seamlessly into CAIPWL to form a complete policy learning algorithm. This is given in Algorithm~\ref{alg:tscaipwl}.

\begin{algorithm}
	\caption{Tree-Search: Exact and Approximate} 
	\label{alg:ts}
	\begin{algorithmic}[1]
		\STATE \textbf{Input:} $\{(X_i, \Ghat_i )\}_{i=1}^n$, depth $L$ and approximation parameter $A$.
		\IF {$L = 1$}
		\STATE Return \Big($\max_{j \in \{1,\dots, d\}} \sum_{j=1}^n\Ghat_i(j)$ , $\arg\max_{j \in \{1,\dots, d\}} \sum_{j=1}^n\Ghat_i(j)$\Big) 
		
		\ELSE
		\STATE Initialize $\text{reward} = - \infty$, $\text{tree} = \emptyset$
		\FOR {$m = 1, 2, \dots, p$}
		\STATE Sort the $X_i$'s according to the $m$-th coordinates (and merge the same values)
		\FOR {$i = 1, 2, \dots, \ceil{\frac{n}{A}}-1$}
		\STATE $(\text{reward\_left}, \text{tree\_left}) =$Tree-Search$(\{(X_l, \Ghat_i )\}_{l=1}^{iA}, L-1, A)$
		\STATE $(\text{reward\_right}, \text{tree\_right}) =$Tree-Search$(\{(X_l, \Ghat_i )\}_{l=iA+1}^{n}, L-1, A)$
		\IF {$\text{reward\_left} + \text{reward\_right} > \text{reward}$}
		\STATE $\text{reward} = \text{reward\_left} + \text{reward\_right} $
		\STATE  $\text{tree} =$ Tree-search$ [m, \frac{X_{iA}(m) + X_{iA + 1}(m)}{2}, \text{tree\_left},\text{tree\_right} ] $
		\ENDIF
		\ENDFOR
		\ENDFOR
		\STATE Return $(\text{reward}, \text{tree})$
		\ENDIF
	\end{algorithmic}
\end{algorithm}

\begin{algorithm}
	\caption{Tree-Search Based CAIPWL} 
	\label{alg:tscaipwl}
	\begin{algorithmic}[1]
		\STATE \textbf{Input:} Dataset $\{(X_i, \W_i, Y_i)\}_{i=1}^n$, depth $L$ and approximation parameter $A$.
		\STATE Choose $K > 1$.
		\FOR {$k = 1, 2, \dots, K$}
		\STATE Build estimators $\hat{\mu}^{-k}(\cdot) = \begin{bmatrix}
		\hat{\mu}_{a^1}^{-k}(\cdot) \\
		\hat{\mu}_{a^2}^{-k}(\cdot)\\
		...\\
		\hat{\mu}_{a^d}^{-k}(\cdot) 
		\end{bmatrix}, \hat{e}^{-k}(\cdot) =\begin{bmatrix}
		\hat{e}_{a^1}^{-k}(\cdot) \\
		\hat{e}_{a^2}^{-k}(\cdot)\\
		...\\
		\hat{e}_{a^d}^{-k}(\cdot) 
		\end{bmatrix}$
		using the rest $K-1$ folds.
		\ENDFOR
		\STATE Collect $\Ghat_i  = \frac{Y_i - \hat{\mu}_{\W_i}^{-k(i)}(X_i)}{\hat{e}^{-k(i)}_{\W_i}(X_i)} \cdot  \W_i
		+
		\begin{bmatrix}
		\hat{\mu}_{a^1}^{-k(i)}(X_i) \\
		\hat{\mu}_{a^2}^{-k(i)}(X_i)\\
		...\\
		\hat{\mu}_{a^d}^{-k(i)}(X_i) 
		\end{bmatrix}, i = 1, \dots, n $
		\STATE Return Tree-Search$(\{(X_i, \Ghat_i )\}_{i=1}^n, L, A)$.
	\end{algorithmic}
\end{algorithm} 

Finally, 
as a quick summary, we mention that the customized tree-search algorithm have several benefits over the MIP approach: 1) it is much more efficient compared to the MIP based approach (for instance, a problem of size (1000, 10, 3, 3) takes less than a second to solve on the same computing environment); 2) the running time characterization is explicit and hence one can project the total running time for a given problem size at hand before deciding whether the time investment is worthwhile; 3) it has an extra knob that allows computational efficiency to be smoothly traded off with accuracy; 4) it runs even faster when features are discrete (a common scenario that arises in practice), whereas the MIP approach sees no difference in computational efficiency in continuous versus discrete settings. See Appendix~\ref{subsec:complexity} for a detailed discussion of the complexity of the tree search algorithm.

\subsection{Simulations}\label{subsec:sim}
In this section, we present a simple set of simulations to demonstrate the efficacy of the learning algorithm.
\subsubsection{Setup}
\begin{figure}[t]
	\centering
	\includegraphics[width=0.5\linewidth]{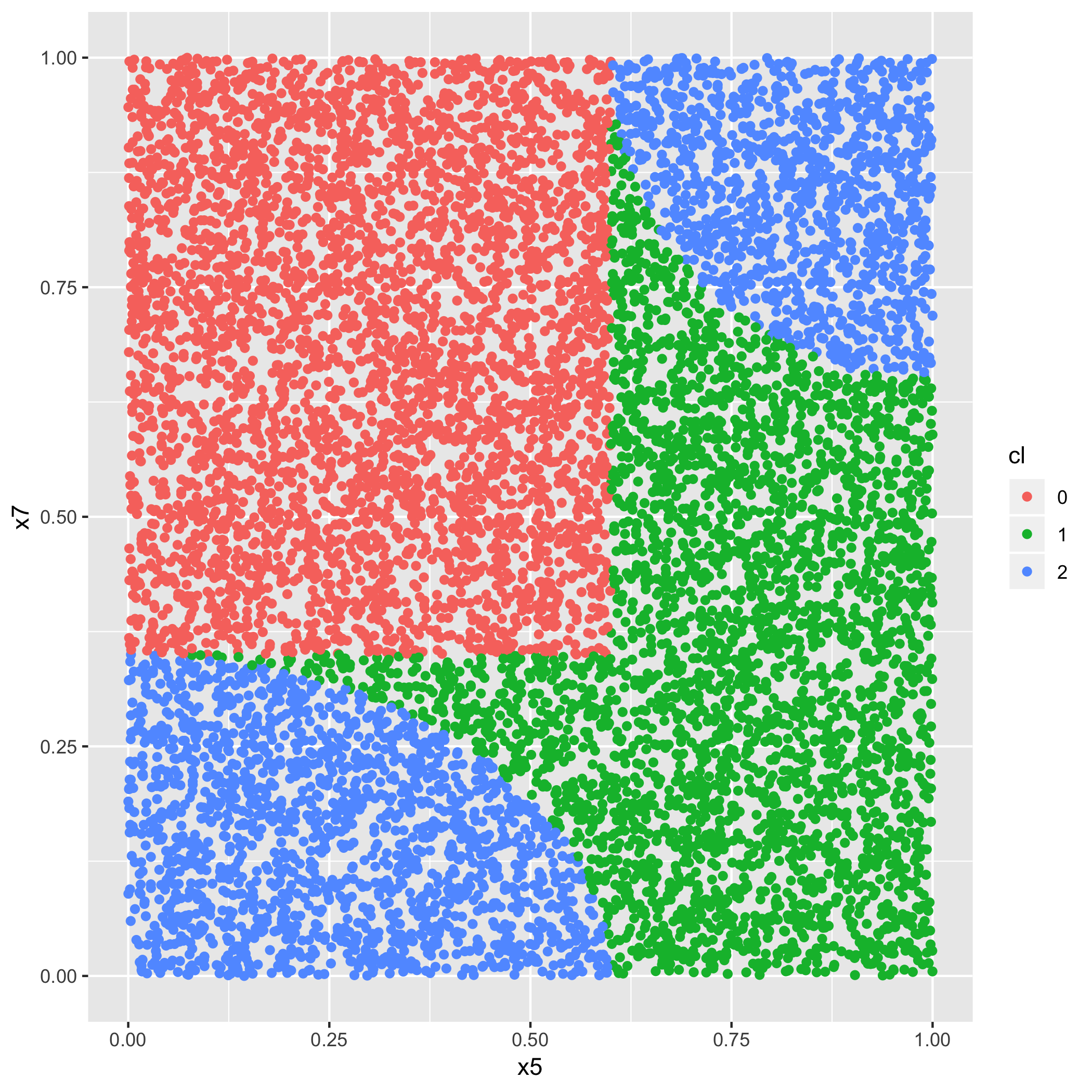}
	\caption{A pictorial illustration of the setup of the data-generating distribution. There are two dimensions (out of the total 10 dimensions of the entire feature space) that determine the action selection probabilities and the rewards: $x_5$ and $x_7$. The underlying region is then dividedn into three disjoint regions: Region $0$ (read), Region $1$ (green) and Region $2$ (blue). The figure provides a two-dimensional slice of the entire $10$-dimensional feature space. The probabilities of selecting an action is given in Table~\ref{table:prob}. Finally, the optimal action in each region coincides with the region number.}
	\label{fig:gen}
\end{figure}

We start by describing the data-generating process from which the training dataset is collected. Each $X_i$ is a $10$-dimensional real random vector drawn uniformly at random from \textit{iid} from $[0, 1]^{10}$.
We denote the $10$ dimensions by $x_0, x_1, \dots, x_9$.
There are three actions, which we label as $0, 1, 2$. 
The probability of each action being selected depends on the features and is succinctly summarized by Table~\ref{table:prob}. In the table, the three regions, Region 0, Region 1 and Region 2 form a partition of 
the entire feature space $[0, 1]^{10}$:
\begin{enumerate}
	\item Region 0:  $\{x \in [0, 1]^{10} \mid 0 \le x_5 < 0.6,  0.35 < x_7 \le 1\}$.
	\item Region 2:  $\{x \in [0, 1]^{10} \mid \frac{x_5^2}{0.6^2} + \frac{x_7^2}{0.35^2} < 1\} \cup \{x \in [0, 1]^{10} \mid \frac{(x_5-1)^2}{0.4^2} + \frac{(x_7-1)^2}{0.35^2} < 1\} $.
	\item Region 1: $[0, 1]^{10} -$ (Region $0$ $\cup$ Region $2$).
\end{enumerate}
See Figure~\ref{fig:gen} for a quick visualization of the three regions (each in a different color).
Finally, each reward $Y_i$ is generated \textit{iid} from $\mathcal{N}(\mu_{A_i}(X_i), 4)$, where $A_i$ is generated \textit{iid} from $X_i$ according to the probabilities in Table~\ref{table:probs} and the mean reward function $\mu_a(x)$ is given as follows:

\begin{table}[]
	\begin{tabular}{l|llll|}
		\cline{2-5}
		& Region 0 & Region 1 & Region 2 &  \\ \hline
		Action 0 & 0.2     & 0.2     & 0.4      &  \\ \hline
		Action 1 & 0.6     & 0.6      & 0.2      &  \\ \hline
		Action 2 & 0.2     & 0.2     & 0.4      &  \\ \hline
	\end{tabular}
	\caption{The probabilities of selecting an action based on features. In Regions $0$ and $1$, the probabilities of selecting $0$, $1$, $2$ are 0.2, 0.6 and 0.2 respectively. In Region $2$, the respective probabilities are 0.4, 0.2 and 0.4.}
	\label{table:probs}
\end{table}

\begin{equation*}
\mu_a(x) = \begin{cases}
3-a, &\text{if $x \in $ Region $0$}\\
2 - \frac{|a-1|}{2}, &\text{if $x \in $ Region $1$}\\
1.5(a- 1), &\text{if $x \in $ Region $2$.}\\
\end{cases}
\end{equation*}
The rewards are constructed such that: 1) the optimal action in each region is the same as the region number (action $0$ is optimal in Region $0$; action $1$ is optimal in Region $1$ and action $2$ is optimal in Region $2$). 2) The noise has variance $4$, which is sizable in the current setting (the largest possible reward in any region is $2$ while the standard deviation of the noise is also $2$. This provides an interesting 
regime for learning a good policy since the signals can easily be obfuscated
by noise at this scale.

\begin{figure*}[t]
	\centering
	\begin{tabular}{ccc}
		\includegraphics[width=0.4\textwidth]{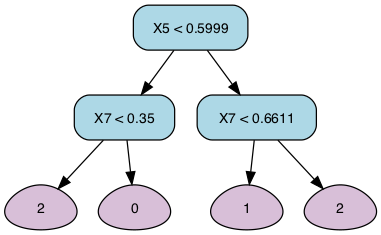} &
		\includegraphics[width=0.4\textwidth]{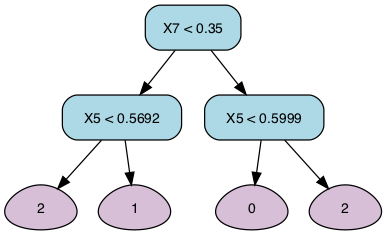} & \\
		(a) & (b)
	\end{tabular} 
	\caption{\textbf{(a) Best depth-3 decision tree for the simulation setup. This is learned by generating 10000 noise-free data points and applying the tree-search algorithm on this clean data. (b) Tree learned by applying the greedy search algorithm on the same clean dataset.}}
	\label{fig:best_trees}
\end{figure*} 

Under the above setup, the best decision tree is given in Figure~\ref{fig:best_trees}(a).
Here, we learn the best decision tree by generating 10000 noise-free data samples and directly performing exact tree-search using Algorithm~\ref{alg:ts} on rewards that are true mean rewards. 
In fact, by inspecting the simulation setup, we can easily see that Figure~\ref{fig:best_trees}(a) provides
the best decision tree (up to some rounding errors): since the rewards differ less in different actions in Region 1 compared to Region 0 and Region 2, selecting the wrong action for Region 1 (green region in Figure~\ref{fig:gen}) is less consequential. Hence, it is best to make the first cutoff at $x_5 = 0.6$ and then split at $x_7$. Further, when splitting at $x_7$ in the second layer,  the left subtree should be split at  $0.35$ exactly because action only action $0$ and action $2$ will be in the leaves (action $1$ is never optimal in either case).
On the other hand, for the right subtree, the split on $x_7$ should be somewhat above 0.65 because there is a trade-off between the green area (action 1) and the blue area (action 2).

 \subsubsection{Empirical Regret Comparisons} 
 
 We run $7$ different learning algorithms on the simulated data and compare the empirical regret performances. We start by explaining in more detail about these five different methods:
 
 \begin{enumerate}
 	\item CAIPWL-opt. This is the proposed tree-search CAIPWL given in Algorithm~\ref{alg:ts}.
 	Here, we use the suffix ``opt" to highlight the fact that the policy optimization step is solved to exact optimality: there is no approximation made (i.e. the approximation parameter $A = 1$). 
 	\item CAIPWL-skip. This is the same as tree-search CAIPWL-opt except we make the approximation of skipping points.
 	Specifically, we set the approximation parameter $A = 10$. 
 	\item CAIPWL-greedy. This is a similar algorithm to CAIPWL-opt, with the important distinction that
 	in the policy optimization step, a greedy approach is used to select the final tree. In other words, the chosen policy does not maximize the estimated policy value. More specifically, at each branch node, greedy selects the split dimension and the split point assuming it is a tree of depth $2$. 
 	This is the approach adopted in~\cite{langford2011doubly}
 	\item IPWL-opt. This approach (\cite{langford2011doubly,swaminathan2015batch}) starts with policy evaluation by using the IPW estimator and then proceeding to select the policy that maximizes the policy value. Here we also added cross-fitting to the vanilla IPWL to increase its performance.
 	\item IPWL-skip. Same as IPWL-opt with skipping points (again $A = 10$).
 	\item IPWL-greedy. Similar to IPWL, with the distinction again being the policy optimization performed using the greedy approach.
 	\item Random. This is the basic benchmark policy that always selects an action uniformly at random given any feature. Note that even though this is a fixed policy, it can be thought of as a constant learning algorithm that does not depend on training data.
 \end{enumerate}

Second, the above methods rely on estimating the propensities $e$ and/or mean rewards $\mu$. We next  explain how such estimation is done. In general, either of the two quantities can be estimated by either a parametric method or a non-parametric method. Our specific choices are as follows:

For estimating propensities $e$, we use multi-class logistic regression to produce $\hat{e}_a(\cdot)$ for each $a$. This is a parametric estimation method that is quite computationally efficient, with free R packages publicly available. 
(Random forest would be another, non-parametric choice.)

For estimating mean rewards $\mu$, we use a non-parametric method, which is more complicated and breaks into several steps. First, we estimate the overall mean reward function $m(x) = \sum_{a \in \mathcal{A}} e_a(x) \mu_a(x)$. This is done by simply regressing $Y_i$'s to $X_i$'s using random forest, which then produces $\hat{m}(x)$ for any given $x$. Second, for each action $a$, we estimate the quantity
$\tau_a(x) = \mu_a(x) - \frac{m(x) - e_a(x) \mu_a(x)}{1 - e_a(x)}$. This quantity can be viewed as the performance difference between this action $a$ and the weighted average of all the other actions.
This quantity can be conveniently estimated using the generalized random forest approach developed in~\cite{athey2016solving}, which also provides a free software package. Third, noting that $\mu_a(x) = m(x) + (1-e_a(x)) \tau_a(x)$, we can combine the above components $\hat{m}(x), \hat{\tau}_a(x), \hat{e}_a(x)$ (the propensities $\hat{e}_a(x)$ are the same estimates using multi-class logistic regression) into the reward estimator $\hat{\mu}_a(x)$, where $\hat{\mu}_a(x) = \hat{m}(x) + (1-\hat{e}_a(x)) \hat{\tau}_a(x)$.

 \begin{table}[]
 	\begin{tabular}{l|llllll|ll}
 		\cline{1-8}
 		& CAIPWL-opt & CAIPWL-skip & CAIPWL-greedy & IPWL-opt & IPWL-skip & IPWL-greedy & Random \\ 
 		\cline{1-7}
 		1000 & 0.1280957    & 0.1266719   &  0.4923080    & 0.2115385   &  0.2133970   &  0.5159509 &  0.8679667  \\ 
 		\cline{1-7}
 		1500 &   0.0767156   & 0.0774815   & 0.3267238    &0.2276709    & 0.2297918    & 0.4955450 &  \\ 
 		\cline{1-7}
 		2000 & 0.0525168   &  0.0544816    & 0.2479887    & 0.2173981    & 0.2184838   & 0.4888422  \\ 
 		\cline{1-7}
 		2500 & 0.0393409   & 0.0399413    & 0.1943858   &  0.2154734   & 0.2143081    & 0.4827773 \\ \hline
 	\end{tabular}
 	\caption{Table of regrets. The first row shows the $7$ different methods. The first column indicates the total number of training data points used. Each number in the table is the average regret of the corresponding method, computed by performing 400 different simulation runs and obtaining the average. For each simulation run, the regret is computed by learning a policy using the corresponding learning algorithm and then evaluating the regret by using a 15000-datapoint test set generated independently from the training data. The number of folds used in the entire experiments is $5$ for all methods (except Random).}
 	\label{table:prob}
 \end{table}

\begin{figure}[t]
	\centering
	\includegraphics[width=0.5\linewidth]{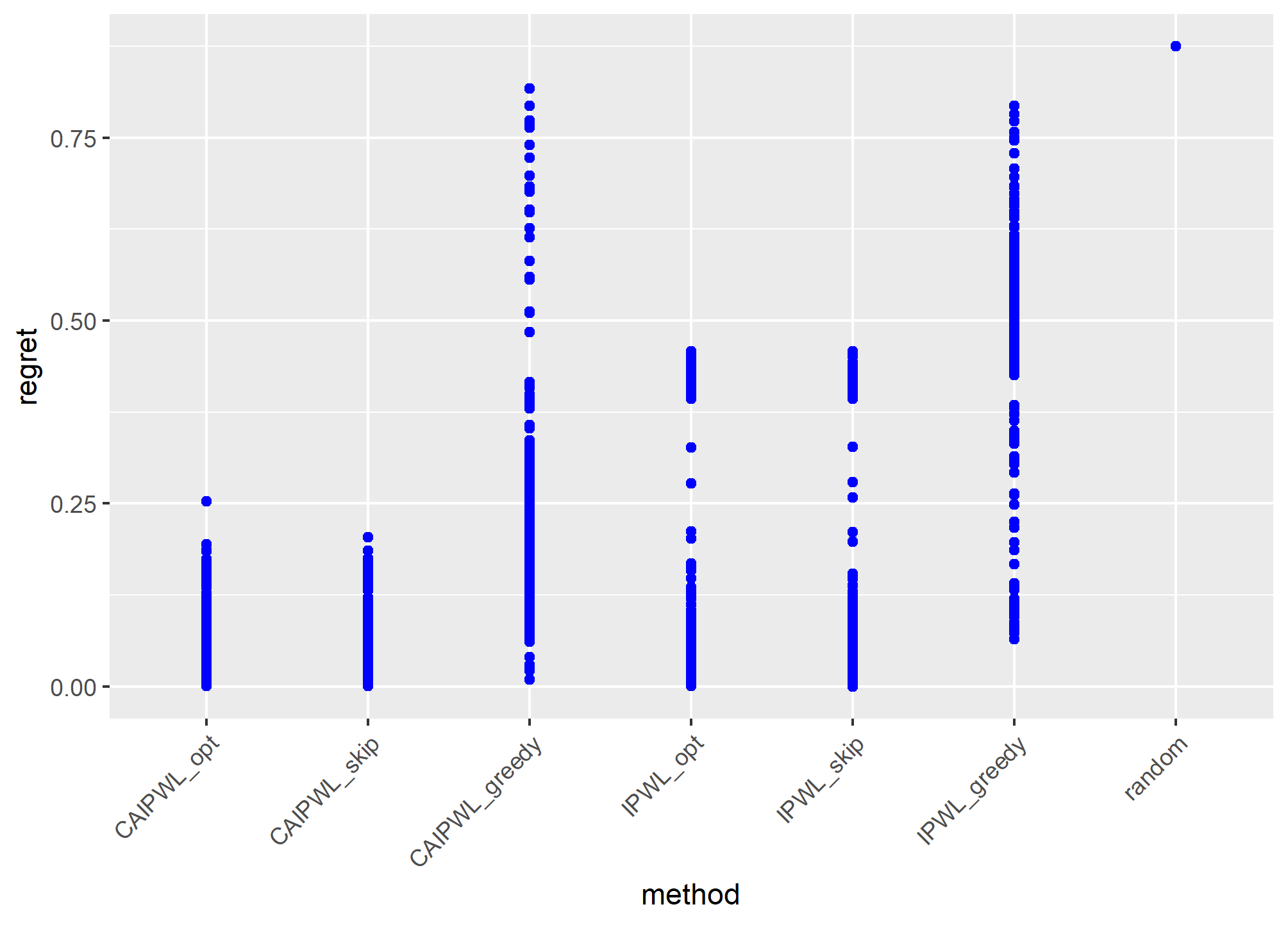}
	\caption{A picture plotting all the regrets for the 400 runs for each of the 7 methods under $n = 2500$. It is generated by plotting a single dot for a regret value in each run and for each method. Consequently, the denser the area, the more runs that fall into that regret range. Note that the regret of the random policy does not vary since it is a fixed numder under a fixed simulation setup.}
	\label{fig:2500}
\end{figure}

\begin{figure}[t]
	\centering
	\includegraphics[width=0.5\linewidth]{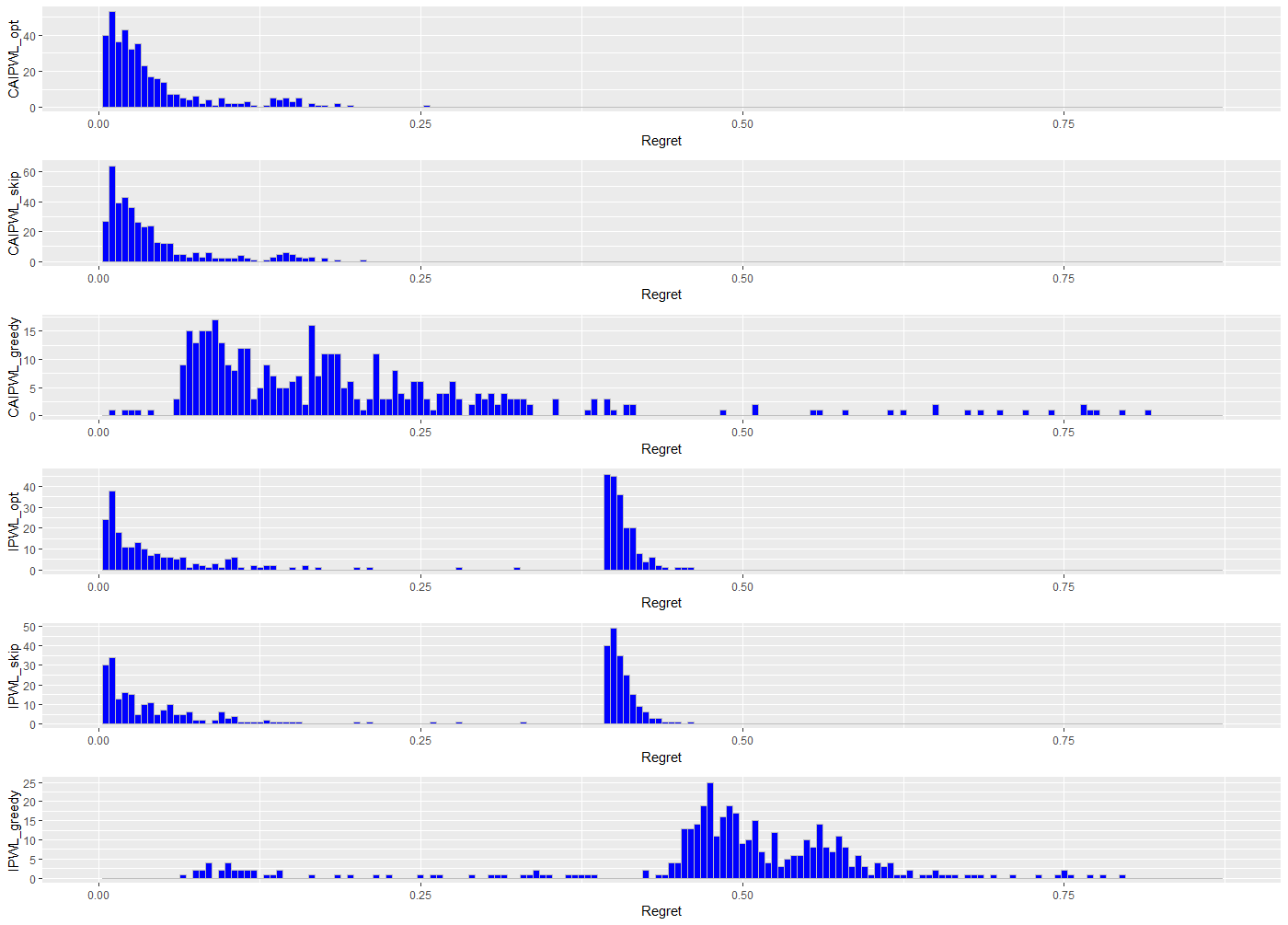}
	\caption{A histogram showing empirical counts for regret values across 400 simulation runs under $n = 2500$. Each subfigure is a histogram for a particular method.}
	\label{fig:2500_histo}
\end{figure}

Finally, once a tree is computed from a learning algorithm, we then evaluate its performance by computing the regret using a freshly generated test set containing 15000 datapoints. In particular, each feature vector is drawn according to the underlying distribution. After that, the learned tree and the optimal tree will each select an action based on that feature vector. Finally, with the feature vector and the corresponding actions, we directly generate the true mean rewards. We then average the differences (reward of optimal tree minus reward of learned tree) across these 15000 points. We do so for each of the $7$ methods, which constitutes as a single run. We then perform 400 separate different runs for each $n$ (ranging from 1000 to 2500, with increments of 500 each) and average the corresponding empirical regrets across all the runs.
The final results are cataloged in Table~\ref{table:prob}. For ease of visualization, the empirical regret distributions for $2500$ training samples are also displayed in Figure~\ref{fig:2500} and Figure~\ref{fig:2500_histo}.

Several important observations are worth making here from the simulation results.
First, when the optimization procedure is fixed, CAIPWL is more (statistically) efficient than IPWL.
Furthermore, IPWL has a much higher variance in terms of the learned policy's policy.
This is because IPWL estimates the policy value by solely weighting the rewards by estimated propensities, which can vary considerably particularly when the actual propensity is small. CAIPWL, on the other hand, mitigates this high variability by contributing an additional model estimate that complements the pure weight-by-propensity approach. In particular, with 2500 datapoints, CAIPWL is already close-to-optimal.
Second, all else equal, greedy performed much worse than opt or skip while opt and skip are comparable (with opt slightly better). To latter is easily understandable: skipping 10 points each in a dataset of size at least 1000 points essentially amounts to divide the interval into 100 quantiles; further, as mentioned earlier, such division is dynamic based on the training data rather than fixed.
To understand the former, we note that the greedy optimization procedure has two important sources of errors. First, as with all other optimization procedures, the estimates are noise-prone and hence the selection of split value can be off. Second, and more importantly, even without noise, the tree learned by greedy is prone to structural-error: it can simply fail to choose the right variable to split because its selection process is myopic. This point is best illustrated by Figure~\ref{fig:best_trees}(b), where 10000 noise-free data points are fed into the greedy learning optimization procedure to produce the best possible greedy tree.
Note that here, greedy first splits on $x_7$ rather than $x_5$. This is because when only one cut can be made, from the data-generating distribution given in Figure~\ref{fig:gen},  it is clear that splitting at $x_7 = 0.35$ is the best option because the difference between red and blue when splitting along $x_7$ is larger than that when splitting along $x_5$ (the green region is less important because the reward differences among different actions are much smaller). Consequently, when greedy learns the wrong structure of the tree (i.e. which variable to split on at each level), the errors will be much larger when combined with the already noisy estimations. Finally, comparing the above results yields the following overall insight: when either estimation or optimization falls short, the other can provide a performance boost. Consequently,
in practice, it is important to combine the best in estimation and optimization to achieve a powerful learning algorithm. In that regard, CAIPWL-skip provides the best computationally-efficient choice for trees in a low signal-to-noise regime.

\section{Application on a Voting Dataset}

In this section, we apply our method to a voting dataset on August 2006 primary election.
This dataset was originally collected by~\cite{gerber2008social} to study the motivations
for why people vote. Our goal in this section is to apply policy learning algorithms to this
dataset and illustrate some interesting findings.

\subsection{Dataset description}
We start with a quick description of the dataset and only focus on aspects that are relevant to our current policy learning context. 
The dataset contains 180002 data points (i.e. $n = 180002$), each corresponding to a single voter in a different household. The voters span the entire state of Michigan. There are 10 voter characteristics we use as features: year of birth, sex, household size, city, g2000, g2002, g2004, g2000, p2002, p2004. The first 4 features are self-explanatory. The next three features are outcomes for whether a voter has voted for general elections in 2000, 2002 and 2004 respectively: $1$ was recorded if the voter did vote and $0$ was recorded if the voter didn't vote. The last three features are outcomes for whether a voter has voted for primary in 2000, 2002 and 2004. As pointed out in~\cite{gerber2008social}, these 10 features are commonly used as covariates for predicting whether an individual voter will vote\footnote{Many other features were also recorded in the dataset. However, they are not individual features: most of them are features on a county level, such as percentages of different races etc. A cleaned up version of the data can be found at https://github.com/gsbDBI/ExperimentData/tree/master/Social.}.

There are five actions in total, as listed below:

\textbf{Nothing}: No action is performed.

\textbf{Civic}: A letter with ``Do your civic duty" is emailed to the household before the primary election.

\textbf{Monitored} : A letter with ``You are being studied" is emailed to the household before the primary election. Voters receiving this letter are informed that whether they vote or not in this election will be observed.

\textbf{Self History}: A letter with the voter's past voting records as well as the voting records of other voters who live in the same household is mailed to the household before the primary election. The letter also indicates that, once the election is over, a follow-up letter on whether the voter has voted will be sent to the household.

\textbf{Neighbors}: A letter with the voting records of this voter, the voters living in the same household, the voters who are neighbors of this household is emailed to the household before the primary election. The letter also indicates that ``all your neighbors will be able to see your past voting records" and that follow-up letters will be sent so that whether this voter has voted in the upcoming election will become public knowledge among the neighbors.

In collecting this dataset, these five actions are chosen at random independent of everything else, with probabilities equal to $\frac{10}{18},\frac{2}{18},\frac{2}{18},\frac{2}{18},\frac{2}{18}$ (in the same order as listed above). Finally, the outcome is whether a voter has voted in the 2006 primary election, which is either $1$ or $0$.

\subsection{Direct Application of CAIPWL}
With the above data, we can directly run policy learning algorithms to obtain a policy, which maps the features to one of the five actions. We apply both CAIPWL and IPWL on the dataset, using depth-$3$ trees, where exact optimization is performed using the tree-search algorithm given in the previous section.
The propensities are estimated using random forests (and clipped at the lower bound $0.1$ for stability).
The model estimates are performed using the same method as described in Section~\ref{subsec:sim}.
We then compare the performance of trees learned via these two algorithms to that of the following six policies: 1) randomly selecting an action, 2) always choose \textbf{Nothing}, 3) always choose \textbf{Civic},
4) always choose \textbf{Monitored}, 5) always choose \textbf{Self History}, 6) always choose \textbf{Neighbors}. 

\begin{figure*}[t]
	\centering
	\begin{tabular}{ccc}
		\includegraphics[width=0.5\textwidth]{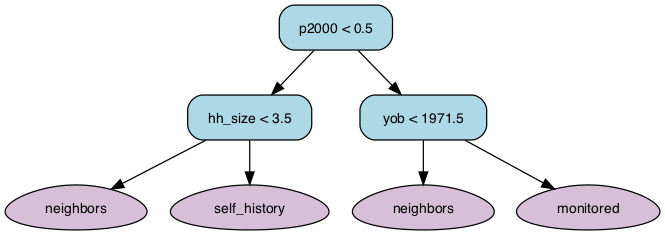} &
		\includegraphics[width=0.5\textwidth]{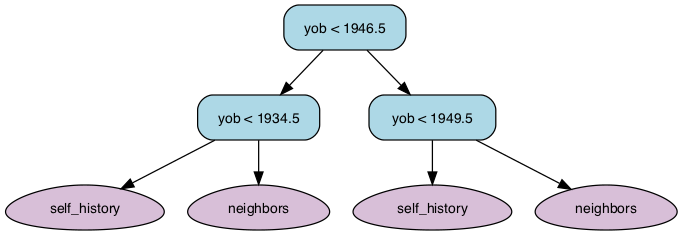} & \\
		(a) & (b)
	\end{tabular} 
	\caption{(a) Tree learned from CAIPWL when applied on the dataset directly. Recall that $p2000$ is a binary variable indicating whether the voter voted in the 2000 primary ($0$ means no vote and $1$ means voted); hh$\_$size represents the household size and yob represents year of birth. This tree has comparable performance to always choosing the \textbf{Neighbors} action. Note that when a voter didn't vote in the 2000 primary and when his or her householid size is greater than or equal to $4$, the tree recommends the \textbf{Self History} action. This makes intuitive sense because a large household size where all the eligible voters' past voting records are shown will presumably generate sufficient pressure.  Finally, for a voter who did vote in the 2000 primary, the tree would recommend a mild action for a young person (age less than or equal to $24$). This tree indicates that, for a certain subpopulation, there is no need to use the most aggressive \textbf{Neighbors} action--an alternative action that is milder can still achieve comparable effect. (b) Tree learned from IPWL when applied on dataset directly. This tree has suboptimal performance and all the variables split on year of birth.}
	\label{fig:tree_from_voting_data_no_injection}
\end{figure*}

To compare the performance, we divide the entire dataset into $5$ folds. Each time we train on the $4$ folds to obtain two depth-$3$ tree policies, one for CAIPWL and one for IPWL. We then use the fifth fold to evaluate the performance of these two depth-$3$ tree policies as well as the performance of the other six policies. We repeat this procedure five times, each time using a different fold as the held-out test data.
We use the AIPW estimator for evaluation: if $\pi$ is a given policy and $F_{test}$ is the set of indices for test data, then the test value is computed by $Q^{test} (\pi) = \sum_{i \in F_{test}} \langle \pi(X_i), \frac{Y_i - \hat{\mu}_{\W_i}(X_i)}{e_{\W_i}} \cdot  \W_i
+
\begin{bmatrix}
\hat{\mu}_{a^1}(X_i) \\
\hat{\mu}_{a^2}(X_i)\\
\hat{\mu}_{a^3}(X_i)\\
\hat{\mu}_{a^4}(X_i)\\
\hat{\mu}_{a^5}(X_i) 
\end{bmatrix} \rangle$, where $e_a$ is known exactly in this case for each action $a$ and each $\hat{\mu}_{a}(\cdot)$ is learned from the rest four folds: we do so because the individual terms $\langle \pi(X_i), \frac{Y_i - \hat{\mu}_{\W_i}(X_i)}{e_{\W_i}} \cdot  \W_i
+
\begin{bmatrix}
\hat{\mu}_{a^1}(X_i) \\
\hat{\mu}_{a^2}(X_i)\\
\hat{\mu}_{a^3}(X_i)\\
\hat{\mu}_{a^4}(X_i)\\
\hat{\mu}_{a^5}(X_i) 
\end{bmatrix} \rangle$ are then \textbf{iid}, and we can perform $t$-test to compare performance difference between two policies. 
Similarly, given a test fold $F_{test}$,  the mean policy value difference between two policies $\pi_1$ and $\pi_2$ can be computed as $Q^{test} (\pi_1) - Q^{test} (\pi_2) =  \sum_{i \in F_{test}} \langle \pi_1(X_i) - \pi_2(X_i), \frac{Y_i - \hat{\mu}_{\W_i}(X_i)}{e_{\W_i}} \cdot  \W_i
+
\begin{bmatrix}
\hat{\mu}_{a^1}(X_i) \\
\hat{\mu}_{a^2}(X_i)\\
\hat{\mu}_{a^3}(X_i)\\
\hat{\mu}_{a^4}(X_i)\\
\hat{\mu}_{a^5}(X_i) 
\end{bmatrix} \rangle$.
Consequently, to test whether $\pi_1$ is significantly different from $\pi_2$, we can 
perform $t$-test on the list of values $\langle \pi_1(X_i) - \pi_2(X_i), \frac{Y_i - \hat{\mu}_{\W_i}(X_i)}{e_{\W_i}} \cdot  \W_i
+
\begin{bmatrix}
\hat{\mu}_{a^1}(X_i) \\
\hat{\mu}_{a^2}(X_i)\\
\hat{\mu}_{a^3}(X_i)\\
\hat{\mu}_{a^4}(X_i)\\
\hat{\mu}_{a^5}(X_i) 
\end{bmatrix} \rangle$, where $i \in F_{test}$. The null hypothesis is that the mean difference is $0$ and the alternative hypothesis is that it isn't. 
Note that since the number of data points is large in this case, estimating $\mu_{a}(\cdot)$ based on the four folds is very close to estimating $\mu_{a}(\cdot)$ using all five folds.
Finally, $Q^{test} (\pi_1) - Q^{test} (\pi_2) $ gives the mean policy value difference on a particular test fold, and we can average all five test folds to yield the average mean policy value difference. 
Similarly, we perform an aggregate $t$-test on all test folds. 
The results are then reported in Table~\ref{table:data}, where each number in the first row shows the mean policy value difference between the policy learned from CAIPWL and the policy in each column. 
In particular, a positive number indicates that the policy learned from CAIPWL does better than the column policy on the test data; a negative number indicates otherwise.
Similarly, the second row shows the mean policy value difference between the policy learned from IPWL and the policy in each column. Further, Figure~\ref{fig:tree_from_voting_data_no_injection} provides one of the trees learned using CAIPWL and IPWL.

Several interesting observations to point out here. 
First, it is clear that out of the 6 policies, always choosing \textbf{Neighbors} is the best policy. This point is confirmed by~\cite{gerber2008social}, which reported that sending the \textbf{Neighbors} letter increases the voter turnout most signficantly. Second, both CAIPWL and IPWL perform better than each of the first five policies, where the performance differences are both significant (all the p-values are $< 0.0001$).
However, IPWL performs signficantly worse than \textbf{Neighbors} while CAIPWL is comparable to \textbf{Neighbors}: the difference is insignificant ($p$-value is 0.47). This in particular implies that CAIPWL is significantly better than IPWL. Third, putting the above two observations together, we note that there is no (significant) heterogenity in this voting data. In this case, CAIPWL is able to learn a policy of comparable performance to the best policy \textbf{Neighbors}, while IPWL failed to do so.
Note that Table~\ref{table:data} provides only one set of five runs on the entire dataset (with a particular random partition into 5 disjoint folds). With a different random partition, the results will vary. However, one thing that is common across different sets of such runs is that CAIPWL is always comparable to the best policy \textbf{Neighbors} (either insignificantly better or insignificantly worse), while IPWL is always significantly worse.
Finally, it is important to point out the computational advantage of the tree-search algorithm we have developed: the training phase each takes in $180002 * \frac{4}{5} = 144000$ data points and the computation finished within a few seconds on a laptop.
In this case, all the features are categorical and thus, as mentioned in the previous section, our algorithm is further capable of taking advantage of this structure to run more efficiently than if all the features are numerical. 

\begin{table}[]
	\begin{tabular}{l|llllll|l}
		\cline{1-8}
		& Random & Nothing & Civic & Monitored & Self History & Neighbors  \\ 
		\cline{1-7}
		Value diff (CAIPWL) &   $0.046^{***}$   & $0.082^{***}$   & $0.061^{***}$    & $0.056^{***}$    & $0.031^{***}$    & $0.00042^{\Delta}$ &  \\ 
		\cline{1-7}
		\cline{1-7}
		Value diff (IPWL)  &   $0.042^{***}$   & $0.078^{***}$   & $0.057^{***}$    & $0.052^{***}$    & $0.027^{***}$    & $-0.0038^{**}$ &  \\ \hline
	\end{tabular}
	\caption{A performance comparison between policies in the rows and policies in the columns for the dataset. Both CAIPWL and IPWL are significantly better than first five column policies. Always choosing Neighbors is the best policy: IPWL is signficantly worse than it and CAIPWL is not significantly better. In other words, there is no heterogenity in this dataset.
		$**$: p-value $< 0.001$. $***$: p-value $< 0.0001$. $\Delta$: p-value $=0.27$. }
	\label{table:data}
\end{table}

\begin{table}[]
	\begin{tabular}{l|llllll|l}
		\cline{1-8}
		& Random & Nothing & Civic & Monitored & Self History & Neighbors  \\ 
		\cline{1-7}
		Value diff (CAIPWL) &   $0.054^{***}$   & $0.085{***}$   & $0.054^{***}$   & $0.049^{***}$ & $0.047^{***}$        & $0.027^{***}$ &  \\ 
		\cline{1-7}
		\cline{1-7}
		Value diff (IPWL)  &   $0.048^{***}$   & $0.079^{***}$   & $0.049^{***}$  & $0.043^{***}$  & $0.041^{***}$       & $0.021^{***}$ &  \\ \hline
	\end{tabular}
	\caption{A performance comparison between policies in the rows and policies in the columns for the dataset with injected heterogeneity. Both CAIPWL and IPWL are significantly better than each of the six column policies. Further, CAIPWL is significantly better than IPWL.}
	\label{table:data2}
\end{table}

\begin{figure*}[t]
	\centering
	\begin{tabular}{ccc}
		\includegraphics[width=0.5\textwidth]{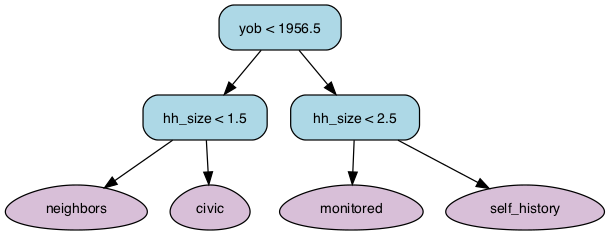} &
		\includegraphics[width=0.5\textwidth]{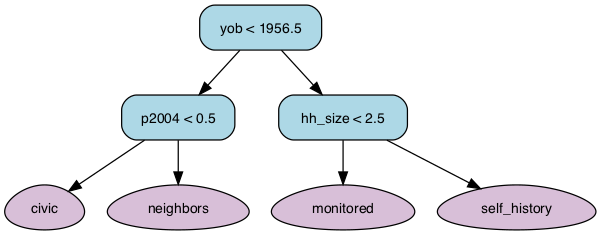} & \\
		(a) & (b)
	\end{tabular} 
	\caption{(a) Tree learned from CAIPWL when applied on dataset with heterogeneity injection. This tree is consistently learned from all five times. (b) Tree learned from IPWL applied on dataset with heterogeneity injection. This is one of the trees learned across five times.}
	\label{fig:tree_from_voting_data}
\end{figure*} 

\subsection{Application of CAIPWL on the Dataset with Injected Heterogenity}
The previous subsection demonstrates that when there is no heterogenity, CAIPWL is able to perform comparably with the best action. Our goal in this subsection is to demonstrate that, if there \textit{were} heterogenity in the voting data, our algorithm CAIPWL would have been able to exploit it to learn a better policy. To do that, we inject a small amount of heterogenity into the data and check whether our policy learning algorithm would be able to identify it.

More specifically, we imagine different subgroups of people respond differently to the different actions.
In particular, each of the following three groups has a slight preference for (i.e. responds more positively to) a different action:
\begin{enumerate}
	\item People who are younger than 50 (i.e. born after 1956, since the data was collected in 2006) and who have small households (no larger than 2 people in the household) prefer \textbf{Monitored}.
	
	\item People who are younger than 50 and who have large households (more than 2 people in the household) prefer \textbf{Self History}.
	
	\item People who are older than (or as old as) 50 and who are not single (more than 1 person in the household) prefer \textbf{Civic}.
\end{enumerate}
Next, we inject these heterogeneous preferences into the dataset.
We keep the signal injection at its bare minimum using the following procedure.
First, for all datapoints whose action in data is \textbf{nothing}, we do not make any modification (this is about 100000 data points). Second, for the datapoints whose action is one of the other four treatments, we apply the following transformation to the outcome variable:
\begin{enumerate}
	\item
	Flip the outcome from $0$ to $1$ w.p. $p$ if action in data matches preferred action.
	\item 
	Flip the outcome from $1$ to $0$ w.p. $p$ if action in data doesn't match preferred action.
\end{enumerate}
Here we set $p = 0.07$, which is a small probability of flipping.
We then apply the policy learning algorithms to this modified dataset. The results are given in Table~\ref{table:data2}. Note that this time, both CAIPWL and IPWL are able to identify the heterogenity and deliver trees that perform significantly better than all six column policies. Note that \textbf{Neighbors} is still the best policy among all the column policies: this is because the injected heterogenity is so weak that it does not completely override the strong effect of \textbf{Neighbors} in the original data.

Another important thing to note is that CAIPWL performs better than IPWL. This is not only seen from Table~\ref{table:data2}, but also made clear from the actual trees learned, as shown in Figure~\ref{fig:tree_from_voting_data}. From Figure~\ref{fig:tree_from_voting_data}, it is clear that CAIPWL has learned the right tree, where all the three subgroups have been assigned to the best action. Note also that \textbf{Neighbors} is still the best action for people who are older than 50 (i.e. born before 1956) and who live in single-person households, because there is no modification for this group of people. We emphasize this tree is consistently learned by CAIPWL across all training folds. On the other, IPWL does not learn the optimal tree, even though it can identify some relevant features in the learned tree. For instance, one tree learned from IPWL is shown in Figure~\ref{fig:tree_from_voting_data} (b), where it has identified household size and age, but has incorrectly split on $p2004$. Across different training folds, IPWL learns different suboptimal trees, where it usually splits on one variable that is not correct. In summary, the above empirical results demonstrate both the important role played by policy learning algorithms in exploiting heterogenity to make better decisions and the superiority of CAIPWL.

 \section{Conclusion and Future Work}

In this paper, we have provided a framework for the multi-action offline policy learning problem, which contributes to the broad landscape of data-driven decision making  by providing guidance for service-decision provisioning using observational data.
We believe much exciting work remains. For instance, in certain settings, one may wish to use other policy classes that are more expressive (such as neural networks), in which case different computational algorithms must be deployed. Additionally, actions can be continuous, in which case a broader framework is needed to handle the infinite-action settings. Another important direction is to apply the framework developed in this paper to different empirical applications. We leave them for future work.

\section{Acknowledgements}

We gratefully acknowledge the support from the Sloan Foundation and ONR grant N00014-17-1-2131.

\theendnotes

\bibliographystyle{ormsv080}
\bibliography{references}
\begin{APPENDIX} {Auxiliary Results and Missing Proofs}
	\section{Auxiliary Results}


We first state two useful concentration inequalities: the first is Bernstein's inequality, and the second is Hoeffding's inequality (see~\cite{boucheron2013concentration}).
\begin{lemma}\label{lem:bern}
	Let $0 < M < \infty$ be a fixed constant and let $X_1, X_2, \dots, X_n$ be independent, zero-mean (real-valued) random variables with $|X_i| \le  M \text{ a.s.} $  and 
	$\sum_{i=1}^n \E[X_i^2] = v$.
	Then for any $t > 0$, the following holds:
	\begin{equation}
	\P\Big[\Big|\sum_{i=1}^n X_i\Big| \ge t\Big] \le 
	2 \exp\Big[\frac{t^2}{2(v + \frac{Mt}{3})}\Big].
	\end{equation}
\end{lemma}

\begin{lemma}\label{lem:hoef}
	Let $X_1, \dots, X_n$ be independent (real-valued) random variables with $X_i \in  [a_i, b_i] \text{ a.s.}$, and let $S_n = \sum_{i=1}^n (X_i - \E[X_i])$,
	then for every $t > 0$, we have:
	$$\P\{|S_n| \ge  t \} \le 2\exp\Big(-\frac{2t^2}{\sum_{i=1}^n (b_i -a_i)^2}\Big).$$
\end{lemma}

Next, we state Talagrand's inequality (see~\cite{talagrand1994}).
\begin{lemma}\label{lem:tal}
	Let $X_1, \dots, X_n$ be independent $\feas$-valued random variables and $\mathcal{F}$ be a class of functions where each $f: \feas \rightarrow \reals$ in $\mathcal{F}$ satisfies $\sup_{x \in \feas} |f(x) |\le 1$. Then:
	\begin{equation}
	\P\Bigg\{\Bigg|\sup_{f \in \mathcal{F}} |\sum_{i=1}^n f(X_i) | - \E\Big[ \sup_{f \in \mathcal{F}} |\sum_{i=1}^n f(X_i) |\Big] \Bigg| \ge t \Bigg\} 
	\le 2 \exp\Big( -\frac{1}{2} t \log(1 + \frac{t}{V})\Big), \forall t > 0,
	\end{equation}
	where $V$ is any number satisfying
	$V \ge \E \Big[\sup_{f \in \mathcal{F}} \sum_{i=1}^n f^2(X_i) \Big]$.
\end{lemma}

As remarked in~\cite{gine2006}, by a further argument that uses randomization and contraction, Talagrand's inequality yields the following bound (which will also be useful later):

\begin{lemma}\label{lem:tal2}
	Let $X_1, \dots, X_n$ be independent $\feas$-valued random variables and $\mathcal{F}$ be a class of functions where $\sup_{x \in \feas} |f(x) |\le U$ for some $U > 0$, and let $Z_i$ be \textbf{iid} Rademacher random variables: $\P[Z_i = 1] = \P[Z_i = -1] = \frac{1}{2}$. We have:
	$$\E\Big[\sup_{f \in \mathcal{F}} \sum_{i=1}^n f^2(X_i) \Big] \le n \sup_{f \in\mathcal{F}} \E[f^2(X_i)] + 8U\E\Big[\sup_{f \in \mathcal{F}}\Big|\sum_{i=1}^n Z_i f(X_i) \Big|\Big].$$
\end{lemma}	

We end this section with a review of the definition of VC dimension. More details can be found in~\cite{mohri2012foundations}.

\begin{definition}
	Given a policy class $\Pi$ of binary-valued functions $\pi : \feas \rightarrow \{1, -1\}$.
	\begin{enumerate}
		\item The growth function $G_{\Pi} : \mathbb{N} \rightarrow \mathbb{N}$ for the policy class $\Pi$ is defined as: $G_{\Pi}(m) = \max_{ \{x_1, \dots, x_m\} \subset \feas} \big|  \{  (\pi(x_1), \pi(x_2), \dots, \pi(x_m)) \mid \pi \in \Pi     \}  \big|$, where $\mathbb{N}$ is the set of natural numbers. 
		\item
		When a set $ \{x_1, \dots, x_m\} $ of $m$ points yields $2^m$ possible different labelings
		$(\pi(x_1), \pi(x_2), \dots, \pi(x_m))$ as $\pi$ ranges over $\Pi$, then we say $ \{x_1, \dots, x_m\} $  is shattered by $\Pi$.
		\item The VC-dimension of $\Pi$ is the largest set of points that can be shattered by $\Pi$:
		$$VC(\Pi) = \max \{m \mid G_{\Pi}(m) = 2^m\}.$$
		If $\Pi$ shatters every finite set of points, then $VC(\Pi) = \infty$.
	\end{enumerate}
\end{definition}

\section{Proof of Lemma~\ref{lem:ipd}}
\setcounter{proofstep}{0}
		
	We first verify the triangle inequality as follows:
	$\forall \pi_1, \pi_2, \pi_3,$
	\begin{equation}
	\begin{split}
	\D(\pi_1, \pi_2) &= \sqrt{\frac{\sum_{i=1}^n |\langle \gamma_i,\pi_1(x_i) - \pi_3(x_i) + \pi_3(x_i)- \pi_2(x_i)\rangle|^2}{\sup_{\pi_a, \pi_b}\sum_{i=1}^n |\langle \gamma_i,\pi_a(x_i) - \pi_b(x_i)\rangle|^2}} \\
	& =  \sqrt{\frac{\sum_{i=1}^n |\langle \gamma_i,\pi_1(x_i) - \pi_3(x_i)\rangle + \langle \gamma_i, \pi_3(x_i)- \pi_2(x_i)\rangle|^2}{\sup_{\pi_a, \pi_b}\sum_{i=1}^n |\langle \gamma_i,\pi_a(x_i) - \pi_b(x_i)\rangle|^2}}  \\
	&\le  \sqrt{\frac{\sum_{i=1}^n |\langle \gamma_i,\pi_1(x_i) - \pi_3(x_i)\rangle|^2}{\sup_{\pi_a, \pi_b}\sum_{i=1}^n |\langle \gamma_i,\pi_a(x_i) - \pi_b(x_i)\rangle|^2}} +
	\sqrt{\frac{\sum_{i=1}^n |\langle \gamma_i, \pi_3(x_i)- \pi_2(x_i)\rangle|^2}{\sup_{\pi_a, \pi_b}\sum_{i=1}^n |\langle \gamma_i,\pi_a(x_i) - \pi_b(x_i)\rangle|^2}} \\
	& = \D(\pi_1, \pi_3) + \D(\pi_3, \pi_2),
	\end{split}
	\end{equation}
	where the last inequality follows from squaring both sides of the second and third lines and noticing:
	\begin{equation}
	\begin{split}
	&2\sum_{i=1}^n\langle \gamma_i,\pi_1(x_i) - \pi_3(x_i)\rangle \langle \gamma_i,\pi_3(x_i) - \pi_2(x_i)\rangle 
	\le 2\sum_{i=1}^n|\langle \gamma_i,\pi_1(x_i) - \pi_3(x_i)\rangle|\cdot| \langle \gamma_i,\pi_3(x_i) - \pi_2(x_i)\rangle| \\
	& = 2\sqrt{\Big(\sum_{i=1}^n|\langle \gamma_i,\pi_1(x_i) - \pi_3(x_i)\rangle|^2 \Big)
		\Big(\sum_{i=1}^2\cdot| \langle \gamma_i,\pi_3(x_i) - \pi_2(x_i)\rangle|^2\Big)},\\
	\end{split}
	\end{equation}
	where the last inequality follows from Cauchy-Schwartz.
	
	Next, to prove the second statement, let $K = N_H(\epsilon^2, \Pi)$. Without loss of generality, we can assume $K < \infty$, otherwise, the above inequality automatically holds.
	Fix any $n$ points $\{x_1, \dots, x_n\}$.
	Denote by $\{\pi_1, \dots, \pi_K\}$ the set of $K$ policies that $\epsilon^2$-cover $\Pi$. This means that for any $\pi \in \Pi$, there exists a $\pi_j$, such that (here we write $H_M(\cdot, \cdot)$ to emphasize the explicit dependence of the Hamming distance on the sample size): 
	\begin{equation}\label{eq:crucial}
	\forall M> 0, \forall \{\tilde{x}_1, \dots, \tilde{x}_M\},
	H_M(\pi, \pi_j) =  \frac{1}{M} \sum_{i=1}^m \mathbf{1}_{\{\pi(\tilde{x}_i) \neq \pi_j(\tilde{x}_i)\}} \le \epsilon^2.
	\end{equation}
	
	Pick $M = \sum_{i=1}^n \ceil{\frac{m |\langle \gamma_i,\pi(x_i) - \pi_j(x_i)\rangle|^2}{\sup_{\pi_a, \pi_b}\sum_{i=1}^n |\langle \gamma_i,\pi_a(x_i) - \pi_b(x_i)\rangle|^2}}$ (where $m$ is some positive integer) and  
	$\{\tilde{x}_1, \dots, \tilde{x}_M\} = 
	\{x_1, \dots, x_1, x_2, \dots, x_2, \dots, x_n, \dots, x_n\}$,
	where $x_1$ appears $\ceil{\frac{m |\langle \gamma_i,\pi(x_1) - \pi_j(x_1)\rangle|^2}{\sup_{\pi_a, \pi_b}\sum_{i=1}^n |\langle \gamma_i,\pi_a(x_i) - \pi_b(x_i)\rangle|^2}}$ times, $x_2$ appears $\ceil{\frac{m |\langle \gamma_i,\pi(x_2) - \pi_j(x_2)\rangle|^2}{\sup_{\pi_a, \pi_b}\sum_{i=1}^n |\langle \gamma_i,\pi_a(x_i) - \pi_b(x_i)\rangle|^2}}$ times and so on.
	Per the definition of $M$, we have: 
	\begin{equation}
	\begin{split}
	&M = \sum_{i=1}^n \ceil{\frac{m |\langle \gamma_i,\pi(x_i) - \pi_j(x_i)\rangle|^2}{\sup_{\pi_a, \pi_b}\sum_{i=1}^n |\langle \gamma_i,\pi_a(x_i) - \pi_b(x_i)\rangle|^2}}
	\le  \sum_{i=1}^n ( \frac{m |\langle \gamma_i,\pi(x_i) - \pi_j(x_i)\rangle|^2}{\sup_{\pi_a, \pi_b}\sum_{i=1}^n |\langle \gamma_i,\pi_a(x_i) - \pi_b(x_i)\rangle|^2}+ 1 ) \\
	&=\frac{ m\sum_{i=1}^n |\langle \gamma_i,\pi(x_i) - \pi_j(x_i)\rangle|^2}{\sup_{\pi_a, \pi_b}\sum_{i=1}^n |\langle \gamma_i,\pi_a(x_i) - \pi_b(x_i)\rangle|^2}+ n \le  m +n.
	\end{split}
	\end{equation}
	Further, since each $x_i$ appears $\ceil{\frac{m |\langle \gamma_i,\pi(x_i) - \pi_j(x_i)\rangle|^2}{\sup_{\pi_a, \pi_b}\sum_{i=1}^n |\langle \gamma_i,\pi_a(x_i) - \pi_b(x_i)\rangle|^2}}$  times, we have:
	\begin{align*}
	H_M(\pi, \pi_i) &= \frac{1}{M} \sum_{i=1}^M \mathbf{1}_{\{\pi(\tilde{x}_i) \neq \pi_j(\tilde{x}_i)\}} = \frac{1}{M} \sum_{i=1}^n \ceil{\frac{m |\langle \gamma_i,\pi(x_i) - \pi_j(x_i)\rangle|^2}{\sup_{\pi_a, \pi_b}\sum_{i=1}^n |\langle \gamma_i,\pi_a(x_i) - \pi_b(x_i)\rangle|^2}} \mathbf{1}_{\{\pi(x_i) \neq \pi_j(x_i)\}} \\
	&\ge 
	\frac{1}{m+n}\sum_{j=1}^n \frac{m |\langle \gamma_i,\pi(x_i) - \pi_j(x_i)\rangle|^2}{\sup_{\pi_a, \pi_b}\sum_{i=1}^n |\langle \gamma_i,\pi_a(x_i) - \pi_b(x_i)\rangle|^2} \mathbf{1}_{\{\pi(x_i) \neq \pi_j(x_i)\}} \\
	& = 
	\frac{1}{m+n}\sum_{j=1}^n \frac{m |\langle \gamma_i, \{\pi(x_i) - \pi_j(x_i)\} \mathbf{1}_{\{\pi(x_i) \neq \pi_j(x_i)\}} \rangle|^2}{\sup_{\pi_a, \pi_b}\sum_{i=1}^n |\langle \gamma_i, \pi_a(x_i) - \pi_b(x_i)\rangle|^2} \\
	& = 
	\frac{1}{m+n}\sum_{j=1}^n \frac{m |\langle \gamma_i, \pi(x_i) - \pi_j(x_i) \rangle|^2}{\sup_{\pi_a, \pi_b}\sum_{i=1}^n |\langle \gamma_i, \pi_a(x_i) - \pi_b(x_i)\rangle|^2} \\
	& =\frac{m}{m+n}\frac{\sum_{i=1}^n |\langle \gamma_i,\pi_1(x_i) - \pi_2(x_i)\rangle|^2}{\sup_{\pi_a, \pi_b}\sum_{i=1}^n |\langle \gamma_i, \pi_a(x_i) - \pi_b(x_i)\rangle|^2}
	= \frac{m}{m+n}\D^2(\pi, \pi_i).
	\end{align*}
	Letting $m \rightarrow \infty$ (and hence $M \rightarrow \infty$) yields:
	$\lim_{m \rightarrow \infty} H_M(\pi, \pi_j) \ge \D^2(\pi, \pi_j)$,
	where we note that $n$ is fixed and $\D(\pi, \pi_j)$ is still the inner product distance between $\pi$ and $\pi_i$ with respect to the original $n$ points $\{x_1, \dots, x_n\}$.
	Further, since Equation~\eqref{eq:crucial} holds for any $m$, we then have:
	$$\epsilon^2 \ge \lim_{m \rightarrow \infty} H_M(\pi, \pi_j)  \ge \D^2(\pi, \pi_j) .$$
	This immediately implies 
	$ \D(\pi, \pi_j) \le \epsilon$. 
	Consequently, the above argument establishes that for any $\pi \in \Pi$,
	there exists a $\pi_j\in \{\pi_1, \dots, \pi_K\}$, such that
	$ \D(\pi, \pi_j) \le \epsilon$, and therefore $N_{\D}(\epsilon, \Pi, \{x_1, \dots, x_n\}) \le K = N_H(\epsilon^2, \Pi)$.

\section{Proof of Theorem~\ref{thm:rad}}

\setcounter{proofstep}{0}

\begin{proofstep}{Policy approximations}\label{step:2}
	
We give an explicit construction of the sequence of approximation operators that satisfies these four properties.
Set $\epsilon_j = \frac{1}{2^j}$ and let $S_0, S_1, S_2, \dots, S_J$ be a sequence of policy classes (understood to be subclasses of $\Pi$) such that $S_j$ $\epsilon_j$-cover $\Pi$ under the inner product distance: 
$$\forall \pi \in \Pi, \exists \pi^{\prime} \in S_j, 
\D(\pi, \pi^{\prime}) \le \epsilon_j.$$ 

Note that by definition of the covering number (under the inner product distance), we can choose the $j$-th policy class $S_j$ such that $|S_j| = \N_{\D}(2^{-j},  \Pi, \{X_1, \dots, X_n\})$.
Note that in particular, $|S_0| = 1$, since any single policy is enough to $1$-cover all policies
in $\Pi$ (recall that inner product distance between any two policies in $\Pi$ is never more than $1$).

Next, we use the following backward selection scheme to define $A_j$'s.
For each $\pi \in \Pi$, define:
$$A_J(\pi) = \arg\min_{\pi^{\prime} \in S_J}  \D(\pi, \pi^{\prime}).$$
Further, for each $0 \le j < J$ and each $\pi \in \Pi$, inductively define:
\begin{equation}\label{eqref:prox}
A_j(\pi) = \arg\min_{\pi^{\prime} \in S_j}  \D(A_{j+1}(\pi), \pi^{\prime}).
\end{equation}
We next check in order that each of the four properties is satisfied. 
\begin{enumerate}
	\item Pick any $\pi \in \Pi$. By construction, $ \exists \pi^{\prime} \in S_J, 
	\D(\pi, \pi^{\prime}) \le \epsilon_J$. By the definition of $A_J$, we have 
	$\D(\pi, A_J(\pi)) \le 	\D(\pi, \pi^{\prime}) \le \epsilon_J = \frac{1}{2^J}$. Taking maximum over all $\pi \in \Pi$ verifies Property 1.
	
	\item Since $A_j(\pi) = \arg\min_{\pi^{\prime} \in S_j} \D(A_{j+1}(\pi), \pi^{\prime})$ by construction, $A_j(\pi) \in S_j$ for every $\pi \in \Pi$. Consequently, $|\{A_j(\pi) \mid \pi \in \Pi \} | \le |S_j| =   \N_{\D}(2^{-j},  \Pi, \{X_1, \dots, X_n\})$.
	
	\item Since $\D$ satisfies the triangle inequality (by Lemma~\ref{lem:ipd}), we have:
	$\max_{\pi \in \Pi} \D(A_j(\pi), A_{j+1} (\pi)) \le \max_{\pi \in \Pi} \Big\{\D(A_j(\pi), \pi) + \D(A_{j+1}(\pi), \pi)\Big\}  \le  \max_{\pi \in \Pi} \D(A_j(\pi), \pi) + \max_{\pi \in \Pi} \D(A_{j+1}(\pi), \pi) \le 2^{-j} + 2^{-(j+1)} \le 2^{-(j-1)}$.
	
	\item If $A_{j^{\prime}} (\pi) = A_{j^{\prime}} (\tilde{\pi})$, then by construction given in Equation~\eqref{eqref:prox}, we have:
	\begin{equation}
	A_{j^\prime -1} (\pi) = \arg\min_{\pi^{\prime} \in S_{j^\prime}}  \D(A_{j^\prime}(\pi), \pi^{\prime}) = \arg\min_{\pi^{\prime} \in S_{j^\prime}}  \D(A_{j^\prime}(\tilde{\pi}), \pi^{\prime}) = A_{j^\prime -1}(\tilde{\pi}).
	\end{equation}
	Consequently, by backward induction, it then follows that $A_{j} (\pi) = A_{j} (\tilde{\pi})$.
	Therefore, 
	$$|\{(A_j(\pi), A_{j^{\prime}}(\pi)) \mid \pi \in \Pi\}| = |\{A_{j^{\prime}}(\pi) \mid \pi \in \Pi\}|\le \N_{\D}(2^{-j^{\prime}},  \Pi, \{X_1, \dots, X_n\}).$$
\end{enumerate}
\end{proofstep}

\begin{proofstep}{Chaining with concentration inequalities in the negligible regime}\label{step:3}
	
We prove Statement 1 and Statement 2 in turn.	
	\begin{enumerate}
		\item 
		To see this, note that by Cauchy-Schwartz inequality, we have:
		\begin{equation}
		\begin{split}
		&\sup_{\pi \in \Pi}\Big|\frac{1}{n} \sum_{i=1}^n Z_i \langle \Gamma_i, \pi(X_i) - A_{J} (\pi)(X_i)\rangle\Big| \le
		\sup_{\pi \in \Pi}\sqrt{	\frac{1}{n}\sum_{i=1}^n |\langle \Gamma_i, \pi(X_i) - A_{J} (\pi)(X_i)\rangle|^2} \\
		& =  \sup_{\pi \in \Pi}\sqrt{\frac{\sum_{i=1}^n |\langle \Gamma_i, \pi(X_i) - A_{J} (\pi)(X_i)\rangle|^2}{\sup_{\pi_a, \pi_b}\sum_{i=1}^n |\langle \Gamma_i,\pi_a(x_i) - \pi_b(x_i)\rangle|^2}}
		\sqrt{\frac{\sup_{\pi_a, \pi_b}\sum_{i=1}^n |\langle \Gamma_i,\pi_a(X_i) - \pi_b(X_i)\rangle|^2}{n}} \\
		&= \sup_{\pi \in \Pi}\D(\pi, A_J(\pi)) \sqrt{\frac{\sup_{\pi_a, \pi_b}\sum_{i=1}^n |\langle \Gamma_i,\pi_a(X_i) - \pi_b(X_i)\rangle|^2}{n}}\\
		& \le \sup_{\pi \in \Pi}\D(\pi, A_J(\pi)) \sqrt{\frac{\sup_{\pi_a, \pi_b} \sum_{i=1}^n \|\Gamma_i\|^2 \|\pi_a(X_i) - \pi_b(X_i)\|_2^2}{n}} \\
		& \le \sup_{\pi \in \Pi}\D(\pi, A_J(\pi)) \sqrt{\frac{\sum_{i=1}^n \sup_{\pi_a, \pi_b} \|\Gamma_i\|^2 \|\pi_a(X_i) - \pi_b(X_i)\|_2^2}{n}} \le   \sqrt{2} \sup_{\pi \in \Pi}\D(\pi, A_J(\pi)) \sqrt{\frac{\sum_{i=1}^n \|\Gamma_i\|^2}{n}}\\
		&\le  \sqrt{2}\cdot2^{-J}\sqrt{\frac{{\sum_{i=1}^n \|\Gamma_i\|^2}}{n}}   = 
		\sqrt{2} \cdot 2^{-\ceil{\log_2(n) (1 -\omega)}}\sqrt{\frac{{\sum_{i=1}^n \|\Gamma_i\|^2}}{n}}  
		\le 	\frac{ \sqrt{2}}{n^{1-w}} \sqrt{\frac{\sum_{i=1}^n \|\Gamma_i\|^2}{n}},
		\end{split}
		\end{equation}
		where the first inequality follows from $Z_i^2 = 1$ and the last inequality follows from $\D(\pi, A_J(\pi))  \le 2^{-J},\forall \pi \in \Pi$.
		Consequently, we have:
		\begin{equation}
		\begin{split}
		& \sqrt{n} \sup_{\pi \in \Pi}\frac{1}{n} \sum_{i=1}^n Z_i \langle\Gamma_i, \pi(X_i) - A_J(\pi) (X_i)\rangle \le \sqrt{n} \frac{ \sqrt{2}}{n^{1-w}} \sqrt{\frac{\sum_{i=1}^n \|\Gamma_i\|^2}{n}}\\
		& = \frac{ \sqrt{2}}{n^{0.5-w}} \sqrt{\frac{\sum_{i=1}^n \|\Gamma_i\|^2}{n}} \le \frac{ \sqrt{2}}{n^{0.5-w}} \sqrt{\frac{nU^2}{n}} = U\frac{ \sqrt{2}}{n^{0.5-w}},
		\end{split}
		\end{equation}
		where $U$ is an upper bound on $ \|\Gamma_i\|$ (sinc it's bounded).
		Consequently, $\sqrt{n} \E\Big[\sup_{\pi \in \Pi}\frac{1}{n} \sum_{i=1}^n Z_i \langle\Gamma_i, \pi(X_i) - A_J(\pi) (X_i)\rangle\Big] = O(\frac{ 1}{n^{0.5-w}})$, which then immediately implies
		$\lim_{n \rightarrow \infty} \sqrt{n} \E\Big[\sup_{\pi \in \Pi}\frac{1}{n} \sum_{i=1}^n Z_i \langle\Gamma_i, \pi(X_i) - A_J(\pi) (X_i)\rangle\Big] = 0.$
		
		\item 
		Conditioned on $\{X_i, \Gamma_i\}_{i=1}^n$, the random variables $\{Z_i \langle \Gamma_i, A_{\jlow}(\pi)(X_i) - A_{J} (\pi)(X_i)\rangle\}_{i=1}^n$ are independent and zero-mean (since $Z_i$'s are Rademacher random variables). Further, each $Z_i \langle \Gamma_i, A_{\jlow}(\pi)(X_i) - A_{J} (\pi)(X_i)\rangle$ is bounded between $a_i = -\Big|\langle \Gamma_i, A_{\jlow}(\pi)(X_i) - A_{J} (\pi)(X_i)\rangle\Big|$ and $b_i = \Big|\langle \Gamma_i, A_{\jlow}(\pi)(X_i) - A_{J} (\pi)(X_i)\rangle\Big|$.
		
		By the definition of inner product distance, we have:
		$\D(A_{\jlow}(\pi), A_{J} (\pi)) =  \sqrt{\frac{2\sum_{i=1}^n |\langle \Gamma_i,A_{\jlow}(\pi)(X_i) - A_{J} (\pi)(X_i)\rangle|^2}{\sup_{\pi_a, \pi_b}\sum_{i=1}^n |\langle \Gamma_i,\pi_a(X_i) - \pi_b(X_i)\rangle|^2}},$ which, upon rearrangement, yields:
		$$\D^2(A_{\jlow}(\pi), A_{J} (\pi)) \sup_{\pi_a, \pi_b}\sum_{i=1}^n |\langle \Gamma_i,\pi_a(X_i) - \pi_b(X_i)\rangle|^2= \sum_{i=1}^n 2|\langle \Gamma_i,A_{\jlow}(\pi)(X_i) - A_{J} (\pi)(X_i)\rangle|^2.$$ 
		
		We are now ready to apply Hoeffding's inequality:
		\begin{equation}\label{eq:hoeff}
		\begin{split}
		&\mathbf{P}\Big[\Big| \sum_{i=1}^n Z_i \langle \Gamma_i, A_{\jlow}(\pi)(X_i) - A_{J} (\pi)(X_i)\rangle\Big| \ge t\Big] \le 2\exp\Big(- \frac{2t^2}{\sum_{i=1}^n(b_i-a_i)^2} \Big) \\
		&\le  2\exp\Big(-\frac{t^2}{2\sum_{i=1}^n\Big|\langle \Gamma_i, A_{\jlow}(\pi)(X_i) - A_{J} (\pi)(X_i)\rangle\Big|^2}\Big).
		\end{split}
		\end{equation}
		Setting $t =  a 2^{3 - \jlow} \sqrt{\sum_{i=1}^n \|\Gamma_i\|^2}$ (for some $a > 0$), we have:
		\begin{equation}\label{eq:hoeff1}
		\begin{split}
		& \mathbf{P}\Big[\Big| \frac{1}{\sqrt{n}}\sum_{i=1}^n Z_i \langle \Gamma_i, A_{\jlow}(\pi)(X_i) - A_{J} (\pi)(X_i)\rangle\Big| \ge a 2^{3 - \jlow} \sqrt{\frac{\sum_{i=1}^n \|\Gamma_i\|^2}{n}} \Big]\\ &=\mathbf{P}\Big[\Big| \sum_{i=1}^n Z_i \langle \Gamma_i, A_{\jlow}(\pi)(X_i) - A_{J} (\pi)(X_i)\rangle\Big| \ge t\Big]\le 2\exp\Big(-\frac{t^2}{2\sum_{i=1}^n\Big|\langle \Gamma_i, A_{\jlow}(\pi)(X_i) - A_{J} (\pi)(X_i)\rangle\Big|^2}\Big) \\
		&\le  2\exp\Bigg(-\frac{ a^2 4^{3 - \jlow}   }{2\frac{\sum_{i=1}^n\Big|\langle \Gamma_i, A_{\jlow}(\pi)(X_i) - A_{J} (\pi)(X_i)\rangle\Big|^2}{\sum_{i=1}^n \|\Gamma_i\|^2}}\Bigg) \\
		&	=2\exp\Bigg(-\frac{ a^2 4^{3 - \jlow}   }{4\frac{\sum_{i=1}^n\Big|\langle \Gamma_i, A_{\jlow}(\pi)(X_i) - A_{J} (\pi)(X_i)\rangle\Big|^2}{\sup_{\pi_a, \pi_b \in \Pi}\sum_{i=1}^n |\langle \Gamma_i,\pi_a(X_i) - \pi_b(X_i)\rangle|^2}   \frac{\frac{1}{2}\sup_{\pi_a, \pi_b \in \Pi}\sum_{i=1}^n |\langle \Gamma_i,\pi_a(X_i) - \pi_b(X_i)\rangle|^2}{\sum_{i=1}^n \|\Gamma_i\|^2}}\Bigg)\\	
		& \le 2\exp\Bigg(-\frac{ a^2 4^{3 - \jlow}   }{4\frac{\sum_{i=1}^n\Big|\langle \Gamma_i, A_{\jlow}(\pi)(X_i) - A_{J} (\pi)(X_i)\rangle\Big|^2}{\sup_{\pi_a, \pi_b \in \Pi}\sum_{i=1}^n |\langle \Gamma_i,\pi_a(X_i) - \pi_b(X_i)\rangle|^2}   }\Bigg) = 2\exp\Big(-\frac{ a^2 4^{3 - \jlow}   }{4   \D^2(A_{\jlow}(\pi), A_{J} (\pi))}\Big) \\
		& \le 2\exp\Big(-\frac{ a^2 4^{3 - \jlow}   }{4   \{\sum_{j =\jlow}^{J-1} \D(A_{j}(\pi), A_{j+1} (\pi))\}^2}\Big) 
		= 2\exp\Big(-\frac{ a^2 4^{3 - \jlow}   }{4   \{\sum_{j =\jlow}^{J-1} 2^{-(j-1)}\}^2}\Big)  \\
		&\le  2\exp\Big(-\frac{ a^2 4^{3 - \jlow}   }{4^{3 - \jlow}}\Big) = 2\exp(-a^2).
		\end{split}
		\end{equation}
		
		Equation~\eqref{eq:hoeff1} holds for any policy $\pi \in \Pi$. By a union bound, we have:
		\begin{equation}\label{eq:hoeff2}
		\begin{split}
		& \mathbf{P}\Bigg[\sup_{\pi \in \Pi} \Bigg\{\Big| \frac{1}{\sqrt{n}}\sum_{i=1}^n Z_i \langle \Gamma_i, A_{\jlow}(\pi)(X_i) - A_{J} (\pi)(X_i)\rangle\Big| \Bigg\}\ge a 2^{3 - \jlow} \sqrt{\frac{\sum_{i=1}^n \|\Gamma_i\|^2}{n}} \Bigg]\\ 
		& \le 2 \Big|\{(A_{\jlow}(\pi), A_J(\pi)) \mid \pi \in \Pi \}\Big| \exp(-a^2) \le 2  \N_{\D}(2^{-J},  \Pi, \{X_1, \dots, X_n\}) \exp(-a^2) \\
		& \le 2  N_{H}(2^{-2J},  \Pi) \exp(-a^2) 
		\le 2C \exp(2^{2J\omega})\exp(-a^2)  \le 2C \exp(2^{2\omega(1-\omega)\log_2(n) } -a^2),
		\end{split}
		\end{equation}
		where the second inequality follows from Property 4 in Step 2, the third inequality
		folows from Lemma~\ref{lem:ipd}, the fourth inequality follows from Assumption~\ref{assump:1}
		and the last inequality follows from $J = \ceil{(1-\omega)\log_2(n)} \le (1-\omega)\log_2(n) + 1$ (and the term $\exp(2^{2\omega})$ is absorbed into the constant $C$).
		Next, set $a = \frac{2^{\jlow}}{\sqrt{\log n \frac{\sum_{i=1}^n \|\Gamma_i\|^2}{n}}}$, we have:
		
		\begin{equation}\label{eq:hoeff2}
		\begin{split}
		& \mathbf{P}\Bigg[\sup_{\pi \in \Pi} \Bigg\{\Big| \frac{1}{\sqrt{n}}\sum_{i=1}^n Z_i \langle \Gamma_i, A_{\jlow}(\pi)(X_i) - A_{J} (\pi)(X_i)\rangle\Big| \Bigg\}\ge \frac{8}{\sqrt{\log n}}\Bigg]  \le 2C \exp(2^{2\omega(1-\omega)\log_2(n) } -a^2) \\
		&= 2C \exp\Bigg(2^{2\omega(1-\omega)\log_2(n) } - 
		\frac{2^{(1-\omega)\log_2(n) }}{\frac{\sum_{i=1}^n \|\Gamma_i\|^2}{n}\log n}\Bigg) 
		= 	2C \exp\Bigg(2^{2\omega(1-\omega)\log_2(n) } - 
		\frac{2^{(1-2\omega + 2\omega)(1-\omega)\log_2(n) }}{\frac{\sum_{i=1}^n \|\Gamma_i\|^2}{n}\log n}\Bigg)\\ 
		&=  2C \exp\Bigg(2^{2\omega(1-\omega)\log_2(n) }
		\Big\{ 1- \frac{2^{(1-2\omega)(1-\omega)\log_2(n)}}{\frac{\sum_{i=1}^n \|\Gamma_i\|^2}{n}\log n}\Big\}\Bigg) = 2C \exp\Bigg(2^{2\omega(1-\omega)\log_2(n) }
		\Big\{ 1- \frac{n^{(1-2\omega)(1-\omega)}}{\frac{\sum_{i=1}^n \|\Gamma_i\|^2}{n}\log n}\Big\}\Bigg)\\
		&=2C \exp\Bigg(-n^{2\omega(1-\omega) }
		\Big\{ \frac{n^{(1-2\omega)(1-\omega)}}{\frac{\sum_{i=1}^n \|\Gamma_i\|^2}{n}\log n} - 1\Big\}\Bigg)
		\le 2C \exp\Bigg(-n^{2\omega(1-\omega) }
		\Big\{ \frac{n^{(1-2\omega)(1-\omega)}}{U^2\log n} - 1\Big\}\Bigg),
		\end{split}
		\end{equation}
		where again $U$ is an upper bound of $\|\Gamma_i\|$.
		Since $\omega < \frac{1}{2}$ by Assumption~\ref{assump:1}, 
		$\lim_{n \rightarrow \infty} \frac{n^{(1-2\omega)(1-\omega)}}{U^2\log n} = \infty$.
		This means for all large $n$, with probability at least $1- 2C \exp\Big(-n^{2\omega(1-\omega) }\Big)$,
		$\sup_{\pi \in \Pi} \Bigg\{\Big| \frac{1}{\sqrt{n}}\sum_{i=1}^n Z_i \langle \Gamma_i, A_{\jlow}(\pi)(X_i) - A_{J} (\pi)(X_i)\rangle\Big| \Bigg\}\le \frac{8}{\sqrt{\log n}}$, therefore immediately implying:
		$\lim_{n \rightarrow \infty} \sqrt{n} \E\Big[\sup_{\pi \in \Pi} \Big|\frac{1}{n} \sum_{i=1}^n Z_i \langle\Gamma_i, A_J(\pi)(X_i) - A_{\jlow}(\pi) (X_i)\rangle\Big|\Big] = 0.$
	\end{enumerate}
\end{proofstep}

\begin{proofstep}{Chaining with concentration inequalities in the effective regime}\label{step:4}
	We start by noting that:
	$ \mathcal{R}_n(\PiD)
	= \E\Big[\sup_{\pi_a, \pi_b \in \Pi}\frac{1}{n} \Big|\sum_{i=1}^n Z_i \langle\Gamma_i, \pi_a(X_i) - \pi_b(X_i)\rangle \Big|\Big] \\
	 = \E\Big[\sup_{\pi_a, \pi_b \in \Pi}\frac{1}{n} \Big|\sum_{i=1}^n Z_i \langle\Gamma_i, \{\pi_a(X_i) - A_J(\pi_a)(X_i)\} 
	-\{\pi_b(X_i) - A_J(\pi_b)(X_i)\}\rangle \Big|\Big] \\
	 + \E\Big[\sup_{\pi_a, \pi_b \in \Pi}\frac{1}{n} \Big|\sum_{i=1}^n Z_i \langle\Gamma_i, \{A_J(\pi_a)(X_i) - A_{\jlow}(\pi_a)(X_i)\} 
	-\{A_J(\pi_b)(X_i) - A_{\jlow}(\pi_b)(X_i)\}\rangle \Big|\Big] \\
	+ \E\Big[\sup_{\pi_a, \pi_b \in \Pi}\frac{1}{n} \Big|\sum_{i=1}^n Z_i \langle\Gamma_i, \sum_{j=1}^{\jlow}\{A_j(\pi_a)(X_i) - A_{j-1}(\pi_a)(X_i)\} 
	-\sum_{j=1}^{\jlow}\{A_j(\pi_b)(X_i) - A_{j-1}(\pi_b)(X_i)\}\rangle \Big|\Big]\\
	= 2\E\Big[\sup_{\pi \in \Pi}\frac{1}{n} \Big|\sum_{i=1}^n Z_i \langle\Gamma_i, \sum_{j=1}^{\jlow} \Big\{A_j(\pi)(X_i) - A_{j-1}(\pi)(X_i)\Big\}\rangle\Big|\Big] + o(\frac{1}{\sqrt{n}}).
	.$
	
	Next, for each $j \in \{1,\dots, \jlow \}$, setting $t_j = a_j 2^{2 - j} \sqrt{\sup_{\pi_a, \pi_b}\sum_{i=1}^n |\langle \Gamma_i, \pi_a(X_i) - \pi_b(X_i)\rangle|^2}$ and applying Hoeffding's inequality:
	\begin{equation}\label{eq:hoeff3}
	\begin{split}
	 &\mathbf{P}\Bigg[\Big| \frac{1}{\sqrt{n}}\sum_{i=1}^n Z_i \langle \Gamma_i, A_{j}(\pi)(X_i) - A_{j-1} (\pi)(X_i)\rangle\Big| \ge a_j 2^{2 - j} \sqrt{\frac{ \sup_{\pi_a, \pi_b \in \Pi}\sum_{i=1}^n |\langle \Gamma_i, \pi_a(X_i) - \pi_b(X_i)\rangle|^2}{n}} \Bigg]\\ &=\mathbf{P}\Big[\Big| \sum_{i=1}^n Z_i \langle \Gamma_i, A_{j}(\pi)(X_i) - A_{j-1} (\pi)(X_i)\rangle\Big| \ge t_j\Big]\le 2\exp\Big(-\frac{t_j^2}{2\sum_{i=1}^n\Big|\langle \Gamma_i, A_{\jlow}(\pi)(X_i) - A_{J} (\pi)(X_i)\rangle\Big|^2}\Big) \\
	&\le  2\exp\Bigg(-\frac{ a_j^2 4^{2 - j}   }{2\frac{\sum_{i=1}^n\Big|\langle \Gamma_i, A_{\jlow}(\pi)(X_i) - A_{J} (\pi)(X_i)\rangle\Big|^2}{\sup_{\pi_a, \pi_b \in \Pi}\sum_{i=1}^n |\langle \Gamma_i, \pi_a(X_i) - \pi_b(X_i)\rangle|^2}}\Bigg)  = 2\exp\Big(-\frac{ a_j^2 4^{2 - j}   }{2   \D^2(A_{j}(\pi), A_{j-1} (\pi))}\Big) \\
	 &\le 2\exp\Big(-\frac{ a_j^2 4^{2 - j}   }{2  \cdot 4^{-(j-2)}}\Big) =  2\exp\Big(-\frac{ a_j^2 }{2}\Big),
	\end{split}
	\end{equation}
	where the last inequality follows from Property 3 in Step 1. 
	For the rest of this step, we denote for notational convenience $M(\Pi) \triangleq  \sup_{\pi_a, \pi_b}\sum_{i=1}^n |\langle \Gamma_i, \pi_a(X_i) - \pi_b(X_i)\rangle|^2$, as this term will be repeatedly used.
Setting $a_j^2 = 2\log \Big(\frac{2j^2}{\delta} \N_H(4^{-j}, \Pi)\Big)$, we then apply a union bound to obtain:
	 
	 $\mathbf{P}\Big[\sup_{\pi \in \Pi}\Big| \frac{1}{\sqrt{n}}\sum_{i=1}^n Z_i \langle \Gamma_i, A_{j}(\pi)(X_i) - A_{j-1} (\pi)(X_i)\rangle\Big| \ge a_j 2^{2 - j} \sqrt{\frac{ M(\Pi)}{n}} \Big]\\
	 \le  2  \Bigg|\Bigg\{\Big(A_{j}(\pi), A_{j-1}(\pi)\Big) \mid \pi \in \Pi \Bigg\}\Bigg|\exp\Big(-\frac{ a_j^2 }{2}\Big)\le 2  \N_{\D}(2^{-j},  \Pi, \{X_1, \dots, X_n\})\exp\Big(-\frac{ a_j^2 }{2}\Big) \\
	 \le 2 \N_H(4^{-j}, \Pi)\exp\Big(-\frac{ a_j^2 }{2}\Big) = 2 \N_H(4^{-j}, \Pi)\exp\Big(-\log \Big(\frac{2j^2}{\delta} \N_H(4^{-j}, \Pi)\Big)\Big) = \frac{\delta}{j^2}.$
	
	Consequently, by a further union bound:
	\begin{equation}
	\begin{split}
	&\P\Bigg[\sup_{\pi \in \Pi}\frac{1}{\sqrt{n}} \Bigg|\sum_{i=1}^n Z_i \Bigg\langle\Gamma_i, \sum_{j=1}^{\jlow} \Big\{A_j(\pi)(X_i) - A_{j-1}(\pi)(X_i)\Big\}\Bigg\rangle\Bigg| \ge \sum_{j=1}^{\jlow} a_j 2^{2 - j} \sqrt{\frac{M(\Pi)}{n}}\Bigg] \\
	 &\le 
	\P\Bigg[\sum_{j=1}^{\jlow}\sup_{\pi \in \Pi}\frac{1}{\sqrt{n}} \Bigg|\sum_{i=1}^n Z_i \Big\langle\Gamma_i,  A_j(\pi)(X_i) - A_{j-1}(\pi)(X_i) \Big\rangle\Bigg| \ge \sum_{j=1}^{\jlow} a_j 2^{2 - j} \sqrt{\frac{ M(\Pi)}{n}}\Bigg]\\
	& \le
	\sum_{j=1}^{\jlow} \P\Bigg[\sup_{\pi \in \Pi}\frac{1}{\sqrt{n}} \Bigg|\sum_{i=1}^n Z_i \Big\langle\Gamma_i,  A_j(\pi)(X_i) - A_{j-1}(\pi)(X_i) \Big\rangle\Bigg| \ge  a_j 2^{2 - j} \sqrt{\frac{ M(\Pi)}{n}}\Bigg] \le \sum_{j=1}^{\jlow}  \frac{\delta}{j^2} <  \sum_{j=1}^{\infty}  \frac{\delta}{j^2} < 1.7 \delta.
	\end{split} 
	\end{equation}
	Take $\delta_k = \frac{1}{2^k}$ and apply the above bound to each $\delta_k$ yields that with probability at least $1 - \frac{1.7}{2^k}$,
	\begin{equation}
	\begin{split}
	&\sup_{\pi \in \Pi}\frac{1}{\sqrt{n}} \Bigg|\sum_{i=1}^n Z_i \Bigg\langle\Gamma_i, \sum_{j=1}^{\jlow} \Big\{A_j(\pi)(X_i) - A_{j-1}(\pi)(X_i)\Big\}\Bigg\rangle\Bigg| \le 4\sqrt{2}\sum_{j=1}^{\jlow} \sqrt{\log \Big(2^{k+1}j^2 \N_H(4^{-j}, \Pi)\Big)} 2^{- j} \sqrt{\frac{ M(\Pi)}{n}}\Bigg] \\
	& \le 4\sqrt{2}\sqrt{\frac{M(\Pi)}{n}}
	\sum_{j=1}^{\jlow}  2^{- j} \sqrt{\log \Big(2^{k+1}j^2 \N_H(4^{-j}, \Pi)\Big) } \\
	& \le 4\sqrt{2}\sqrt{\frac{M(\Pi)}{n}}
	\sum_{j=1}^{\jlow}  2^{- j} \Big(\sqrt{k+1} + \sqrt{2\log j} + \sqrt{\log(\N_H(4^{-j}, \Pi))}\Big)\\
	& \le 4\sqrt{2}\sqrt{\frac {M(\Pi)}{n}}
	\Big\{\sqrt{k+1} \sum_{j=1}^{\infty}  2^{- j} (1 + \sqrt{2\log j}) +  \sum_{j=1}^{\jlow} 2^{- j}\sqrt{\log(\N_H(4^{-j}, \Pi))}\Big\}\\
	& \le 4\sqrt{2}\sqrt{\frac{ M(\Pi)}{n}}
	\Big\{\sqrt{k+1} \sum_{j=1}^{\infty}  \frac{2j}{2^j}  +  \frac{1}{2}\sum_{j=1}^{\jlow} 2^{- j}\Big(\sqrt{\log\N_H(4^{-j}, \Pi)} + \sqrt{\log\N_H(1, \Pi)}\Big) 
	\Big\}\\
	& \le 4\sqrt{2}\sqrt{\frac{ M(\Pi)}{n}}
	\Big\{4\sqrt{k+1}  +  \sum_{j=1}^{\jlow} 2^{- j-1} \sqrt{\log\N_H(4^{-j}, \Pi)} + \frac{1}{2} \sum_{j=1}^{\infty} 2^{- j}\sqrt{\log\N_H(1, \Pi)} \Big\} \\
	& = 4\sqrt{2}\sqrt{\frac{M(\Pi)}{n}}
	\Big\{4\sqrt{k+1}  +  \sum_{j=0}^{\jlow} 2^{- j-1} \sqrt{\log\N_H(4^{-j}, \Pi)} 
	\Big\} < 4\sqrt{2}\sqrt{\frac{M(\Pi)}{n}}
	\Big\{4\sqrt{k+1}  +  \int_{0}^1 \sqrt{\log\N_H(\epsilon^2, \Pi)}d\epsilon\Big\}\\
	& = 4\sqrt{2}\sqrt{\frac{ \sup_{\pi_a, \pi_b}\sum_{i=1}^n |\langle \Gamma_i, \pi_a(X_i) - \pi_b(X_i)\rangle|^2}{n}}
	\Big\{4\sqrt{k+1}  +  \kappa(\Pi)
	\Big\},\\
	\end{split}
	\end{equation}
	where the last inequality follows from setting $\epsilon = 2^{-j}$ and upper bounding the sum using
	the integral. Consequently, for each $k = 0, 1,\dots$, we have:
	\begin{equation}\label{eq:prob_bound}
	\begin{split}
	\P\Bigg[&\sup_{\pi \in \Pi}\frac{1}{\sqrt{n}} \Bigg|\sum_{i=1}^n Z_i \Bigg\langle\Gamma_i, \sum_{j=1}^{\jlow} \Big\{A_j(\pi)(X_i) - A_{j-1}(\pi)(X_i)\Big\}\Bigg\rangle\Bigg|  \\
	&\ge  4\sqrt{2}\sqrt{\frac{ \sup_{\pi_a, \pi_b}\sum_{i=1}^n |\langle \Gamma_i, \pi_a(X_i) - \pi_b(X_i)\rangle|^2}{n}}
	\Big\{4\sqrt{k+1}  +  \kappa(\Pi)
	\Big\}   \Bigg] \le \frac{1.7}{2^k}.\\
	\end{split} 
	\end{equation}
	
	We next turn the probability bound given in Equation~\eqref{eq:prob_bound} into a bound on its (conditional) expectation.
	Specifically, define the (non-negative) random variable $R = \sup_{\pi \in \Pi}\frac{1}{\sqrt{n}} \Bigg|\sum_{i=1}^n Z_i \Bigg\langle\Gamma_i, \sum_{j=1}^{\jlow} \Big\{A_j(\pi)(X_i) - A_{j-1}(\pi)(X_i)\Big\}\Bigg\rangle\Bigg|$
	and let $F_R(\cdot)$ be its cumulative distribution function (conditioned on $\{X_i, \Gamma_i\}_{i=1}^n$).
	Per its definition, we have:
	$$1 - F_R\Bigg(4\sqrt{2}\sqrt{\frac{ \sup_{\pi_a, \pi_b}\sum_{i=1}^n |\langle \Gamma_i, \pi_a(X_i) - \pi_b(X_i)\rangle|^2}{n}}
	\Big\{4\sqrt{k+1}  +  \kappa(\Pi)
	\Big\}  \Bigg) \le \frac{1.7}{2^k}.$$
	Consequently, we have:
	\begin{equation}
	\begin{split}
	&\E\Big[ R \mid \{X_i, \Gamma_i\}_{i=1}^n\Big] = \int_{0}^{\infty} (1 - F_R(r))dr
	\\
	&\le \sum_{k=0}^{\infty} \frac{1.7}{2^k} 4\sqrt{2}\sqrt{\frac{ \sup_{\pi_a, \pi_b}\sum_{i=1}^n |\langle \Gamma_i, \pi_a(X_i) - \pi_b(X_i)\rangle|^2}{n}}
	\Big\{4\sqrt{k+1}  +  \kappa(\Pi)\Big\}  \\
	& = 4\sqrt{2}\sqrt{\frac{ \sup_{\pi_a, \pi_b}\sum_{i=1}^n |\langle \Gamma_i, \pi_a(X_i) - \pi_b(X_i)\rangle|^2}{n}} \Big\{ \sum_{k=0}^{\infty}\frac{1.7}{2^k} 4\sqrt{k+1} + \sum_{k=0}^{\infty}\frac{1.7}{2^k} \kappa(\Pi)\Big\}\\
	& \le 6.8\sqrt{2}\sqrt{\frac{ \sup_{\pi_a, \pi_b}\sum_{i=1}^n |\langle \Gamma_i, \pi_a(X_i) - \pi_b(X_i)\rangle|^2}{n}} \Big\{ \sum_{k=0}^{\infty}\frac{4(k+1)}{2^k}  +  2\kappa(\Pi)\Big\} \\
	& =  6.8\sqrt{2}\sqrt{\frac{ \sup_{\pi_a, \pi_b}\sum_{i=1}^n |\langle \Gamma_i, \pi_a(X_i) - \pi_b(X_i)\rangle|^2}{n}} \Big\{2\kappa(\Pi) + 16\Big\}. \\
	\end{split}
	\end{equation}
	
	Taking expectation with respect to $\{X_i, \Gamma_i\}_{i=1}^n$, we obtain:
	\begin{equation}\label{eq:lower1}
	\begin{split}
	&\E\Big[ \sup_{\pi \in \Pi}\frac{1}{\sqrt{n}} \Bigg|\sum_{i=1}^n Z_i \Bigg\langle\Gamma_i, \sum_{j=1}^{\jlow} \Big\{A_j(\pi)(X_i) - A_{j-1}(\pi)(X_i)\Big\}\Bigg\rangle\Bigg| \Big] \\
	&\le 6.8\sqrt{2}\Big\{2\kappa(\Pi) + 16\Big\} \E\Big[\sqrt{\frac{ \sup_{\pi_a, \pi_b}\sum_{i=1}^n |\langle \Gamma_i, \pi_a(X_i) - \pi_b(X_i)\rangle|^2}{n}}\Big] \\
	& \le 13.6\sqrt{2}\Big\{\kappa(\Pi) + 8\Big\} \sqrt{\E\Big[\frac{ \sup_{\pi_a, \pi_b}\sum_{i=1}^n |\langle \Gamma_i, \pi_a(X_i) - \pi_b(X_i)\rangle|^2}{n}\Big] }.
	\end{split}
	\end{equation}
\end{proofstep}

\begin{proofstep}{Refining the lower range bound using Talagrand's inequality}\label{step:5}
	
To prove  Equation~\eqref{eq:lower3}, we apply Lemma~\ref{lem:tal2} to the current context: we identify
	$X_i$ in Lemma~\ref{lem:tal2} with $(X_i, \Gamma_i)$ here and $f(X_i, \Gamma_i) = \langle \Gamma_i, \pi_a(X_i) - \pi_b(X_i)\rangle$. Since $\Gamma_i$ is bounded, $f(X_i, \Gamma_i) \le \|\Gamma_i\|_2 \|\pi_a(X_i) - \pi_b(X_i)\|_2 \le \sqrt{2}  \|\Gamma_i\|_2 \le U, \forall \pi_a, \pi_b \in \Pi$ for some constant $U$.
	Consequently, we have:
	\begin{equation}
	\begin{split}
	&\E\Big[\sup_{\pi_a, \pi_b \in \Pi}\sum_{i=1}^n \Big(\langle \Gamma_i, \pi_a(X_i) - \pi_b(X_i)\rangle\Big)^2\Big] \\
	&\le n \sup_{\pi_a, \pi_b \in \Pi} \E \Big[\Big(\langle \Gamma_i, \pi_a(X_i) - \pi_b(X_i)\rangle\Big)^2 \Big]+ 
	8U\E\Big[\sup_{\pi_a, \pi_b \in \Pi} \sum_{i=1}^n \Big|Z_i  \langle \Gamma_i, \pi_a(X_i) - \pi_b(X_i)\rangle \Big|\Big]. \\
	\end{split}
	\end{equation}
	
	Dividing both sides by $n$ then yields:
	\begin{equation}\label{eq:lower2}
	\begin{split}
	&\E\Big[\frac{ \sup_{\pi_a, \pi_b}\sum_{i=1}^n |\langle \Gamma_i, \pi_a(X_i) - \pi_b(X_i)\rangle|^2}{n}\Big] \\
	& \le  \sup_{\pi_a, \pi_b \in \Pi} \E \Big[\Big(\langle \Gamma_i, \pi_a(X_i) - \pi_b(X_i)\rangle\Big)^2 \Big]+ 
	8U\E\Big[\sup_{\pi_a, \pi_b\in \Pi} \frac{1}{n} \sum_{i=1}^n Z_i \Big| \langle \Gamma_i, \pi_a(X_i) -\pi_b(X_i)\rangle\Big|\Big] \\
	& =  \sup_{\pi_a, \pi_b \in \Pi} \E \Big[\langle \Gamma_i, \pi_a(X_i) - \pi_b(X_i)\rangle^2 \Big] + 8U \mathcal{R}_n(\PiD).
	\end{split}
	\end{equation}
	
	Therefore, by combining Equation~\eqref{eq:lower1} with Equation~\eqref{eq:lower2}, we have:
	\begin{equation}\label{eq:lower}
	\begin{split}
	&\E\Big[ \sup_{\pi \in \Pi}\frac{1}{\sqrt{n}} \Bigg|\sum_{i=1}^n Z_i \Bigg\langle\Gamma_i, \sum_{j=1}^{\jlow} \Big\{A_j(\pi)(X_i) - A_{j-1}(\pi)(X_i)\Big\}\Bigg\rangle\Bigg| \Big] \\
	& \le 13.6\sqrt{2}\Big\{\kappa(\Pi) + 8\Big\} \sqrt{\sup_{\pi_a, \pi_b \in \Pi} \E \Big[\langle \Gamma_i, \pi_a(X_i) - \pi_b(X_i)\rangle^2 \Big] + 8U\mathcal{R}_n(\PiD)}.
	\end{split}
	\end{equation}
	
	Finally, combining Equation~\eqref{eq:rad_decompose}, we have:
	\begin{equation}
	\begin{split}
	& \sqrt{n}\mathcal{R}_n(\PiD) = \E\Big[\sup_{\pi_a,\pi_b \in \Pi}\frac{1}{\sqrt{n}} \Big|\sum_{i=1}^n Z_i \langle\Gamma_i, \pi_a(X_i) - \pi_b(X_i)\rangle \Big|\Big]  \\
	& \le 2\E\Big[\sup_{\pi_a,\pi_b \in \Pi}\frac{1}{\sqrt{n}} \Big|\sum_{i=1}^n Z_i \langle\Gamma_i, \sum_{j=1}^{\jlow} \Big\{A_j(\pi)(X_i) - A_{j-1}(\pi)(X_i)\Big\}\rangle\Big|\Big] + o(1) \\
	&\le 27.2\sqrt{2}\Big\{\kappa(\Pi) + 8\Big\} \sqrt{\sup_{\pi_a, \pi_b \in \Pi} \E \Big[\langle \Gamma_i, \pi_a(X_i) - \pi_b(X_i)\rangle^2 \Big] + 8U\mathcal{R}_n(\PiD)} +  o(1).
	\end{split}
	\end{equation}
	
	Dividing both sides of the above inequality by $\sqrt{n}$ yields:
	\begin{equation}
	\begin{split}\label{eq:final}
	\mathcal{R}_n(\Pi) &\le 27.2\sqrt{2}(\kappa(\Pi) + 8) \sqrt{\frac{\sup_{\pi_a, \pi_b \in \Pi} \E \Big[|\langle \Gamma_i, \pi_a(X_i) - \pi_b(X_i)\rangle|^2 \Big] + 8U\mathcal{R}_n(\PiD)}{n}} + o(\frac{1}{\sqrt{n}})\\
	& \le 27.2\sqrt{2}(\kappa(\Pi) + 8) \Big\{\sqrt{\frac{\sup_{\pi_a, \pi_b \in \Pi} \E \Big[|\langle \Gamma_i, \pi_a(X_i) - \pi_b(X_i)\rangle|^2 \Big]}{n}} + \sqrt{\frac{8U\mathcal{R}_n(\PiD)}{n}} \Big\}+ o(\frac{1}{\sqrt{n}}).
	\end{split}
	\end{equation}
	The above equation immediately implies
	$\mathcal{R}_n(\Pi)  = O(\sqrt{\frac{1}{n}}) + O(\sqrt{\frac{\mathcal{R}_n}{n}})$, which one can solve to obtain
	$\mathcal{R}_n(\Pi) = O(\sqrt{\frac{1}{n}}).$
	Plugging it into Equation~\eqref{eq:final} then results:
	\begin{equation}
	\begin{split}
	\mathcal{R}_n(\Pi) & \le 27.2\sqrt{2}(\kappa(\Pi) + 8) \Big\{\sqrt{\frac{\sup_{\pi_a, \pi_b \in \Pi} \E \Big[|\langle \Gamma_i, \pi_a(X_i) - \pi_b(X_i)\rangle|^2 \Big]}{n}} + \sqrt{\frac{ O(\sqrt{\frac{1}{n}})  }{n}} \Big\}+ o(\frac{1}{\sqrt{n}}) \\
	& = 27.2\sqrt{2}(\kappa(\Pi) + 8) \Big\{\sqrt{\frac{\sup_{\pi_a, \pi_b \in \Pi} \E \Big[|\langle \Gamma_i, \pi_a(X_i) - \pi_b(X_i)\rangle|^2 \Big]}{n}} \Big\} + o(\frac{1}{\sqrt{n}}).
	\end{split}
	\end{equation}
\end{proofstep}

\section{Proof of Theorem~\ref{thm:uniform_counter}}
\setcounter{proofstep}{0}
\begin{proofstep}{Expected uniform bound on maximum deviation}\label{step:1}
	
First, denote 
$\mu_a(X_i) =  
\begin{bmatrix}
\mu_{a^1}(X_i) \\
\mu_{a^2}(X_i)\\
...\\
\mu_{a^d}(X_i) 
\end{bmatrix}$,
we can then compute the expectation of the influence function:
\begin{equation}
\begin{split}
&\E[\QI (\pi)] = \E[\frac{1}{n} \sum_{i=1}^n \langle \pi(X_i), \Gamma_i \rangle] =
\E[\langle \pi(X_i), \Gamma_i \rangle] = \E\Big[ \E[\langle \pi(X_i), \Gamma_i \rangle]\Big| X_i]\Big] \\
& =\E\Big[ \Big\langle \pi(X_i), \E[ \frac{Y_i(A_i) - \mu_{\W_i}(X_i)}{e_{\W_i}(X_i)} \cdot  \W_i \Big| X_i] + \mu_a(X_i)\Big  \rangle\Big] \\
&= \E\Big[ \Big\langle \pi(X_i), \E[ \sum_{j=1}^d\frac{Y_i(a^j) - \mu_{a^j}(X_i)}{e_{a^j}(X_i)} \cdot  \mathbf{1}_{\{\W_i = a^j\}} a^j| X_i] + \mu_a(X_i)\Big  \rangle\Big] \\
&= \E\Big[ \Big\langle \pi(X_i), \sum_{j=1}^d\frac{Y_i(a^j) - \mu_{a^j}(X_i)}{e_{a^j}(X_i)} \cdot  \P(\W_i = a^j | X_i) \cdot a^j] + \mu_a(X_i)\Big  \rangle\Big] \\
&= \E\Big[ \Big\langle \pi(X_i), \sum_{j=1}^d\frac{Y_i(a^j) - \mu_{a^j}(X_i)}{e_{a^j}(X_i)} \cdot  e_{a^j}(X_i) \cdot a^j] + \mu_a(X_i)\Big  \rangle\Big]\\
&= \E\Big[ \Big\langle \pi(X_i), \sum_{j=1}^d \Big(Y_i(a^j) - \mu_{a^j}(X_i)\Big) \cdot a^j + \mu_a(X_i)\Big  \rangle\Big] \\
&= \E\Big[ \Big\langle \pi(X_i), \sum_{j=1}^d Y_i(a^j) \cdot a^j \Big  \rangle\Big] 
+ \E\Big[ \Big\langle \pi(X_i), \mu_a(X_i) - \sum_{j=1}^d\mu_{a^j}(X_i) \cdot a^j  \Big  \rangle\Big] \\
& =  \E\Big[ \Big\langle \pi(X_i), \sum_{j=1}^d Y_i(a^j) \cdot a^j \Big  \rangle\Big]
= \sum_{j=1}^d \E\Big[ \langle \pi(X_i), Y_i(a^j) \cdot a^j \rangle\Big]
= \sum_{j=1}^d \E\Big[ Y_i(a^j) \mathbf{1}_{\{\pi(X_i) = a^j\}} \Big]\\
&= \sum_{j=1}^d \E\Big[ Y_i(\pi(X_i)) \mathbf{1}_{\{\pi(X_i) = a^j\}} \Big]
= \E\Big[ Y_i(\pi(X_i)) \sum_{j=1}^d \mathbf{1}_{\{\pi(X_i) = a^j\}} \Big]
= \E\Big[ Y_i(\pi(X_i)) \Big] = Q(\pi).
\end{split}
\end{equation}
Consequently, $\E[\tilde{\Delta}(\pi_1, \pi_2)] = \E[\QI (\pi_1)] - \E[\QI (\pi_2)]= Q(\pi_1) - Q(\pi_2) =\Delta(\pi_1, \pi_2)$.

Finally, classical results on Rademacher complexity~\cite{bartlett2002rademacher} then give:
\begin{equation}
\begin{split}
&\E\Big[\sup_{\pi_1, \pi_2 \in \Pi} \Big|\tilde{\Delta}(\pi_1, \pi_2) - \Delta(\pi_1, \pi_2)\Big| \Big] \le 2\E\Big[\mathcal{R}_n(\Pi \times \Pi)\Big] \le 4\Big[\mathcal{R}_n(\Pi)\Big] \\
& \le 54.4\sqrt{2}(\kappa(\Pi) + 8) \sqrt{\frac{\sup_{\pi_a, \pi_b \in \Pi}\E[\langle\Gamma_i, \pi_a(X_i) - \pi_b(X_i)\rangle^2]}{n}} + o(\frac{1}{\sqrt{n}}),
\end{split}
\end{equation}
where the last inequality follows from Theorem~\ref{thm:rad}.
\end{proofstep}

\begin{proofstep}{High probability bound on maximum deviation via Talagrand inequality}\label{step:2}
Since $\langle \Gamma_i , \pi_1(X_i) - \pi_2(X_i) \rangle \le U$ (surely),
it follows from Lemma~\ref{lem:tal2} that: 
\begin{equation}
\begin{split}
& \E\Big[\sup_{\pi_1, \pi_2 \in \Pi} \sum_{i=1}^n \Big\{\langle \Gamma_i , \pi_1(X_i) - \pi_2(X_i) \rangle - \E\Big\{\langle \Gamma_i , \pi_1(X_i) - \pi_2(X_i)\rangle\Big\}  \Big\}^2\Big]  \\
&  \le n \sup_{\pi_1, \pi_2 \in \Pi} \V(\langle \Gamma_i , \pi_1(X_i) - \pi_2(X_i) \rangle)
\\
& + 8 U\E\Big[\sup_{\pi_1, \pi_2 \in \Pi}\Big|\sum_{i=1}^n Z_i \Big(\langle \Gamma_i , \pi_1(X_i) - \pi_2(X_i) \rangle - \E[\langle \Gamma_i , \pi_1(X_i) - \pi_2(X_i)\rangle] \Big)  \Big|\Big]\\
& \le 
n\sup_{\pi_1, \pi_2 \in \Pi} \E \Big\{\Big|\langle \Gamma_i , \pi_1(X_i) - \pi_2(X_i) \rangle\Big|^2\Big\} + 
8U \E\Big[\sup_{\pi_1, \pi_2 \in \Pi}\Big|\sum_{i=1}^n Z_i \langle \Gamma_i , \pi_1(X_i) - \pi_2(X_i) \rangle   \Big|\Big] \\
&+
8U \E\Big[\sup_{\pi_1, \pi_2 \in \Pi}\Big|\sum_{i=1}^n  Z_i  \E[\langle \Gamma_i , \pi_1(X_i) - \pi_2(X_i)\rangle]   \Big|\Big] \\
&\le n\sup_{\pi_1, \pi_2 \in \Pi} \E \Big\{\Big|\langle \Gamma_i , \pi_1(X_i) - \pi_2(X_i) \rangle\Big|^2\Big\} + 
8U \E\Big[\sup_{\pi_1, \pi_2 \in \Pi}\Big|\sum_{i=1}^n Z_i \langle \Gamma_i , \pi_1(X_i) - \pi_2(X_i) \rangle   \Big|\Big] \\
& + 8U \E\Big[\sup_{\pi_1, \pi_2 \in \Pi}\Big|\sum_{i=1}^n \frac{1}{n} Z_i \langle \Gamma_i , \pi_1(X_i) - \pi_2(X_i)\rangle   \Big|\Big]\\
& = n\sup_{\pi_1, \pi_2 \in \Pi} \E \Big\{\Big|\langle \Gamma_i , \pi_1(X_i) - \pi_2(X_i) \rangle\Big|^2\Big\} + 
16U \E\Big[\sup_{\pi_1, \pi_2 \in \Pi}\Big|\sum_{i=1}^n Z_i \langle \Gamma_i , \pi_1(X_i) - \pi_2(X_i) \rangle   \Big|\Big],
\end{split}
\end{equation}
where the last inequality follows from Jensen by noting that: 
\begin{equation}
\begin{split}
&\E\Big[\sup_{\pi_1, \pi_2 \in \Pi}\Big|\sum_{i=1}^n  Z_i  \E[\langle \Gamma_i , \pi_1(X_i) - \pi_2(X_i)\rangle] \Big|\Big]\le 	\E\Big[\sup_{\pi_1, \pi_2 \in \Pi}\E\Big|\sum_{i=1}^n \frac{1}{n} Z_i \langle \Gamma_i , \pi_1(X_i) - \pi_2(X_i)\rangle   \Big|\Big] \\
& \le \E\Big[\sup_{\pi_1, \pi_2 \in \Pi}\Big|\sum_{i=1}^n \frac{1}{n} Z_i \langle \Gamma_i , \pi_1(X_i) - \pi_2(X_i)\rangle   \Big|\Big].
\end{split}
\end{equation}

Consequently, with the identification that $f(X_i) = \frac{\langle \Gamma_i , \pi_1(X_i) - \pi_2(X_i) \rangle - \E[\langle \Gamma_i , \pi_1(X_i) - \pi_2(X_i)]}{2U}$, we have:

\begin{equation}\label{eq:V_bound0}
\begin{split}
&\E\Big[\sup_{f \in \mathcal{F}} \sum_{i=1}^n f^2(X_i) \Big] = \E \Big[\sum_{i=1}^n\Big(\frac{\langle \Gamma_i , \pi_1(X_i) - \pi_2(X_i) \rangle - \E[\langle \Gamma_i , \pi_1(X_i) - \pi_2(X_i)]}{2U} \Big)^2 \Big] \\
&\le \frac{n}{4U^2}\sup_{\pi_1, \pi_2 \in \Pi} \E \Big\{\Big|\langle \Gamma_i , \pi_1(X_i) - \pi_2(X_i) \rangle\Big|^2\Big\} + 
\frac{16}{2U} \E\Big[\sup_{\pi_1, \pi_2 \in \Pi}\Big|\sum_{i=1}^n Z_i \langle \Gamma_i , \pi_1(X_i) - \pi_2(X_i) \rangle   \Big|\Big].
\end{split}
\end{equation}

Next, setting $t = 2\sqrt{2(\log \frac{1}{\delta}) V} + 2\log \frac{1}{\delta}$, we then have:
\begin{equation}\label{eq:V_bound}
\begin{split}
&\exp(-\frac{t}{2}\log(1+ \frac{t}{V})) =\exp\Big(-\frac{2\sqrt{2\log \frac{1}{\delta} V} + 2 \log \frac{1}{\delta}}{2} \log(1 + \frac{2\sqrt{2\log \frac{1}{\delta} V} + 2 \log \frac{1}{\delta}}{V})\Big) \\
&\le \exp\Big(-\frac{2\sqrt{2(\log \frac{1}{\delta}) V} + 2 \log \frac{1}{\delta}}{2}  \frac{\frac{2\sqrt{2(\log \frac{1}{\delta}) V} + K \log \frac{1}{\delta}}{V} }{1 + \frac{2\sqrt{2(\log \frac{1}{\delta}) V} + 2 \log \frac{1}{\delta}}{V} } \Big)= \exp\Big(-\frac{1}{2} \frac{(2\sqrt{2(\log \frac{1}{\delta}) V} + 2 \log \frac{1}{\delta})^2}{V + 2\sqrt{2(\log \frac{1}{\delta}) V} + 2 \log \frac{1}{\delta}}\Big) \\
&= \exp\Big(\frac{1}{2} (\frac{2\sqrt{2(\log \frac{1}{\delta}) V} + 2 \log \frac{1}{\delta}}{\sqrt{V} + \sqrt{2\log \frac{1}{\delta}}})^2\Big) \le \exp\Big(-\frac{1}{2}  (\sqrt{2\log \frac{1}{\delta}})^2\Big) = \exp(-\log \frac{1}{\delta}) = \delta.
\end{split}
\end{equation}

Consequently, applying Lemma~\ref{lem:tal} with $t = 2\sqrt{2(\log \frac{1}{\delta}) V} + 2\log \frac{1}{\delta}$ and noting that 
$\Big|\langle \Gamma_i , \pi_1(X_i) - \pi_2(X_i) \rangle - \E[\langle \Gamma_i , \pi_1(X_i) - \pi_2(X_i)] \Big|\le 2U$ surely yields:
\begin{equation}
\begin{split}
&\P\Bigg\{\Bigg|\sup_{\pi_1, \pi_2 \in \Pi} \frac{n}{2U}\Big|\tilde{\Delta}(\pi_1, \pi_2) - \Delta(\pi_1, \pi_2)\Big|
- \E\Big\{\sup_{\pi_1, \pi_2 \in \Pi} \frac{n}{2U}\Big|\tilde{\Delta}(\pi_1, \pi_2) - \Delta(\pi_1, \pi_2)\Big| \Big\} \Bigg| \ge t\Bigg\} \\
& = \P\Bigg\{\Bigg|\sup_{\pi_1, \pi_2 \in \Pi} \Big|\sum_{i=1}^n \frac{\langle \Gamma_i , \pi_1(X_i) - \pi_2(X_i) \rangle - \E[\langle \Gamma_i , \pi_1(X_i) - \pi_2(X_i)]}{2U}\Big|
- \\
&\E\Big\{\sup_{\pi_1, \pi_2 \in \Pi} \Big|\sum_{i=1}^n \frac{\langle \Gamma_i , \pi_1(X_i) - \pi_2(X_i) \rangle - \E[\langle \Gamma_i , \pi_1(X_i) - \pi_2(X_i)]}{2U} \Big| \Big\} \Bigg| \ge 2\sqrt{2(\log \frac{1}{\delta}) V} + 2\log \frac{1}{\delta}\Bigg\} \\
& \le 2\exp(-\frac{t}{2}\log(1+ \frac{t}{V}))\le 2\delta.
\end{split}
\end{equation}

This means that 
with probability at least $1 - 2\delta$:
\begin{equation}
\begin{split}
\sup_{\pi_1, \pi_2 \in \Pi} \frac{n}{2U}\Big|\tilde{\Delta}(\pi_1, \pi_2) - \Delta(\pi_1, \pi_2)\Big| \le
\E\Big\{\sup_{\pi_1, \pi_2 \in \Pi} \frac{n}{2U}\Big|\tilde{\Delta}(\pi_1, \pi_2) - \Delta(\pi_1, \pi_2)\Big| \Big\} + 2\sqrt{2(\log \frac{1}{\delta}) V} + 2\log \frac{1}{\delta}.
\end{split}
\end{equation}

Now multiplying both sides by $2U$, dividing both sides by $n$ and plugging the following $V$ value (which by Equation~\eqref{eq:V_bound0} satisfies the requirement in Lemma~\ref{lem:tal}):
$$V = \frac{n}{4U^2}\sup_{\pi_1, \pi_2 \in \Pi} \E \Big\{\Big|\langle \Gamma_i , \pi_1(X_i) - \pi_2(X_i) \rangle\Big|^2\Big\} + 
\frac{16}{2U} \E\Big[\sup_{\pi_1, \pi_2 \in \Pi}\Big|\sum_{i=1}^n Z_i \langle \Gamma_i , \pi_1(X_i) - \pi_2(X_i) \rangle   \Big|\Big],$$
it follows that with probability at least $1 -2\delta$:
\begin{equation}
\begin{split}
&\sup_{\pi_1, \pi_2 \in \Pi} \Big|\tilde{\Delta}(\pi_1, \pi_2) - \Delta(\pi_1, \pi_2)\Big| \le
\E\Big\{\sup_{\pi_1, \pi_2 \in \Pi} \Big|\tilde{\Delta}(\pi_1, \pi_2) - \Delta(\pi_1, \pi_2)\Big| \Big\} + \frac{2}{n}\sqrt{2U^2(\log \frac{1}{\delta}) V} + \frac{2}{n}\log \frac{1}{\delta}\\
&\le \E\Big\{\sup_{\pi_1, \pi_2 \in \Pi} \Big|\tilde{\Delta}(\pi_1, \pi_2) - \Delta(\pi_1, \pi_2)\Big| \Big\} + \frac{2}{n}\sqrt{\frac{n}{2}\sup_{\pi_1, \pi_2 \in \Pi} \E \Big\{\Big|\langle \Gamma_i , \pi_1(X_i) - \pi_2(X_i) \rangle\Big|^2\Big\}\log\frac{1}{\delta}} \\
& + \frac{2}{n}\sqrt{16 U\E\Big[\sup_{\pi_1, \pi_2 \in \Pi}\Big|\sum_{i=1}^n Z_i \langle \Gamma_i , \pi_1(X_i) - \pi_2(X_i) \rangle   \Big|\Big]\log\frac{1}{\delta}}+ \frac{2}{n}\log \frac{1}{\delta} \\
& = \E\Big\{\sup_{\pi_1, \pi_2 \in \Pi} \Big|\tilde{\Delta}(\pi_1, \pi_2) - \Delta(\pi_1, \pi_2)\Big| \Big\} + \sqrt{2\log\frac{1}{\delta}}\sqrt{\frac{\sup_{\pi_1, \pi_2 \in \Pi} \E \Big\{\Big|\langle \Gamma_i , \pi_1(X_i) - \pi_2(X_i) \rangle\Big|^2\Big\}}{n}} \\
& + 8\sqrt{ \frac{U}{n}\E\Big[\sup_{\pi_1, \pi_2 \in \Pi}\frac{1}{n}\Big|\sum_{i=1}^n Z_i \langle \Gamma_i , \pi_1(X_i) - \pi_2(X_i) \rangle   \Big|\Big]\log\frac{1}{\delta}}+ \frac{2}{n}\log \frac{1}{\delta} \\
& =  \E\Big\{\sup_{\pi_1, \pi_2 \in \Pi} \Big|\tilde{\Delta}(\pi_1, \pi_2) - \Delta(\pi_1, \pi_2)\Big| \Big\} + \sqrt{2\log\frac{1}{\delta}}\sqrt{\frac{\sup_{\pi_1, \pi_2 \in \Pi} \E \Big\{\Big|\langle \Gamma_i , \pi_1(X_i) - \pi_2(X_i) \rangle\Big|^2\Big\}}{n}} \\
& + \sqrt{\frac{O(\frac{1}{\sqrt{n}})}{n})} +  O(\frac{1}{n}) \\
& = \E\Big\{\sup_{\pi_1, \pi_2 \in \Pi} \Big|\tilde{\Delta}(\pi_1, \pi_2) - \Delta(\pi_1, \pi_2)\Big| \Big\} + \sqrt{2\log\frac{1}{\delta}}\sqrt{\frac{\sup_{\pi_1, \pi_2 \in \Pi} \E \Big\{\Big|\langle \Gamma_i , \pi_1(X_i) - \pi_2(X_i) \rangle\Big|^2\Big\}}{n}} + O(\frac{1}{n^{0.75}}).
\end{split}
\end{equation}

\end{proofstep}
	
\section{Proof of Lemma~\ref{lem:uniform_counter}}

{\em Proof:}
Take any two policies $\pi^a, \pi^b \in \Pi$ (here we use superscripts $a,b$ because we will also use subscripts $\pi_j$ to access the $j$-th component of a policy $\pi$).
We start by rewriting the $K$-fold doubly robust estimator difference function as follows: 
\begin{equation}
\begin{split}
&\diffd(\pi^a, \pi^b) =\QE_{DR} (\pi^a) - \QE_{DR} (\pi^b)= \frac{1}{n} \sum_{i=1}^n \langle \pi^a(X_i), \factorf_i \rangle
- \frac{1}{n} \sum_{i=1}^n \langle \pi^b(X_i), \factorf_i \rangle =  \frac{1}{n} \sum_{i=1}^n \langle \pi^a(X_i) - \pi^b(X_i), \factorf_i \rangle\\
&= \frac{1}{n} \sum_{i=1}^n \langle \pi^a(X_i) - \pi^b(X_i),  \frac{Y_i - \hat{\mu}_{\W_i}^{-k(i)}(X_i)}{\hat{e}_{\W_i}^{-k(i)}(X_i)} \cdot  \W_i
+
\begin{bmatrix}
\hat{\mu}_{a^1}^{-k(i)}(X_i) \\
\hat{\mu}_{a^2}^{-k(i)}(X_i)\\
...\\
\hat{\mu}_{a^d}^{-k(i)}(X_i) 
\end{bmatrix} \rangle \\
& =\frac{1}{n} \sum_{i=1}^n  \sum_{j=1}^d \Big(\pi_j^a(X_i) - \pi_j^b(X_i)\Big) \Big(\frac{Y_i - \hat{\mu}_{a^j}^{-k(i)}(X_i)}{\hat{e}_{a^j}^{-k(i)}(X_i)} \cdot  \mathbf{1}_{\{\W_i = a^j\}}
+
\hat{\mu}_{a^j}^{-k(i)}(X_i) \Big) \\
&=
\sum_{j=1}^d  \Big\{\frac{1}{n} \sum_{i=1}^n \Big(\pi_j^a(X_i) - \pi_j^b(X_i)\Big) \Big( \frac{Y_i - \hat{\mu}_{a^j}^{-k(i)}(X_i)}{\hat{e}_{a^j}^{-k(i)}(X_i)} \cdot  \mathbf{1}_{\{\W_i = a^j\}}
+
\hat{\mu}_{a^j}^{-k(i)}(X_i) \Big)\Big\} = \sum_{j=1}^d \diffd_{DR}^{j} (\pi^a, \pi^b),
\end{split}
\end{equation}
where $ \diffd_{DR}^{j} (\pi) \triangleq \frac{1}{n} \sum_{i=1}^n \Big(\pi_j^a(X_i) -\pi_j^b(X_i)\Big) \Big( \frac{Y_i - \hat{\mu}_{a^j}^{-k(i)}(X_i)}{\hat{e}_{a^j}^{-k(i)}(X_i)} \cdot  \mathbf{1}_{\{\W_i = a^j\}}
+
\hat{\mu}_{a^j}^{-k(i)}(X_i) \Big)$ and $\pi_j^a(X_i), \pi_j^b(X_i)$ are the $j$-th coordinates of $\pi_j^a(X_i), \pi_j^b(X_i)$ respectively, which are either $1$ or 0 (recall that $\pi_j(X_i) = 1$ for a policy $\pi$ if and only if the $j$-th action is selected).

We can similarly decompose the influence difference function as follows:
\begin{equation}
\begin{split}
&\diffi (\pi^a, \pi^b) = \frac{1}{n} \sum_{i=1}^n \langle \pi^a(X_i) - \pi^b(X_i), \factor_i \rangle= \frac{1}{n} \sum_{i=1}^n \langle \pi^a(X_i) - \pi^b(X_i),  \frac{Y_i - \mu_{\W_i}(X_i)}{e_{\W_i}(X_i)} \cdot  \W_i
+
\begin{bmatrix}
\mu_{a^1}(X_i) \\
\mu_{a^2}(X_i)\\
...\\
\mu_{a^d}(X_i) 
\end{bmatrix} \rangle \\
&=
\sum_{j=1}^d  \Big\{\frac{1}{n} \sum_{i=1}^n \Big(\pi_j^a(X_i) - \pi_j^b(X_i)\Big) \Big( \frac{Y_i - \mu_{a^j}(X_i)}{e_{a^j}(X_i)} \cdot  \mathbf{1}_{\{\W_i = a^j\}}
+
\mu_{a^j}(X_i) \Big)\Big\} = \sum_{j=1}^d \diffi^{j} (\pi^a, \pi^b),
\end{split}
\end{equation}
where $ \diffi^{j} (\pi) \triangleq \frac{1}{n} \sum_{i=1}^n \Big(\pi_j^a(X_i) -\pi_j^b(X_i) \Big)\Big( \frac{Y_i - \mu_{a^j}(X_i)}{e_{a^j}(X_i)} \cdot  \mathbf{1}_{\{\W_i = a^j\}}
+
\mu_{a^j}(X_i) \Big)$.

Since $\diffd_{DR} (\pi^a,\pi^b) - \diffi (\pi^a, \pi^b) 
= \sum_{i=1}^d \Big(\diffd_{DR}^{j} (\pi^a, \pi^b) - \diffi^{j} (\pi^a, \pi^b)\Big)$, we provide an upper bound  for each generic term $\diffd_{DR}^{j} (\pi^a, \pi^b) - \diffi^{j} (\pi^a, \pi^b).$
To do so, we construct the following decomposition:

\begin{equation}
\begin{split}
\diffd_{DR}^{j} (\pi^a, \pi^b) - \diffi^{j} (\pi^a, \pi^b) &=  \frac{1}{n}\sum_{i=1}^n \Big(\pi_j^a(X_i) -\pi_j^b(X_i)\Big) \Big( 
\hat{\mu}_{a^j}^{-k(i)}(X_i) - \mu_{a^j}(X_i) \Big)
\\
&+ \frac{1}{n}\sum_{i=1}^n \Big(\pi_j^a(X_i) - \pi_j^b(X_i)\Big) \mathbf{1}_{\{\W_i = a^j\}} \Big( \frac{Y_i - \hat{\mu}_{a^j}^{-k(i)}(X_i)}{\hat{e}_{a^j}^{-k(i)}(X_i)} 
-  \frac{Y_i - \mu_{a^j}(X_i)}{e_{a^j}(X_i)}  \Big) \\
& =  \frac{1}{n}\sum_{i=1}^n \Big(\pi_j^a(X_i) - \pi_j^b(X_i)\Big) \Big( 
\hat{\mu}_{a^j}^{-k(i)}(X_i) - \mu_{a^j}(X_i) \Big)\Big(1 - \frac{\mathbf{1}_{\{\W_i = a^j\}}}{e_{a^j}(X_i)} \Big)\\
& +
\frac{1}{n}\sum_{\{i \mid \W_i =a^j\}} \Big(\pi_j^a(X_i) - \pi_j^b(X_i)\Big) \Big( 
Y_i - \mu_{a^j}(X_i) \Big)\Big(\frac{1}{\hat{e}_{a^j}^{-k(i)}(X_i)} - \frac{1}{e_{a^j}(X_i)} \Big)\\
&+ 
\frac{1}{n}\sum_{\{i \mid \W_i =a^j\}} \Big(\pi_j^a(X_i) - \pi_j^b(X_i)\Big) \Big( 
\mu_{a^j}(X_i)  - \hat{\mu}_{a^j}^{-k(i)}(X_i)\Big)\Big(\frac{1}{\hat{e}_{a^j}^{-k(i)}(X_i)} - \frac{1}{e_{a^j}(X_i)} \Big).
\end{split}
\end{equation} 

We bound each of the three terms in turn. For ease of reference, denote
\begin{enumerate}
	\item
	$S_1^j(\pi^a, \pi^b) \triangleq \frac{1}{n}\sum_{i=1}^n \Big(\pi_j^a(X_i) -\pi_j^b(X_i)\Big) \Big( 
	\hat{\mu}_{a^j}^{-k(i)}(X_i) - \mu_{a^j}(X_i) \Big)\Big(1 - \frac{\mathbf{1}_{\{\W_i = a^j\}}}{e_{a^j}(X_i)} \Big)$.
	\item  
	$S_2^j(\pi^a, \pi^b) \triangleq \frac{1}{n}\sum_{\{i \mid \W_i =a^j\}} \Big(\pi_j^a(X_i) - \pi_j^b(X_i)\Big) \Big( 
	Y_i(a^j) - \mu_{a^j}(X_i) \Big)\Big(\frac{1}{\hat{e}_{a^j}^{-k(i)}(X_i)} - \frac{1}{e_{a^j}(X_i)} \Big)$.
	\item 
	$S_3^j(\pi^a, \pi^b) \triangleq \frac{1}{n}\sum_{\{i \mid \W_i =a^j\}} \Big(\pi_j^a(X_i) - \pi_j^b(X_i)\Big) \Big( 
	\mu_{a^j}(X_i)  - \hat{\mu}_{a^j}^{-k(i)}(X_i)\Big)\Big(\frac{1}{\hat{e}_{a^j}^{-k(i)}(X_i)} - \frac{1}{e_{a^j}(X_i)} \Big).$
\end{enumerate}

By definition, we have $\diffd_{DR}^{j} (\pi^a, \pi^b) - \diffi^{j} (\pi^a, \pi^b) = S_1^j(\pi^a, \pi^b) + S_2^j(\pi^a, \pi^b) +S_3^j(\pi^a, \pi^b)$.
Define further:
\begin{enumerate}
	\item
	$S_1^{j,k}(\pi^a, \pi^b) \triangleq \frac{1}{n}\sum_{\{i \mid k(i) =k\}}  \Big(\pi_j^a(X_i) - \pi_j^b(X_i)\Big) \Big( 
	\hat{\mu}_{a^j}^{-k}(X_i) - \mu_{a^j}(X_i) \Big)\Big(1 - \frac{\mathbf{1}_{\{\W_i = a^j\}}}{e_{a^j}(X_i)} \Big)$.
	\item
	$S_2^{j,k}(\pi^a, \pi^b) \triangleq \frac{1}{n}\sum_{\{i \mid k(i) =k,\W_i = a^j\}}  \Big(\pi_j^a(X_i) - \pi_j^b(X_i)\Big) \Big( 
	Y_i(a^j) - \mu_{a^j}(X_i) \Big)\Big(\frac{1}{\hat{e}_{a^j}^{-k(i)}(X_i)} - \frac{1}{e_{a^j}(X_i)} \Big)$.
\end{enumerate}
Clearly $S_1^j = \sum_{k=1}^K S_1^{j,k}(\pi^a, \pi^b), S_2^j(\pi^a, \pi^b) = \sum_{k=1}^K S_2^{j,k}(\pi^a, \pi^b)$.	

Now since $\hat{\mu}_{a^j}^{-k}(\cdot)$ is computed using the rest $K-1$ folds,
when we condition on the data in the rest $K-1$ folds, $\hat{\mu}_{a^j}^{-k}(\cdot)$ is fixed
and each term $\pi_j^a(X_i) - \pi_j^b(X_i)$ (where $i$ is in $\{i \mid k(i) =k\}$) is \textbf{iid}.
Consequently, conditioned on $\hat{\mu}_{a^j}^{-k}(\cdot)$,
$S_1^{j,k}$ is a sum of \textbf{iid} bounded random variables with zero mean (and hence $\mathbf{E}\Big[S_1^{j,k}(\pi^a, \pi^b)\Big] = 0$), because:
\begin{equation}\label{eq:unbiased}
\begin{split}
&\mathbf{E}\Bigg[\Big(\pi_j^a(X_i) -\pi_j^b(X_i)\Big) \Big( 
\hat{\mu}_{a^j}^{-k(i)}(X_i) - \mu_{a^j}(X_i) \Big)\Big(1 - \frac{\mathbf{1}_{\{\W_i = a^j\}}}{e_{a^j}(X_i)} \Big)\Bigg] =\\
&\mathbf{E}\Bigg[\mathbf{E}\Bigg[\Big(\pi_j^a(X_i) -\pi_j^b(X_i)\Big) \Big( 
\hat{\mu}_{a^j}^{-k(i)}(X_i) - \mu_{a^j}(X_i) \Big)\Big(1 - \frac{\mathbf{1}_{\{\W_i = a^j\}}}{e_{a^j}(X_i)} \Big)\mid X_i\Bigg]\Bigg]=\\
&\mathbf{E}\Bigg[\Big(\pi_j^a(X_i) -\pi_j^b(X_i)\Big) \Big( 
\hat{\mu}_{a^j}^{-k(i)}(X_i) - \mu_{a^j}(X_i) \Big)\mathbf{E}\Bigg[1 - \frac{\mathbf{1}_{\{\W_i = a^j\}}}{e_{a^j}(X_i)} \mid X_i\Bigg]\Bigg] = 0,
\end{split}
\end{equation}
where the last equality follows from $\mathbf{E}\Bigg[\mathbf{1}_{\{\W_i = a^j\}}\mid X_i\Bigg] = e_{a^j}(X_i).$
Noting that $|\{i \mid k(i) =k\}| = \frac{N}{K}, \forall k = 1, 2, \dots, K$ (since the training data is divided into $K$ evenly-sized folds),
Equation~\eqref{eq:unbiased} then allows us to rewrite $\sup_{\pi^a, \pi^b \in \Pi}|S_1^{j,k}(\pi^a, \pi^b)|$ as follows:

\begin{equation}
\begin{split}
&\sup_{\pi^a, \pi^b \in \Pi}|S_1^{j,k}(\pi^a, \pi^b)| =
\sup_{\pi^a, \pi^b \in \Pi}|S_1^{j,k}(\pi^a, \pi^b) - \mathbf{E}\Big[S_1^{j,k}(\pi^a, \pi^b)\Big]|\\
&=
\frac{1}{K}\sup_{\pi^a, \pi^b \in \Pi}\Big|\frac{1}{\frac{n}{K}}\sum_{\{i \mid k(i) =k\}}  \Big(\pi_j^a(X_i) - \pi_j^b(X_i)\Big) \Big( 
\hat{\mu}_{a^j}^{-k(i)}(X_i) - \mu_{a^j}(X_i) \Big)\Big(1 - \frac{\mathbf{1}_{\{\W_i = a^j\}}}{e_{a^j}(X_i)} \Big)\Big| \\
& = \frac{1}{K}\sup_{\pi^a, \pi^b \in \Pi}\Bigg|\frac{1}{\frac{n}{K}}\sum_{\{i \mid k(i) =k\}}  \Bigg\{\Big(\pi_j^a(X_i) - \pi_j^b(X_i)\Big) \Big( 
\hat{\mu}_{a^j}^{-k(i)}(X_i) - \mu_{a^j}(X_i) \Big)\Big(1 - \frac{\mathbf{1}_{\{\W_i = a^j\}}}{e_{a^j}(X_i)} \Big) \\
&- \mathbf{E}\Big[\Big(\pi_j^a(X_i) -\pi_j^b(X_i)\Big) \Big( 
\hat{\mu}_{a^j}^{-k(i)}(X_i) - \mu_{a^j}(X_i) \Big)\Big(1 - \frac{\mathbf{1}_{\{\W_i = a^j\}}}{e_{a^j}(X_i)} \Big)\Big]\Bigg\}\Bigg|.\\
\end{split}
\end{equation}

Consequently, by identifying $\Big( 
\hat{\mu}_{a^j}^{-k(i)}(X_i) - \mu_{a^j}(X_i) \Big)\Big(1 - \frac{\mathbf{1}_{\{\W_i = a^j\}}}{e_{a^j}(X_i)} \Big)$ with $\factor_i$, we can apply Theorem~\ref{thm:uniform_counter} (and specializing it to the $1$-dimensional case) to obtain: $\forall \delta > 0$, with probability at least $1 - 2\delta$,
\begin{equation}\label{eq:sup_bound}
\begin{split}
& K\sup_{\pi^a, \pi^b \in \Pi}|S_1^{j,k}(\pi^a, \pi^b)| = \sup_{\pi^a, \pi^b \in \Pi}\Bigg|\frac{1}{\frac{n}{K}}\sum_{\{i \mid k(i) =k\}}  \Bigg\{\Big(\pi_j^a(X_i) - \pi_j^b(X_i)\Big) \Big( 
\hat{\mu}_{a^j}^{-k}(X_i) - \mu_{a^j}(X_i) \Big)\Big(1 - \frac{\mathbf{1}_{\{\W_i = a^j\}}}{e_{a^j}(X_i)} \Big) \\
&- \mathbf{E}\Big[\Big(\pi_j^a(X_i) -\pi_j^b(X_i)\Big) \Big( 
\hat{\mu}_{a^j}^{-k(i)}(X_i) - \mu_{a^j}(X_i) \Big)\Big(1 - \frac{\mathbf{1}_{\{\W_i = a^j\}}}{e_{a^j}(X_i)} \Big)\Big]\Bigg\}\Bigg|\le o(\frac{1}{\sqrt{n}}) + 
\Bigg(54.4\sqrt{2}\kappa(\Pi) + 435.2 \\
&+ \sqrt{2\log\frac{1}{\delta}}\Bigg) \sqrt{\frac{\sup_{\pi_a, \pi_b \in \Pi}\E\Bigg[\Big(\pi_j^a(X_i) -\pi_j^b(X_i)\Big)^2 \Big( 
		\hat{\mu}_{a^j}^{-k(i)}(X_i) - \mu_{a^j}(X_i) \Big)^2\Big(1 - \frac{\mathbf{1}_{\{\W_i = a^j\}}}{e_{a^j}(X_i)} \Big)^2\mid \hat{\mu}_{a^j}^{-k}(\cdot)\Bigg]}{\frac{n}{K}}} \\
&\le\Big(54.4\sqrt{2}\kappa(\Pi) + 435.2 + \sqrt{2\log\frac{1}{\delta}}\Big) \sqrt{\frac{\E\Bigg[\Big( 
		\hat{\mu}_{a^j}^{-k(i)}(X_i) - \mu_{a^j}(X_i) \Big)^2\Big(1 - \frac{\mathbf{1}_{\{\W_i = a^j\}}}{e_{a^j}(X_i)} \Big)^2 \mid \hat{\mu}_{a^j}^{-k}(\cdot) \Bigg]}{\frac{n}{K}}} + o(\frac{1}{\sqrt{n}}) \\
&\le (\frac{1}{\eta} - 1)^2 \Big(54.4\sqrt{2}\kappa(\Pi) + 435.2 + \sqrt{2\log\frac{1}{\delta}}\Big)
\sqrt{\frac{K\E\Bigg[\Big( 
		\hat{\mu}_{a^j}^{-k(i)}(X_i) - \mu_{a^j}(X_i) \Big)^2\mid \hat{\mu}_{a^j}^{-k}(\cdot) \Bigg]}{n}}+ o(\frac{1}{\sqrt{n}}),  \\
\end{split}
\end{equation}
where the second-to-last inequality follows from $\sup_{\pi_a, \pi_b \in \Pi}\Big(\pi_j^a(x) -\pi_j^b(x)\Big)^2 \le 1, \forall x \in \feas$ and the last inequality follows from the overlap assumption in Assumption~\ref{assump:classical}.
By Assumption~\ref{assump:consistency}, it follows that $\E\Bigg[\Big( 
\hat{\mu}_{a^j}^{-k(i)}(X_i) - \mu_{a^j}(X_i) \Big)^2\Bigg] \le \frac{s(\frac{K-1}{K}n)}{(\frac{K-1}{K}n)^{t_1}}$.
Consequently, Markov's inequality immediately implies that
$\E\Bigg[\Big( 
\hat{\mu}_{a^j}^{-k(i)}(X_i) - \mu_{a^j}(X_i) \Big)^2 \mid \hat{\mu}_{a^j}^{-k}(\cdot)\Bigg] = O_p\Big(\frac{s(\frac{K-1}{K}n)}{(\frac{K-1}{K}n)^{t_1}}\Big) =O_p\Big(\frac{s(\frac{K-1}{K}n)}{n^{t_1}}\Big)$.
Combining this observation with Equation~\eqref{eq:sup_bound}, we immediately have:
$\sup_{\pi^a, \pi^b \in \Pi}|S_1^{j,k}(\pi^a, \pi^b)| = O_p\Big(\frac{s(\frac{K-1}{K}n)}{n^{1+t_1}}\Big) + o(\frac{1}{\sqrt{n}}) = o_p(\frac{1}{\sqrt{n}}).$
Consequently, $$\sup_{\pi^a, \pi^b \in \Pi}|S_1^{j}(\pi^a, \pi^b)| = \sup_{\pi^a, \pi^b \in \Pi}|\sum_{k=1}^K S_1^{j,k}(\pi^a, \pi^b)| \le \sum_{k=1}^K\sup_{\pi^a, \pi^b \in \Pi}|S_1^{j,k}(\pi^a, \pi^b)| =  o_p(\frac{1}{\sqrt{n}}).$$
By exactly the same argument, we have $\sup_{\pi^a, \pi^b \in \Pi}|S_2^{j,k}(\pi^a, \pi^b)| = o_p(\frac{1}{\sqrt{n}})$, and hence $\sup_{\pi^a, \pi^b \in \Pi}|S_2^{j}(\pi^a, \pi^b)| \le \sum_{k=1}^K\sup_{\pi^a, \pi^b \in \Pi}|S_2^{j,k}(\pi^a, \pi^b)| =  o_p(\frac{1}{\sqrt{n}}).$

Next, we bound the contribution from $S_3^j(\cdot, \cdot)$ as follows:
\begin{equation}
\begin{split}
&\sup_{\pi^a,\pi^b\in \Pi} |S_3^j(\pi^a, \pi^b)| = \frac{1}{n}\sup_{\pi^a,\pi^b\in \Pi} \Bigg|\sum_{\{i \mid \W_i =a^j\}} \Big(\pi_j^a(X_i) - \pi_j^b(X_i)\Big) \Big( 
\mu_{a^j}(X_i)  - \hat{\mu}_{a^j}^{-k(i)}(X_i)\Big)\Big(\frac{1}{\hat{e}_{a^j}^{-k(i)}(X_i)} - \frac{1}{e_{a^j}(X_i)} \Big) \Bigg|\\
& \le \frac{1}{n}\sup_{\pi^a,\pi^b\in \Pi} \sum_{\{i \mid \W_i =a^j\}} \Big|\Big(\pi_j^a(X_i) - \pi_j^b(X_i)\Big)\Big| \cdot \Big|\Big( 
\mu_{a^j}(X_i)  - \hat{\mu}_{a^j}^{-k(i)}(X_i)\Big)\Big|\cdot \Big|\Big(\frac{1}{\hat{e}_{a^j}^{-k(i)}(X_i)} - \frac{1}{e_{a^j}(X_i)} \Big) \Big|\\
& \le \frac{1}{n} \sum_{\{i \mid \W_i =a^j\}} \Big|\Big( 
\mu_{a^j}(X_i)  - \hat{\mu}_{a^j}^{-k(i)}(X_i)\Big)\Big|\cdot \Big|\Big(\frac{1}{\hat{e}_{a^j}^{-k(i)}(X_i)} - \frac{1}{e_{a^j}(X_i)} \Big) \Big|\\
&\le   \sqrt{\frac{1}{n} \sum_{\{i \mid \W_i =a^j\}} \Big( 
	\mu_{a^j}(X_i)  - \hat{\mu}_{a^j}^{-k(i)}(X_i)\Big)^2}\sqrt{\frac{1}{n} \sum_{\{i \mid \W_i =a^j\}} \Big(\frac{1}{\hat{e}_{a^j}^{-k(i)}(X_i)} - \frac{1}{e_{a^j}(X_i)} \Big)^2},\\
\end{split}
\end{equation}
where the last inequality follows from Cauchy-Schwartz. Taking expectation of both sides yields:
\begin{equation}\label{eq:S_3}
\begin{split}
&\E\Big[\sup_{\pi^a,\pi^b\in \Pi} |S_3^j(\pi^a, \pi^b)| \Big] \le \E\Bigg[\sqrt{\frac{1}{n} \sum_{\{i \mid \W_i =a^j\}} \Big( 
	\mu_{a^j}(X_i)  - \hat{\mu}_{a^j}^{-k(i)}(X_i)\Big)^2}\sqrt{\frac{1}{n} \sum_{\{i \mid \W_i =a^j\}} \Big(\frac{1}{\hat{e}_{a^j}^{-k(i)}(X_i)} - \frac{1}{e_{a^j}(X_i)} \Big)^2} \Bigg]\\
& \le \sqrt{ \E\Bigg[\frac{1}{n} \sum_{\{i \mid \W_i =a^j\}} \Big( 
	\mu_{a^j}(X_i)  - \hat{\mu}_{a^j}^{-k(i)}(X_i)\Big)^2\Bigg]}\sqrt{\E\Bigg[\frac{1}{n} \sum_{\{i \mid \W_i =a^j\}} \Big(\frac{1}{\hat{e}_{a^j}^{-k(i)}(X_i)} - \frac{1}{e_{a^j}(X_i)} \Big)^2 \Bigg]} \\
&  = \sqrt{ \frac{1}{n}\sum_{\{i \mid \W_i =a^j\}} \E\Bigg[ \Big( 
	\mu_{a^j}(X_i)  - \hat{\mu}_{a^j}^{-k(i)}(X_i)\Big)^2\Bigg]}\sqrt{\frac{1}{n}\sum_{\{i \mid \W_i =a^j\}} \E\Bigg[ \Big(\frac{1}{\hat{e}_{a^j}^{-k(i)}(X_i)} - \frac{1}{e_{a^j}(X_i)} \Big)^2 \Bigg]} \\
& \le \sqrt{ \frac{1}{n}\sum_{\{i \mid \W_i =a^j\}} \frac{s(\frac{K-1}{K}n)}{\Big(\frac{K-1}{K}n\Big)^{t_1}} }\sqrt{\frac{1}{n}\sum_{\{i \mid \W_i =a^j\}} \frac{s(\frac{K-1}{K}n)}{\Big(\frac{K-1}{K}n\Big)^{t_2}}} 
\le \sqrt{  \frac{s(\frac{K-1}{K}n)}{\Big(\frac{K-1}{K}n\Big)^{t_1}} }\sqrt{ \frac{s(\frac{K-1}{K}n)}{\Big(\frac{K-1}{K}n\Big)^{t_2}}}\\
& = 
\frac{s(\frac{K-1}{K}n)}{\sqrt{\Big(\frac{K-1}{K}n\Big)^{t_1 + t_2}}}
\le  \frac{s(\frac{K-1}{K}n)}{\sqrt{\frac{K-1}{K}n}} = o(\frac{1}{\sqrt{n}}),
\end{split}
\end{equation}
where the second inequality again follows from Cauchy-Schwartz, the third inequality follows from Assumption~\ref{assump:consistency}, Remark~\ref{assump:consistency_eqv} and the fact that each estimator $\hat{u}^{-k}(\cdot)$ is trained on $\frac{K-1}{K}n$ data points, the last inequality follows from Assumption~\ref{assump:consistency} and the last equality follows from $s(n) = o(1)$.
Consequently, by Markov's inequality, Equation~\eqref{eq:S_3} immediately implies $\sup_{\pi^a,\pi^b\in \Pi} |S_3^j(\pi^a, \pi^b)|  = o_p(\frac{1}{\sqrt{n}})$. 
Putting the above bounds for $\sup_{\pi^a,\pi^b\in \Pi} |S_1^j(\pi^a, \pi^b)|  = o_p(\frac{1}{\sqrt{n}})$, $\sup_{\pi^a,\pi^b\in \Pi} |S_2^j(\pi^a, \pi^b)|  = o_p(\frac{1}{\sqrt{n}})$ and $\sup_{\pi^a,\pi^b\in \Pi} |S_3^j(\pi^a, \pi^b)|  = o_p(\frac{1}{\sqrt{n}})$ together, we therefore have the claim established:
\begin{equation}
\begin{split}
& \sup_{\pi^a,\pi^b\in \Pi} |\diffd_{DR}^{j} (\pi^a, \pi^b) - \diffi^{j} (\pi^a, \pi^b)| = \sup_{\pi^a,\pi^b\in \Pi} |\sum_{j=1}^d\diffd_{DR}^{j} (\pi^a, \pi^b) - \diffi^{j} (\pi^a, \pi^b)| \\
&\le \sum_{j=1}^d  \sup_{\pi^a,\pi^b\in \Pi}  |\diffd_{DR}^{j} (\pi^a, \pi^b) - \diffi^{j} (\pi^a, \pi^b)|
= \sum_{j=1}^d \sup_{\pi^a,\pi^b\in \Pi} |S_1^j(\pi^a, \pi^b) + S_2^j(\pi^a, \pi^b) +S_3^j(\pi^a, \pi^b)|  \\
& \le \sum_{j=1}^d \sup_{\pi^a,\pi^b\in \Pi} \Big(|S_1^j(\pi^a, \pi^b) |+ |S_2^j(\pi^a, \pi^b)| +|S_3^j(\pi^a, \pi^b)|\Big) \\
&\le 
\sum_{j=1}^d \sup_{\pi^a,\pi^b\in \Pi} \Big(|S_1^j(\pi^a, \pi^b) |+ \sum_{j=1}^d \sup_{\pi^a,\pi^b\in \Pi}|S_2^j(\pi^a, \pi^b)| +\sum_{j=1}^d \sup_{\pi^a,\pi^b\in \Pi}|S_3^j(\pi^a, \pi^b)|\Big) = o_p(\frac{1}{\sqrt{n}}).\\
\end{split}
\end{equation}
\hfill $\blacksquare$

\section{Proof of Lemma~\ref{lem:entropy_integral}}


We fix any $n > 0$ and $n$ points $x_1, \dots, x_n \in \feas$. 
\setcounter{proofstep}{0}
\begin{proofstep}{Universal approximations of tree paths}\label{step:2}
For any $b \in \mathbf{R}$, denote $\mathbf{1}_b(\cdot)$ to be the indicator function on reals:
$\mathbf{1}_b(r) = \left\{
\begin{array}{ll}
1, & \textbf{if } r < b  \\
0, & \textbf{if } r \ge b.\\
\end{array}
\right.$
We show that given any $m$ real numbers $r_1, \dots, r_m$,
we can always find $\ceil{\frac{L}{\epsilon}}$ indicator functions
$\mathbf{1}_{b_1}(\cdot), \mathbf{1}_{b_2}(\cdot), \dots, \mathbf{1}_{b_{\ceil{\frac{L}{\epsilon}}}}(\cdot)$, such that: 
\begin{equation}\label{eq:indicator_approximate}
\min_{i \in \{1,2\dots, n\}} H(\mathbf{1}_{b}(\cdot), \mathbf{1}_{b_i}(\cdot)) \le \frac{\epsilon}{L}, \forall \mathbf{1}_{b}(\cdot),
\end{equation}
where $H(\mathbf{1}_{b}(\cdot), \mathbf{1}_{b_i}(\cdot))$ is the Hamming distance  between 
the two functions $\mathbf{1}_{b}(\cdot)$ and $\mathbf{1}_{b_i}(\cdot)$ with respect to $r_1, \dots, r_m$
and in the current setting can be written equivalently as:
$H(\mathbf{1}_{b}(\cdot), \mathbf{1}_{b_i}(\cdot)) =
\frac{1}{m} \sum_{j=1}^m \Big|\mathbf{1}_{b}(r_j) -\mathbf{1}_{b_i}(r_j)\Big|.$

To see this, note that it is without loss of generality to take these $m$ numbers to be distinct: otherwise, the duplicated numbers can be removed to create the same set of distinct numbers with a smaller $m$.
We start by sorting the $m$ real numbers in ascending order (and relabelling them if necessary) such that:
$r_1 < r_2 < \dots < r_m.$ Then any $b$ must fall into exactly one of the following
$m+1$ disjoint intervals: $(-\infty, r_1], (r_1, r_2], (r_2, r_3], \dots, (r_m, \infty)$. 
Per the definition of indicator functions, it follows immediately that
for any $b, \tilde{b}$ that fall in the same interval, $\mathbf{1}_{b}(r_j) = \mathbf{1}_{\tilde{b}}(r_j), \forall j \in \{1,2, \dots, m\}$, which is a consequence of partitioning the real line using the left-open-right-closed intervals $(r_j, r_{j+1}]$. Consequently, for such $b$ and $\tilde{b}$,
$H(\mathbf{1}_{b}(\cdot), \mathbf{1}_{\tilde{b}}(\cdot)) = 0.$

To establish the uniform approximation result, we note that there are two possibilities:
\begin{enumerate}
	\item $\frac{\epsilon m }{L} < 1$. In this case, $\ceil{\frac{L}{\epsilon}} \ge m+1$, and we can hence
	place a $b_i$ in each of the $m+1$ intervals (and arbitrarily place the remaning $b_i$'s).
	With this configuration, for any $b$, we can always find a $b_i$ where both $b$ and $b_i$ are in the same interval, in which case $\mathbf{1}_{b}(r_j) = \mathbf{1}_{b_i}(r_j), \forall j \in \{1,2\dots, m\}$ and hence
	$H(\mathbf{1}_{b}(\cdot), \mathbf{1}_{b_i}(\cdot)) =
	\frac{1}{m} \sum_{j=1}^m \Big|\mathbf{1}_{b}(r_j) -\mathbf{1}_{b_i}(r_j)\Big| = 0.$
	\item $\frac{\epsilon m }{L} \ge 1$. In this case, $\ceil{\frac{L}{\epsilon}} \le m$, and we can
	place $\mathbf{1}_{b_1}(\cdot), \dots, \mathbf{1}_{b_{\ceil{\frac{L}{\epsilon}}}}(\cdot)$ sequentially as follows:
	\begin{enumerate}
		\item Place $b_1$ (anywhere) in the interval $(-\infty, r_1)$.
		\item For each $i \ge 1$, Place $b_{i+1}$ in the interval that is $\ceil{\frac{\epsilon m }{L}}$ intervals away from the interval $b_i$ is in, which we refer to for simplicity that $b_i$ is $\ceil{\frac{\epsilon m }{L}}$ intervals away from $b_i$. More specifically, the interval $[r_i, r_{i+1})$ is $1$ interval away from $[r_{i-1}, r_i)$. Note that under this definition,
		if $b_i$ is $l$ interval intervals away from $b$, then there are $l$ $r_j$'s between $b_i$ and $b$. 
	\end{enumerate}
	With this configuration, any $b$ is sandwiched between $b_i$ and $b_{i+1}$ (for some $i$)
	that are $\ceil{\frac{\epsilon m }{L}}$ intervals apart. Consequently,
	one of $b_i$ and $b_{i+1}$ must be at most $\frac{1}{2}\ceil{\frac{\epsilon m }{L}}$ intervals away from $b$.
	Without loss of generality, assume that is $b_i$.
	Then there are at most $\frac{1}{2}\ceil{\frac{\epsilon m }{L}}$ $r_j$'s
	between $b$ and $b_i$: these are the only $r_j$'s where
	$\mathbf{1}_{b}(r_j) \neq \mathbf{1}_{b_i}(r_j)$. Consequently,
	$$H(\mathbf{1}_{b}(\cdot), \mathbf{1}_{b_i}(\cdot)) =
	\frac{1}{m} \sum_{j=1}^m \Big|\mathbf{1}_{b}(r_j) -\mathbf{1}_{b_i}(r_j)\Big| \le \frac{1}{2m}\ceil{\frac{\epsilon m }{L}} \le \frac{1}{m}\frac{\epsilon m }{L} = \frac{\epsilon}{L}.$$
\end{enumerate}
Combining the above two cases yields the conclusion given in Equation~\eqref{eq:indicator_approximate}.

Given any two indicator functions $\mathbf{1}_b(\cdot)$ and $\mathbf{1}_{b^{\prime}}(\cdot)$, we can define:
$\mathbf{1}_b(\cdot) \wedge \mathbf{1}_{b^{\prime}}(\cdot) = \mathbf{1}_b(\cdot) \cdot \mathbf{1}_{b^{\prime}}(\cdot).$
The above argument establishes that for any set of $m_1$ real numbers $r_1, \dots, r_{m_1}$, we can find $\mathbf{1}_{b_1}(\cdot), \mathbf{1}_{b_2}(\cdot), \dots, \mathbf{1}_{b_{\ceil{\frac{L}{\epsilon}}}}(\cdot)$, such that for any  
$\mathbf{1}_{b}(\cdot)$:
$\min_{i \in \{1,2\dots, n\}} H(\mathbf{1}_{b}(\cdot), \mathbf{1}_{b_i}(\cdot)) \le \frac{\epsilon}{L},$ where the Hamming distance is with repsect to $r_1, \dots, r_{m_1}$. Similarly, for any set of $m_2$ real numbers $r_1^\prime, \dots, r_{m_2}^\prime$, we can find $\mathbf{1}_{b_1^\prime}(\cdot), \mathbf{1}_{b_2^\prime}(\cdot), \dots, \mathbf{1}_{b^\prime_{\ceil{\frac{L}{\epsilon}}}}(\cdot)$, such that for any  
$\mathbf{1}_{b^\prime}(\cdot)$:
$\min_{i \in \{1,2\dots, n\}} H(\mathbf{1}_{b^\prime}(\cdot), \mathbf{1}_{b^\prime_i}(\cdot)) \le \frac{\epsilon}{L},$  where the Hamming distance is with repsect to $r_1^\prime, \dots, r^\prime_{m_2}$.
Since each tree path is simply a conjuction of $L$ indicator functions,
induction then immediately establishes the claim.

	Next, fix an assignment of split variables to each branch node.
Then, each branch node is completely determined by the threshold value, 
and consists of a single boolean clause: $x(i) < b$, where $x(i)$ is the split variable chosen for this node, and $b$ is the threshold value.
This boolean clause can be equivalently represented by $\mathbf{1}_b(x(i))$ (i.e. $x(i) < b$ evaluates to true if and only if $\mathbf{1}_b(x(i)) = 1$).
Further, each tree path from the root node to a leaf node is a conjunction of $L$ such boolean clauses, which can then be equivalently represented by multiplying the
$L$ corresponding indicator functions: a leaf node is reached if and only if
all the indicator variables along the tree path to that leaf evaluates to $1$.
Consequently, each tree path is a function that maps $x$ to $\{0, 1\}$ that
takes the form $\Pi_{j=1}^L \mathbf{1}_{b_j}(x(i_j))$.
Utilizing the conclusion from the previous step, and by a straightforward but lengthy induction (omitted),
we establish that by using $\ceil{\frac{L}{\epsilon}}$ indicator functions at each branch node (which may be different across branch nodes), every tree path can be approximated within $\epsilon$ Hamming distance (with respect to $x_1,\dots, x_n$ fixed at the beginning).
\end{proofstep}

\begin{proofstep}{Universal approximations of trees}\label{step:3}
	A tree is specified by the following list of parameters:
	an assignment of the split variables for each branch node;
	a split threshold for each branch node;
	the action assigned for each leaf node.
	\begin{enumerate}
		\item An assignment of the split variables for each branch node.
		\item A split threshold for each branch node.
		\item The action assigned for each leaf node.
	\end{enumerate}
	There are $2^L -1$ branch nodes. For each branch node, there are $p$ possible split variables to choose from. Once all the split variables are chosen for each node, we need to place $\ceil{\frac{L}{\epsilon}}$ indicator variables at each of the $2^L -1$ branch nodes.
	Finally, there are $2^L$ leaves, and for each leaf node, there are $d$ possible ways to assign the action labels, one for each action. Consequently, we need at most
	$p^{2^{L}-1} \ceil{\frac{L}{\epsilon}}^{2^L -1} d^{2^L}$ trees to approximate any depth-$L$ tree within $\epsilon$-Hamming distance (with respect to $x_1,\dots, x_n$ fixed at the beginning).
	This immediately yields that $\N_H(\epsilon, \Pi_L, \{x_1, \dots, x_n\}) \le p^{2^{L}-1} \ceil{\frac{L}{\epsilon}}^{2^L -1} d^{2^L}$. Since this is true for any set of points  $\{x_1, \dots, x_n\}$, 
	we have $\N_H(\epsilon, \Pi_L) \le p^{2^{L}-1} \ceil{\frac{L}{\epsilon}}^{2^L -1} d^{2^L}$, thereby yielding:
	$\log \N_H(\epsilon, \Pi_L) \le (2^{L}-1)\log p + (2^L -1) \log \ceil{\frac{L}{\epsilon}} + 2^L\log d \le (2^{L}-1)\log p + (2^L -1) \log (\frac{L}{\epsilon} +1) + 2^L\log d = O(\log \frac{1}{\epsilon}) $.
	Finally, by computing a integral on the covering number, we obtain the bound on $\kappa(\Pi)$.
	\begin{align*}
	&\kappa(\Pi) = \int_0^1 \sqrt{\log \N_H(\epsilon^2, \Pi)}d\epsilon
	=  \int_0^1 \sqrt{ (2^{L}-1)\log p + 2^L\log d  + (2^L -1) \log (\frac{L}{\epsilon^2} +1)   }d\epsilon\\
	&\le \int_0^1\sqrt{ (2^{L}-1)\log p + 2^L\log d } d\epsilon+ \sqrt{ (2^L -1) }\int_0^1 \sqrt{\log (\frac{L}{\epsilon^2} +1)   }d\epsilon\\
	&= \sqrt{ (2^{L}-1)\log p + 2^L\log d } + \sqrt{ (2^L -1) }\int_0^1 \sqrt{\log (\frac{L}{\epsilon^2} +1)   }d\epsilon \\
	&\le \sqrt{ (2^{L}-1)\log p + 2^L\log d } + \sqrt{ (2^L -1) }\int_0^1 (\frac{L}{\epsilon^2})^{\frac{1}{4}}
	d\epsilon\\
	&= \sqrt{ (2^{L}-1)\log p + 2^L\log d } + \frac{4}{3}L^{\frac{1}{4}}\sqrt{ (2^L -1) },
	\end{align*}
	where the last inequality follows from the (rather crude) inequality $\log (x + 1) \le \sqrt{x}, \forall x \ge 0$. 
\end{proofstep}

\section{Complexity Discussion of Tree Search Algorithm}\label{subsec:complexity}

Here we provide a disucssion on the complexity of Algorithm~\ref{alg:ts}:
\begin{enumerate}
	\item The return value of the tree-search algorithm is a pair that consists of the maximum reward obtainable for the given data and the tree that realizes this maximum reward. In the algorithm description, we used an array-based representation for trees: when a tree is depth-$1$ (i.e. a single leaf-node), it is specified by a single positive integer (representing which action to be taken); when a tree is beyond depth-$1$, each leaf node is still specified by a single positive integer representing the action while each branch node is represented by two numbers (a positive integer specifying which variable to split on and a real number specifying the split value). Note that in this algorithm, unlike MIP, we do not need to normalize all the features to be between $0$ and $1$, which simplifies the pre-processing step for raw data.
	\item The tree-search algorithm also provides a flexible framework that allows accuracy to be traded-off with computational efficiency. In particular, the algorithm takes as input the approximation parameter $A$,
	which specifies how many points to skip over when performing the search. $A = 1$ corresponds to exact search where no accuracy is lost. The larger the $A$ is, the more crude the approximation becomes, and the faster the running time results. In practice, particularly when a lot of training data is available, features tend to be densely packed in an interval. Consequently, skipping $A$ features every time tends to result in neglible impact for small $A$.
	\item Several engineering optimizations can be performed to make the code run faster (both asymptotically and practically). For instance, Step $7$ of Algorithm~\ref{alg:ts} requires sorting the features in every FOR loop of every call on the subtree. This takes $O(n\log n)$ per call in the worst case. Instead, one can sort all the features in each dimension at the beginning and use an appropriate data-structure to put all the (subset of) data in order in $O(n)$ per call.
	As another example, since each recursive call on the subtree is executed many times, one can also avoid redundant computation by saving the optimal tree computed for a given subset of data (i.e. using memoization to speed up recursive computation).
	To avoid clutter, we do not incorporate these engineering details into the description of the tree-search algorithm, although they are implemented in the solver. 
	\item A few words on the running time of tree-search under exact search (i.e $A= 1$), the running time 
	scales $O(2^{L-1}(np)^{L-1}d + p n\log n)$ for $L \ge 2$ (for $L = 1$, the running time is $O(nd)$ since global sorting is not needed); in particular, when $L \ge 3$, the running time is
	$O(2^{L-1}(np)^{L-1}d)$. To see this, first note that even though naively
	it takes $O(nd)$ to identify the best action at a leaf for $n$ data points, one can collapse the leaf layer and the last branch layer to achieve a joint $O(npd)$ running time by a dynamic programming style algorithm (details omitted). Consequently, combined with the global sorting at the beginning,
	when $L=2$, the running time is $O(npd + pn\log n)$. For a general $L$, we denote $T(L, n, p, d)$ to be the running time under the corresponding scale parameters.
	Since at each level we need to search through $O(np)$ possibilities, and since each of these possibilities 
	results in recursively calling the same tree-search algorithm on a tree that is one level shallower,
	we have: 
	$$T(L, n,  p, d) \le np + 2npT(L-1, n, p, d),$$
	where the first $np$ on the right-hand side corresponds to going through the globally sorted data to put the subset of datapoints allocated to this node in sorted order (there are $p$ dimensions, and each dimension has $n$ points) and the second $2np$ on the right-hand side corresponds to the total number of recursive calls made for the subtrees ($np$ calls for the left subtree and $np$ calls for the right subtree).
	From this recursion, one can compute that the total running time is $O(2^{L-1}(np)^{L-1}d + p n\log n)$. Note that for a fixed depth $L$, the running time is now polynomial. Another thing to note is that the above running time is obtained when all the feature values are distinct (i.e. when features assume continuous values). However, when each feature can only assume a finite number of discrete values, the running time will improve, since the search will only occur at places where features change.
	Specifically, if the $i$-th feature takes $K_i$ possible values, then we have the following recursion:
	$$T(L, n,  p, d) \le np + 2 (\sum_{i=1}^p K_i ) T(L-1, n, p, d),$$
	where the first $np$ on the right-hand side again corresponds to going through the globally sorted data to put the subset of datapoints allocated to this node in sorted order (there are $p$ dimensions, and each dimension has $n$ points); but now there are only $2 (\sum_{i=1}^p K_i )$ recursive calls to be made for two the subtrees since the algorithm only searches at points where features change. 
	In this case, we still have $O(npd + pn\log n)$ time for $L = 2$ since one still needs to pass through the entire data to identify the best action for each dimension and each action (which amounts to $O(pnd)$ time total) and $O(pn\log n)$ for global sorting\footnote{Since features are now discrete, the sorting can be done in $O(pn)$ time using bucket sort, resulting in $O(pnd)$ total time. However, in practice, bucket sort may have a larger constant than quicksort, and the timing savings, if any, are inconsequential. Consequently, we stick with the $O(npd + pn\log n)$ running time for $L = 2$, even though it is slightly worse than $(npd)$. Also, when $L = 1$, the running time is still $O(nd)$.}.
	Combining this base case with the recursion established earlier, one can then derive
	that the total running time is
	$O(2^{L-1}(\sum_{i=1}^p (K_i))^{L-2}pnd + pn\log n ), L \ge 2$.
	Note that in the continuous feature case, each $K_i = n$ and hence $\sum_{i=1}^p (K_i) = np$,
	which degenerates to the previously established running time  $O(2^{L-1}(np)^{L-1}d + p n\log n)$.
	It is important to point out that time savings can be quite significant when working with discrete features.
	For instance, as commonly in practice, when all features are binary, the running time becomes $O(2^{L-1}p^{L-1}nd + pn\log n )$, which is (essentially) linear in the number of data points $n$. More generally, the running time will be near-linear if the size of the feature domain is small and the dimension is not too high.
	
	\item  Algorithm~\ref{alg:ts} also provides a convenient interface that allows one to make some approximation in order to gain further computational efficiency. The main idea behind the approximation algorithm is that when $n$ is large, many features can be closely clustered around each other. In such settings, searching over all breakpoints (between two distinct features) can be wasteful. Instead, it is more efficient to skip over several points at once when searching. This is akin to dividing the entire feature interval into, for instance, percentiles, and then only search at the endpoint of each percentile. However,
	the downside of this approach is that pre-dividing the interval in this fixed way may not leverage the data efficiently, because data may be dense in certain regions and sparse in others. Consequently, choosing to skip points is a dynamic way of dividing the feature space into smaller subspaces that bypasses the drawback brought about by the fixed division scheme.
	In the algorithm, $A$ is the approximation parameter that controls how aggressive one is in skipping: the larger the $A$, the more aggressive the skipping becomes. From a computational standpoint, if we skip $A$ points each time, then the total number of recursive calls decrease by a factor of $A$ at each node, and will hence result in significant time savings. In particular, by a similar calculation, we can show that the total running time when skipping $A$ points each time is $O(2^{L-1}npd(\frac{n}{A}p)^{L-2} + p n\log n)$ for $L \ge 2$.
\end{enumerate}
\end{APPENDIX}

\end{document}